\documentclass[10pt,twocolumn,letterpaper]{article}

\usepackage{iccv}
\usepackage{times}
\usepackage{epsfig}
\usepackage{graphicx}
\usepackage{amsmath}
\usepackage{amssymb}
\usepackage{color}

\usepackage{soul}

\usepackage{footmisc}

\newcommand{\argmax}{\operatornamewithlimits{argmax}}

\newcommand{\sign}{\operatornamewithlimits{sign}}

\usepackage{url}

\usepackage[numbers]{natbib}
\usepackage{enumitem}

\usepackage{fixmath}
\renewcommand{\vec}[1]{\mathbold{#1}}

\newcommand{\cV}{\mathcal{V}}
\newcommand{\cE}{\mathcal{E}}

\renewcommand{\phi}{\varphi}

\usepackage[pagebackref=true,breaklinks=true,letterpaper=true,colorlinks,bookmarks=false]{hyperref}
\hypersetup{pdfauthor={Tuan-Hung Vu, Anton Osokin, Ivan Laptev},pdftitle={Context-aware CNNs for person head detection}}

\iccvfinalcopy 

\def\iccvPaperID{758} 

\ificcvfinal\fi

\begin{document}

\title{
Context-aware CNNs for person head detection
}
\author{Tuan-Hung Vu\thanks{WILLOW project-team, D\'epartment d'Informatique de l'Ecole Normale Sup\'erieure, ENS/INRIA/CNRS
UMR 8548, Paris, France } \qquad Anton Osokin\thanks{SIERRA project-team, D\'epartment
d'Informatique de l'Ecole Normale Sup\'erieure, ENS/INRIA/CNRS UMR 8548, Paris, France} \qquad Ivan Laptev\footnotemark[1]\vspace{.3cm}\\
INRIA/ENS\vspace{-.0cm}\\
%
%
}

\maketitle
\thispagestyle{empty}

\begin{abstract}

Person detection is a key problem for many computer vision tasks. While face detection has reached maturity, detecting people under a full variation of camera view-points, human poses, lighting conditions and occlusions is still a difficult challenge. In this work we focus on detecting human heads in natural scenes. Starting from the recent local \mbox{R-CNN} object detector, we extend it with two types of contextual cues.
First, we leverage person-scene relations and propose a Global CNN model trained to predict positions and scales of heads directly from the full image.
Second, we explicitly model pairwise relations among objects and
train a Pairwise CNN model using a structured-output surrogate loss.
The Local, Global and Pairwise models are combined into a joint CNN framework.
To train and test our full model, we introduce a large dataset composed of $369,846$ human heads annotated in $224,740$ movie frames. We evaluate our method and demonstrate improvements of person head detection against several recent baselines in three datasets. We also show improvements of the detection speed provided by our model.

\end{abstract}

\section{Introduction}

Common images and videos primarily focus on people. Indeed, about $35\%$ of pixels in movies and YouTube videos as well as about $25\%$ of pixels in photographs belong to people~\cite{Laptev13hdr}. This strong bias together with the growing amount of daily videos and photographs urge reliable methods for person analysis in visual data.

Person detection is a key component for many tasks including person identification, action recognition, age and gender recognition, autonomous driving, cloth recognition and many others. While face detection has reached maturity~\cite{mathias2014face}, the more general task of finding people in images and video still remains to be very challenging. For example, state-of-the-art object detectors~\cite{girshick14} reach only $65\%$ Average Precision for the person class on the Pascal VOC benchmark. Common difficulties arise from variations in human pose, background clutter, motion blur, low image resolution, occlusions and poor lighting conditions. 

\begin{figure}[t!]
    \centering
    \includegraphics[trim = 10mm 0mm 10mm 0mm, clip, width=\linewidth]{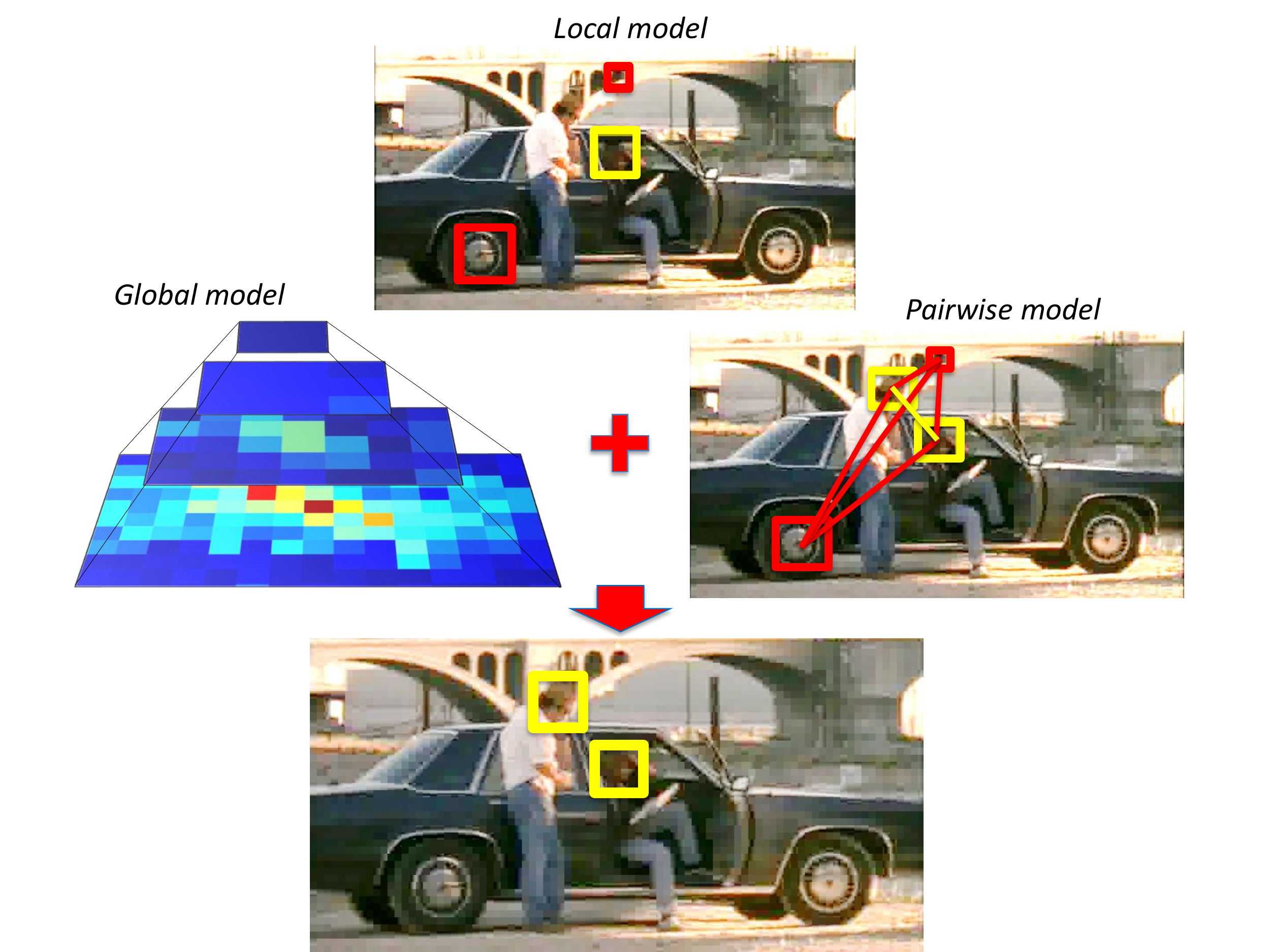}\vspace{.1cm}
    \caption{\small
             Results of head detection for a sample movie frame.
             The output of our method (bottom) is obtained from the combination of Local, Global and Pairwise CNN models.
             Bounding boxes illustrate detections: yellow~-- correct, red~-- false.
             Links between detections correspond to the pairwise potentials of our model: yellow~-- attractive, red~-- repulsive.
    }
    \label{fig:teaser}
\end{figure}

Recent advances in Convolutional Neural Networks (CNN)~\cite{lecun98} have brought significant progress in image classification~\cite{krizhevsky12} and other vision tasks. In particular, CNN-based object detectors such as R-CNN~\cite{girshick14} have shown large gains compared to previous models~\cite{felzenszwalb10dpm,Vandeaande11}. Most of existing methods, however, treat objects independently and model appearance inside object bounding boxes only. Meanwhile, information available in the scene around objects~\cite{Torralba03contextual} as well as relations among objects~\cite{desai11} are known to provide complementary contextual cues for recognition. Such cues are likely to be particularly helpful when object appearance lacks discriminative cues due to low image resolution, poor lighting and other factors.

In this work we build on the recent CNN model for object detection~\cite{girshick14} and extend it to contextual reasoning. 
We particularly focus on person detection and aim to locate human heads on images coming from video data. The choice of heads is motivated by frequent occlusions of other body parts. When visible, however, other body parts and the rest of the scene constrain locations of heads in the image. Moreover, interactions between people put constraints on the relative positions and appearance of heads. We aim to leverage such constraints for detection by introducing the following two models.

First, we propose a {\em Global} CNN model which we train to predict coarse locations and scales of objects given the full low-resolution image on the input. In contrast to our base {\em Local} model limited to object appearance only, the Global model uses all pixels of the image for prediction. Interestingly, we find this simple model to provide quite accurate localization of heads across positions and scales of the image. Second, we introduce a {\em Pairwise} CNN model that explicitly models relations among pairs of objects. Motivated by~\citet{desai11}, we build a joint score function for multiple object hypotheses in the image. This score function considers the relative positions, scales and appearance of heads. All parameters of the score function depend on the image data and are learned by optimizing a structured-output loss function. 
Our final joint model combines Local, Global and Pairwise CNN models (see Figure~\ref{fig:teaser}).

To train and test our model, we introduce a new large dataset with $369,846$ human heads annotated in $224,740$ video frames from 21 movies. We show the importance of our large dataset for training and evaluate our method on the new and two existing datasets. The results demonstrate improvements of the proposed contextual CNN model compared to other recent baselines including R-CNN~\cite{girshick14} on all three datasets. We also demonstrate a speed-up of object detection provided by our Global model. Our new dataset and the code are publicly available from the project web-page~\cite{projectwebpage}.

The rest of the paper is organized as follows. We review related work in Section~\ref{sec:relatedwork}. Section~\ref{sec:model} describes the parts of our contextual CNN model. Section~\ref{sec:datasets} introduces datasets followed by the presentation of experimental results in Section~\ref{sec:experiments}. Section~\ref{sec:conclusions} concludes the paper.

\section{Related works \label{sec:relatedwork}}

The history of object detection with neural networks dates back to the 90s~\cite{Vaillant93}, but methods of these group have started to outperform others, e.g DPM~\cite{felzenszwalb10dpm}, only after the seminal work of~\citet{krizhevsky12}.
\citet{szegedy13} and \citet{sermanet14} applied CNN as a sliding window detector at multiple scales.
The R-CNN model~\cite{girshick14} is a combination of a CNN and a support vector machine (SVM) operating on object proposals generated by the selective search~\cite{uijlings2013selective}. The pipeline of our Local model is similar to the one of R-CNN (see Section~\ref{sec:localmodel} for details).

The use of image context was proposed to support object detection in~\cite{Torralba03contextual}.
Contextual information can be modeled at a global scene level as well as at the level of object relations.
%
%
For example, \citet{murphy2003using} propose a CRF model for jointly solving the task of object detection and scene classification.
\citet{modolo2015context} uses context forest to predict object location and to speed-up object detection using global scene information.
\citet{erhan2014} uses CNN to predict coordinates of object bounding boxes. 
Our Global CNN model predicts likely locations and scales of objects by producing a multi-scale heat map for the whole image.

\citet{desai11} models spatial constellations of objects in the image and constructs an energy with unary and pairwise potentials.
Unary potentials represent the confidence of object hypotheses based on the local image evidence, while pairwise potentials model spatial arrangement of objects in the image. \citet{hoai14} substitute the pairwise dependencies with a latent variable that represents the preferable configuration of object hypotheses. In both works~\cite{desai11,hoai14} binary potentials do not depend on the actual image data, moreover, unary potentials are trained independently of the joint model.
Our Pairwise model exploits object context, i.e.\ builds a graphical model (an energy function) reasoning about multiple image locations jointly.
Our approach is richer compared to~\cite{desai11} and~\cite{hoai14} as it allows pairwise dependencies to be conditioned on the image data and we can train the base detector jointly with the graphical model on top of it.

Our Pairwise CNN model incorporates the structured-output loss. The idea of combining the structured-prediction objective with neural networks has been explored in~\cite{bottou97,lecun98}. 
Recently~\citet{domke2013nips} and~\citet{chen14} use the dual message passing formulation of the inference task to construct a joint objective of the CNN parameters and the message-passing variables.
This approach was applied to the small scale denoising and binary segmentation tasks in~\cite{domke2013nips} and to the image tagging and word recognition tasks~\cite{chen14}.
\citet{jaderberg15} shows how to directly combine the structured SVM (SSVM)~\cite{taskar03,tsochantaridis05} objective with the procedure of training a CNN for text recognition.
CNNs with structured prediction have been recently explored for the task of human pose estimation.
\citet{chen2014pose} propose a model with data-dependent pairwise potentials but the different parts of the model were trained separately.
\citet{tompson14} construct a specific NN that mimicked the behaviour of several rounds of a message-passing inference algorithm.
Our Pairwise model is trained with an explicit structured-output surrogate loss with an external inference routine inside and enables to fine-tune all the parameters of the model jointly.


\section{Context-aware CNN model}
\label{sec:model}
This section presents main components of our contextual CNN model.
In Section~\ref{sec:localmodel}, we describe our Local model building on R-CNN~\cite{girshick14}.
In Section~\ref{sec:globalmodel}, we introduce the Global CNN model trained to score object proposals using the context of the full image.
Section~\ref{sec:pairwisemodel} describes our extension of CNNs with a structured-output loss function aimed to model pairwise relations between objects.\\[-0.5cm]

\subsection{Local model}
\label{sec:localmodel}
Our Local model follows R-CNN~\cite{girshick14} and uses selective search proposals~\cite{Vandeaande11} to restrict the set of object hypotheses. We extend the bounding box of each proposal with a small margin to capture local image context around objects. The image patch corresponding to each proposal is then resized to fit the input layer of the CNN. As we are interested in head detection, we select bounding boxes with square-like aspect ratios $\mathcal R\in [2/3, 3/2]$ and refer to them as candidates.


The R-CNN model is based on the AlexNet architecture~\cite{krizhevsky12} pre-trained on the ImageNet dataset~\cite{deng2009imagenet}.
We have considered several alternatives including VGG-S~\cite{chatfield2014}, VGG-verydeep-16~\cite{simonyan14} and~\citet{Oquab14}.
In our experiments VGG-S slightly outperformed AlexNet but was significantly slower in both training and testing.
\mbox{VGG-verydeep-16} showed better performance but was much slower.
The network of~\citet{Oquab14} had  better accuracy and similar speed compared to AlexNet (see Section~\ref{sec:exp:localArch} for details).
For experiments in this paper we use the pre-trained network of~\citet{Oquab14} extended by one fully-connected
layer (with 2048 nodes) initialized randomly and followed by ReLu and DropOut.

To train the network, we optimize parameters by minimizing the sum of independent log-losses using stochastic gradient descent (SGD) with momentum.
Differently from R-CNN which deploys the second pass of training using SVM, we use the outputs of CNN to score candidates. We found this training procedure to work better for our problem compared to the standard R-CNN training. More details on our training procedure can be found in Appendix~\ref{sec:app:local}.\\[-0.1cm]

\subsection{Global model}
\label{sec:globalmodel}
Our Global model uses image-level information to reason about locations of objects in the image.
The Global model is a CNN that takes the whole image as input and outputs a score for each cell of a multi-scale heat map.
The input image is isotropically rescaled and zero-padded to fit the standard CNN input of $224 \times 224$ pixels.
The output of the network is defined as a multi-scale grid of scores, corresponding to object hypotheses with coarsely discretized locations and scales in the image (see Figure~\ref{fig:globalModel_examples}).
Object hypotheses form a grid of $\textsc{c}=284$ square cells of four sizes (28x28, 56x56, 112x112 and 224x224 pixels) and the stride corresponding to the $50\%$ of cell size.
Except the output layer, the architecture of the Global CNN is identical to our Local model described in Section~\ref{sec:localmodel}.

The Global CNN is trained with SGD, minimizing the sum of~$\textsc{c}$ log-loss functions, one per each grid cell~$c\in\{1\cdots\textsc{c}\}$,
\begin{equation}
   \ell(\:\vec{f}_c(\mathbf{x})\:,\:y_c\:)
     = \!\!\sum_{y\in\{0,1\}}{\log(1+\exp{( (-1)^{y_c+y+1} \, f_{c,\,y}(\mathbf{x}))})} ~,
\end{equation}
where $\vec{f}_c(\mathbf{x}) \in \mathbb{R}^2$ is the output of the network for grid cell $c$ of input image $\mathbf{x}$; $y_c\in \{0,1\}$ is the label indicating the class of the grid cell $c$: \textit{background} or \textit{head}. 
We set the label of a grid cell to~\textit{head} if the Intersection-over-Union (IoU) overlap-ratio between the cell and any ground-truth bounding box in the image $\mathbf{x}$ is larger than~$0.3$, otherwise the label is set to~\textit{background}.

Due to the coarse resolution of grid cells, our Global model does not provide accurate localization. We therefore use the Global model to rescore the candidates of Local and Pairwise models. For this purpose, we match each candidate with the corresponding grid cell and compute affine combination of their scores. Each candidate is matched to a grid cell with the maximum IoU overlap-ratio. The parameters of affine score combination are optimized by cross validation on the validation set.

\begin{figure*}
\begin{center}
\includegraphics[trim = 0mm 5mm 0mm 10mm, clip, width=0.95\textwidth]{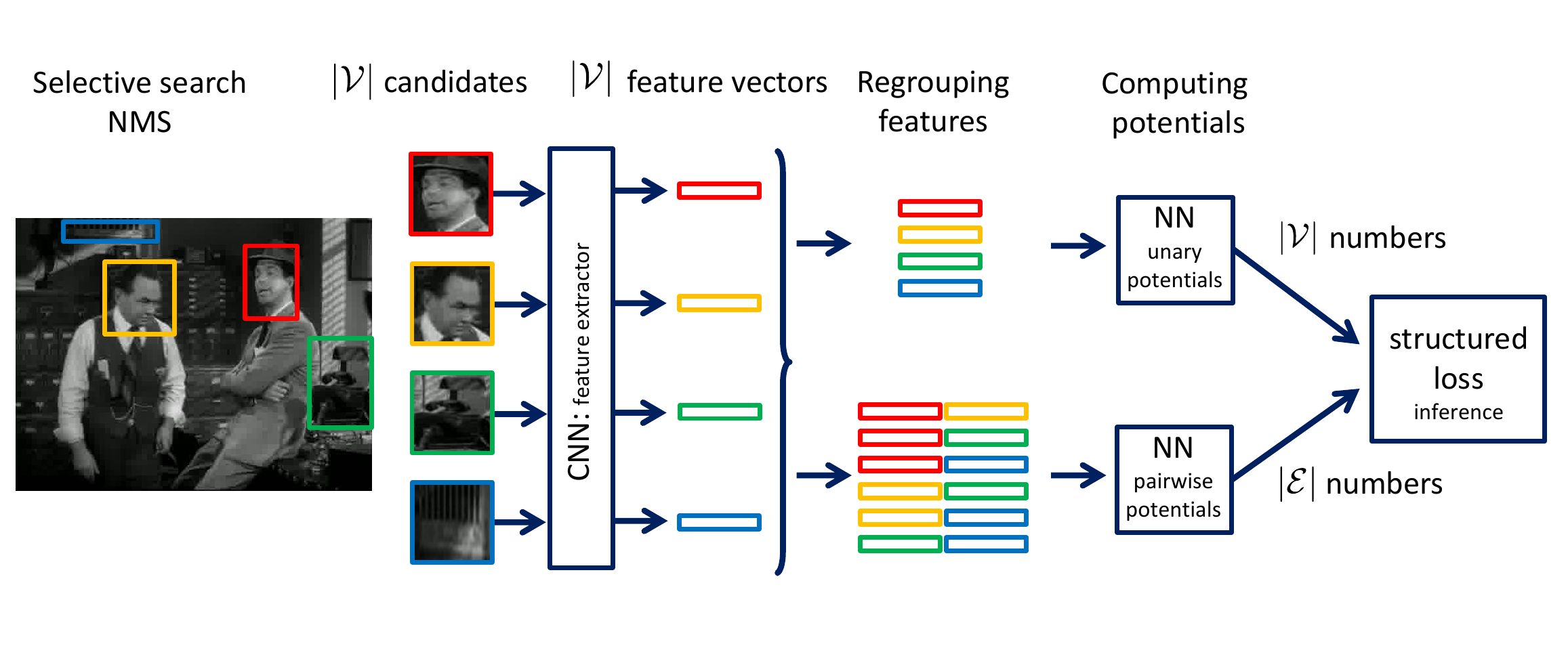}
\end{center}
\vspace{-0.8cm}
\caption{Pairwise model for the object detection task.}
\vspace{-0.3cm}
\label{fig:structuredNetworkScheme}
\end{figure*}

\subsection{Pairwise model}
\label{sec:pairwisemodel}
In this section we describe our Pairwise model that aims to jointly reason about multiple object candidates.
Following~\citet{desai11} we formulate the model as a joint score function where variables correspond to object candidates.
In the prior work~\cite{desai11,felzenszwalb10dpm,hoai14} unary potentials of the score function are defined by the response of the local object detector at corresponding locations, whereas higher-order potentials model spatial relations between candidates. Our Pairwise model enriches the model of~\cite{desai11} by making all potentials of the score function~\eqref{eq:pairwiseEnergyFull} dependent on the image data and, in contrast to~\cite{chen2014pose}, allows to perform the joint training of all parameters. We describe details of our model in Section~\ref{sec:pairwiseModelDetails}.

We train parameters of our model by minimizing the structured surrogate loss using stochastic gradient descent algorithm.
The details of our training procedure are presented in Section~\ref{sec:trainingPairwise}.

\subsubsection{Model formulation \label{sec:pairwiseModelDetails}}
\paragraph{Score function.}
Consider a set of~$\cV$ candidate bounding boxes (nodes) extracted from an image. 
Let each bounding box have a binary variable~$y_i$, $i\in\cV$ assigned to it. We associate label~$1$ with the object class and label~$0$ with the background class.
We assume that the ground-truth labels~$\hat{y}_i$ are available for all candidates in training images.

For each pair of nodes we choose an order based on the coordinates of corresponding bounding boxes: the left box is defined to be the first, the right one~-- the second.
Let~$\cE$ denote the set of oriented pairs of candidates (set of edges).
We cluster all edges based on relative locations and scales of bounding boxes\footnote{To cluster edges we apply k-means algorithm with $K = 20$ to a subset of oriented edges in training images. Edges in this subset connect object candidates with positive labels as well as any other candidates with high scores of the pre-trained Local model.
For the clustering we use relative location features (horizontal and vertical displacements, ratio of sizes) converted to the log scale and normalized to have zero mean and unit standard deviation. Further details of the clustering are available in Appendix~\ref{sec:app:pairwise}.} and denote the cluster index of edge~$(i,j) \in \cE$ by~$k_{ij} \in \{1,\dots,K\}$.

Inspired by~\citet{desai11}, we construct a joint score function~$S(\vec{y}; \vec{w})$ that ties together the labels of candidates in the same image:
\begin{equation}
\label{eq:pairwiseEnergyFull}
S(\vec{y}; \vec{w}) =\!
\sum_{i \in \cV} \theta^U_{i}(y_{i}; \vec{w})+\!\!\!\!\sum_{(i,j) \in \cE} \!\!\theta^P_{ij}(y_i, y_j, k_{ij}; \vec{w}),
\end{equation}
where $\vec{w}$ denotes trainable parameters, $\theta^U_{i}$ and $\theta^P_{ij}$ are unary and pairwise potentials depending on~$\vec{w}$,
 and $\vec{y} = (y_i)_{i \in \cV}$~ is a vector of all binary variables.

Note, that different values of potentials in~\eqref{eq:pairwiseEnergyFull} can lead to exactly the same score function~$S$.
We rewrite Eq.~\eqref{eq:pairwiseEnergyFull} in the more compact form (the set of all representable functions of binary variables stays the same):
\begin{equation}
\label{eq:pairwiseEnergyCompact}
S(\vec{y}; \vec{w}) =
\sum_{i \in \cV} y_{i} \: \theta^U_{i}\!(\vec{w}) + \sum_{(i,j) \in \cE} y_i y_j \: \theta^P_{ij, k_{ij}}\!(\vec{w})
\end{equation}
where unary potentials~$\theta^U_{i}$ and pairwise potentials~$\theta^P_{ij, k_{ij}}$ are represented by real values.

\paragraph{Connecting the score function and the image.}
Now we connect the image with potentials of the score function~\eqref{eq:pairwiseEnergyCompact} using several feed-forward neural networks.
First, from the Local model described in Section~\ref{sec:localmodel} we create a feature extractor (FE), i.e.\ a function~$\phi^E$ that constructs feature vector~$\vec{f}_i$ for the image data~$\vec{x}_i$ of candidate~$i$: $\vec{f}_i=\phi^E(\vec{x}_i, \vec{w}^E)$. Here~$\vec{w}^E$ is a vector of trainable parameters of FE.

To connect features~$\vec{f}_i$ with potentials in~\eqref{eq:pairwiseEnergyCompact} we construct two additional feed-forward networks: the unary network (UN) and the pairwise network (PN).
The unary network~$\phi^U$ maps the feature vector~$\vec{f}_i$ of a candidate~$i$ to the value of the corresponding unary potential, i.e.\ $\theta^U_{i} = \phi^U(\vec{f}_i, \vec{w}^U  )$. The pairwise network~$\varphi^P$ maps the concatenated feature vectors of its two candidates to a vector~$\vec{\theta}^P_{ij}$ where the $k$-th component~$\theta^P_{ij, k}$ corresponds to the one of~$K$ cluster indices, i.e.\ $\theta^P_{ij, k_{ij}} = \phi^P_{k_{ij}}(\vec{f}_i, \vec{f}_j, \vec{w}^P )$. Vectors~$\vec{w}^U$ and~$\vec{w}^P$ are the trainable parameters of the UN and PN, correspondingly.

In our experiments we found the following architectures to work best.
The FE was of the same structure as our Local model (based on the network of~\citet{Oquab14}) leading to~$2048$ features.
In both UN and PN we use just one fully-connected layer. Addition of more hidden layers did not improve results.

\paragraph{Precision-recall evaluation.}
Object detection methods are typically evaluated in terms of precision-recall (PR) and average precision (AP) values.
To construct the precision-recall curve given the joint score~\eqref{eq:pairwiseEnergyCompact}, we follow the approach of~\citet{desai11}.
For each candidate bounding box~$i$, we compute an individual score~$s_i(\vec{w})$ defined as the difference of the max-marginals of the joint score
\begin{equation}
\label{eq:maxMarginalDifference}
s_i(\vec{w}) = \max_{\vec{y}: y_i = 1} S(\vec{y}; \vec{w}) - \max_{\vec{y}: y_i = 0} S(\vec{y}; \vec{w}).
\end{equation}
The individual scores are used in the standard precision-recall evaluation pipeline~\cite{pascal2010voc}.

When the number of candidates is small, i.e.\ $|\cV| \leq 20$, both maximization problems of~\eqref{eq:maxMarginalDifference} can be solved exactly using exhaustive search.
When the number of candidates becomes larger, the exhaustive search becomes too slow. In this case one can use the cascade of QPBO~\cite{Kolmogorov07qpbo} and TRW-S~\cite{Kolmogorov06trws} methods to approximate~$s_i$. Specifically, QPBO allows to quickly determine the optimal label for some candidates. On our dataset QPBO works surprisingly well, i.e.,\ in many cases it is able to label all nodes. If some nodes are unlabeled by QPBO, one can apply the exhaustive search when the number of unlabelled nodes is at most 20 and TRW-S otherwise.

We have tried using 16 and 32 candidates per image. The exact inference is tractable only in the first case. In this paper we use 16 candidates per image as the large number of candidates did not improve performance on our validation set.

\subsubsection{Training the model \label{sec:trainingPairwise}}
We train parameters of our model by minimizing a structured surrogate loss using the stochastic gradient descent algorithm\footnote{As common in the deep learning literature we ignore the non-differentiability issues and assume that in practice we can always compute the gradient.}\!\!.
The algorithm for parameter update consists of the following four steps:
\begin{enumerate}[noitemsep,nolistsep]
\item\label{step:nms}Select the set of candidates by applying the non-maximum suppression~\cite{felzenszwalb10dpm} on top of the scores produced by the Local model.
\item\label{step:forward} Perform the forward pass through the model to compute potentials of the joint score function.
\item\label{step:inference} Perform the inference to compute the structured loss and its gradient (see below).
\item\label{step:backProp} Back-propagate the gradient through the model.
\end{enumerate}
We explain details of the algorithm below.

\paragraph{Structured surrogate loss.} A structured loss is a function that maps the current values of parameters, image data $\vec{x} = (\vec{x}_i)_{i \in \cV}$ and the ground-truth labeling~$\hat{\vec{y}} = (\hat{y}_i)_{i \in \cV}$ to a real number.
A popular choice for the surrogate loss for structured-prediction tasks is the structured SVM (SSVM) objective~\cite{taskar03,tsochantaridis05}:
\begin{equation}
\label{eq:ssvmLoss}
\ell_{\text{SVM}}( \vec{w}, \hat{\vec{y}}, \vec{x} ) = \max_{\vec{y}} \Bigl( S(\vec{y}; \vec{w}, \vec{x}) + h(\vec{y}, \hat{\vec{y}}) \Bigr) - S(\hat{\vec{y}}; \vec{w}, \vec{x})
\end{equation}
where $h(\vec{y}, \hat{\vec{y}}) \geq 0$ measures the agreement between the two labelings. Possible choices for~$h$ include the Hamming loss, the Hamming loss with penalties normalized by the frequency of classes, or higher-order losses making use of assumption that each ground-truth object is assigned to exactly one object candidate~\cite{osokin14}.
Notice, that in~\eqref{eq:ssvmLoss} the joint score~$S$ depends on parameters~$\vec{w}$ and image data~$\vec{x}$ implicitly through potentials $\theta^U$ and $\theta^P$.

However, in our experiments we have observed that the SSVM loss is less suited for the detection task, i.e.\ optimizing the objective~\eqref{eq:ssvmLoss} does not lead to good results in terms of precision-recall measure.
To tackle this problem, we propose a new surrogate loss which directly imposes penalties on the wrong values of individual scores~\eqref{eq:maxMarginalDifference} extracted from the joint score~$S$.
Specifically, this loss can be written as
\begin{equation}
\label{eq:scoreLoss}
\ell( \vec{w}, \hat{\vec{y}}, \vec{x} ) = \sum_{i: \hat{y}_i = 1} v( s_i(\vec{w}, \vec{x}) ) + \sum_{i: \hat{y}_i = 0} v( -s_i(\vec{w}, \vec{x}) )
\end{equation}
where $v$ can be any non-increasing function bounded from below. We use $v(t) = \log(1 + \exp(-t))$ which brings us closer to the training of conventional detector with a soft-max loss.

\paragraph{Gradient of the structured loss.}
To optimize the structured loss w.r.t.\ the model parameters~$\vec{w}$, we need to compute the gradient of the objective w.r.t.\ model parameters. We can always achieve this goal using the back-propagation method under two assumptions: 1) the gradient can be back-propagated through the modules of the model, i.e.\ all the partial derivatives of $\phi^E$, $\phi^U$, $\phi^P$ w.r.t.\ the input and the parameters can be computed; 2) the scores of the candidates~\eqref{eq:maxMarginalDifference} can be computed exactly.

To start the back-propagation procedure, we compute the gradient of structured loss w.r.t.\ potentials $\theta^U_{i}$, $\theta^P_{ij, k}$ of the joint score function~$S$.
\citet{jaderberg15} have in details explained how to do this for the SSVM loss~\eqref{eq:ssvmLoss}. Here we explain how to differentiate the loss~\eqref{eq:scoreLoss}.
First, the gradient of the loss~\eqref{eq:scoreLoss} w.r.t.\ the scores can be expressed as
$$
\frac{d\ell}{d s_i} = (-1)^{\hat{y}_i+1} v'( s_i (-1)^{\hat{y}_i+1} ), \; v'(t) = \frac{-\exp(-t)}{1 + \exp(-t)}.
$$
The gradient of the score (when existent) w.r.t.\ potentials can be computed exactly if we can compute all max-marginals exactly:
\begin{equation*}
\frac{d s_i}{d \theta^U_p } = y_p^{i,1} - y_p^{i,0}, \;\; \frac{d s_i}{d \theta^P_{pq, k} } = (y_p^{i,1} y_q^{i,1} - y_p^{i,0} y_q^{i,0}) [k_{ij} = k]
\end{equation*}
where $y_q^{i,t}$ is the $q$-th component of $\vec{y}^{i,t} = \argmax\limits_{\vec{y}: y_i = t} S(\vec{y}; \vec{w})$ for $t\in\{0,1\}$. Here, $[ \cdot ]$ is the Iverson bracket notation.
Combining the two derivatives via the chain rule we get
\begin{equation*}
\frac{d\ell}{d \theta^U_p } = \sum_{i \in \cV} \frac{d\ell}{d s_i} \frac{d s_i}{d \theta^U_p }, \;\; \frac{d\ell}{d \theta^P_{pq, k} } = \sum_{i \in \cV} \frac{d\ell}{d s_i} \frac{d s_i}{d \theta^P_{pq, k} }.
\end{equation*}

\paragraph{Back-propagation of the gradient.}
The next step of the back-propagation procedure is to compute the derivatives of the loss w.r.t.\ parameters of the UN and PN
\begin{equation}
\label{eq:dLdWup}
\frac{d\ell}{d \vec{w}^U} \!\!=\!\!\sum_{i \in \cV} \!\frac{d\ell}{d \theta^U_i } \frac{d \theta^U_i }{d \vec{w}^U}, \,
\frac{d\ell}{d \vec{w}^P} \!\!=\!\!\!\!\!\sum_{(i,j) \in \cE} \sum_{k=1}^K \!\frac{d\ell}{d\theta^P_{ij, k} } \frac{d\theta^P_{ij, k}}{d \vec{w}^P}
\end{equation}
and w.r.t.\ the output of the feature extractor
\begin{align}
\notag
\frac{d\ell}{d\vec{f}_i} = \frac{d\ell}{d \theta^U_i } \frac{d\theta^U_i }{d\vec{f}_i}
&+
\sum_{j : (i,j) \in \cE} \frac{d\ell}{d\theta^P_{ij, k_{ij}} }\frac{d\theta^P_{ij, k_{ij}}}{d\vec{f}_i} \\
\label{eq:dLdF}
&+
\sum_{j : (j,i) \in \cE} \frac{d\ell}{d\theta^P_{ji, k_{ji}} }\frac{d\theta^P_{ji, k_{ji}}}{d\vec{f}_i}.
\end{align}
Notice that all the derivatives of potentials w.r.t.\ parameters and features can be computed by propagating the gradient through networks $\phi^U$ and $\phi^P$\!\!.
Finally, propagation of the gradient~\eqref{eq:dLdF} through $\phi^E$ gives us the direction  of the update for parameters~$\vec{w}^E$ of the FE.
\section{Datasets \label{sec:datasets}}
In this section we present our new head detection dataset, HollywoodHeads (HH), and discuss two other datasets we use for evaluation: TVHI~\cite{patron2012structured,hoai14} and Casablanca~\cite{ren2008finding}.

\subsection{HollywoodHeads dataset \label{sec:dataset}}

HollywoodHeads dataset contains $369,846$ human heads annotated in $224,740$ video frames from 21 Hollywood movies\footnote{\label{note:hh}List of movies used in HollywoodHeads dataset. Training set: \textit{American beauty},
\textit{As Good As It Gets},
\textit{Big Fish},
\textit{Big Lebowski},
\textit{Bringing out the dead},
\textit{Capote},
\textit{Clerks},
\textit{Crash},
\textit{Dead Poets Society},
\textit{Double Indemnity},
\textit{Erin Brockovich},
\textit{Fantastic 4},
\textit{Fargo},
\textit{Fear And Loathing In Las Vegas},
\textit{Fight Club}.
Validation set:
\textit{Five Easy Pieces},
\textit{Forrest Gump},
\textit{Gang Related}.
Test set:
\textit{Gandhi},
\textit{Charade},
\textit{I Am Sam}.}\!\!.
The movies vary in genres and represent different time epochs.
To create annotation, we have manually annotated tracks of human heads in action-rich movie clips.
For each head track, head bounding boxes, i.e.,~the smallest axis-parallel rectangles including all visible pixels of the head, were manually annotated on several key frames. The bounding boxes on remaining frames were linearly interpolated and manually verified to be correct.
In total, we have collected $2,380$ clips with $3,872$ human tracks, spanning over $3.5$ hours of video.
The dataset is divided into the training, validation and test subsets which have no overlap in terms of movies\footref{note:hh}\!\!. 
Given the redundancy of consequent video frames, we have temporally subsampled videos in the validation and test subsets.
In summary, the training set of HollywoodHeads contains $216,719$ frames from $15$ movies, the validation set contains $6,719$ frames from $3$ movies and the test set contains $1,302$ frames from another set of $3$ movies. Human heads with poor visibility (e.g.,~strong occlusions, low lighting conditions) were marked by the ``difficult'' flag and were excluded from the evaluation.
The HollywoodHeads dataset is available from~\cite{projectwebpage}.

\subsection{TVHI dataset}
The extended TV Human Interaction (TVHI) dataset~\cite{patron2012structured,hoai14} consists of $1,313$ frames of TV show episodes annotated with bounding boxes of human upper bodies. Frames are split into the two sets: $599$ for training and $714$ for testing.
To evaluate head detection using upper-body annotation, we have applied bounding-box regression to the output of head detectors~\cite{mathias2014face}.
The parameters of regression were tuned on the TVHI training subset for each tested method.


\subsection{Casablanca dataset}
\label{sec:casadataset}
The Casablanca dataset~\cite{ren2008finding} contains $1,466$ frames from the movie ``Casablanca''. The frames are annotated with head bounding boxes, however, the annotation of frontal heads is typically reduced to face bounding boxes and, therefore differs in the scale and aspect ratio from the HollywoodHeads annotation.
Given some mistakes in the original annotation of~\cite{ren2008finding},
we have added missing bounding boxes for heads of all people in the foreground. We have also applied bounding-box regression~\cite{mathias2014face} to compensate for differences in annotation policies.
\section{Experiments}
\label{sec:experiments}
This section presents our experimental results.
First, we demonstrate the effect of different combinations of proposed models (Section~\ref{sec:exp:modelVariants}) and provide the comparison with the state-of-the-art~(Section~\ref{sec:exp:rcnnComparison}).
Section~\ref{sec:exp:localArch} compares different architectures of the Local model.
We then justify the need of our new large dataset for training~(Section~\ref{sec:exp:trainingSize}) and show improvements in computational complexity that can be achieved with the Global model~(Section~\ref{sec:exp:speedUp}).

To evaluate the detection performance, we use the standard Average Precision (AP) measure based on the Precision-Recall (PR) curve~\cite{pascal2010voc}. Detections having high overlap ratio with the ground truth (IoU~\textgreater~0.5) are considered as true positives. Multiple detections assigned to the same ground truth are penalized and declared as false positives. Matches to ``difficult'' head annotations are ignored in the evaluation, i.e.\ such detections are considered neither as true positives nor as false positives.
%

\subsection{Results of context-aware models \label{sec:exp:modelVariants}}
\begin{figure}[!]
\begin{center}
\includegraphics[trim = 7.4cm 1.7cm 7.4cm 1.7cm, clip, width=\linewidth]{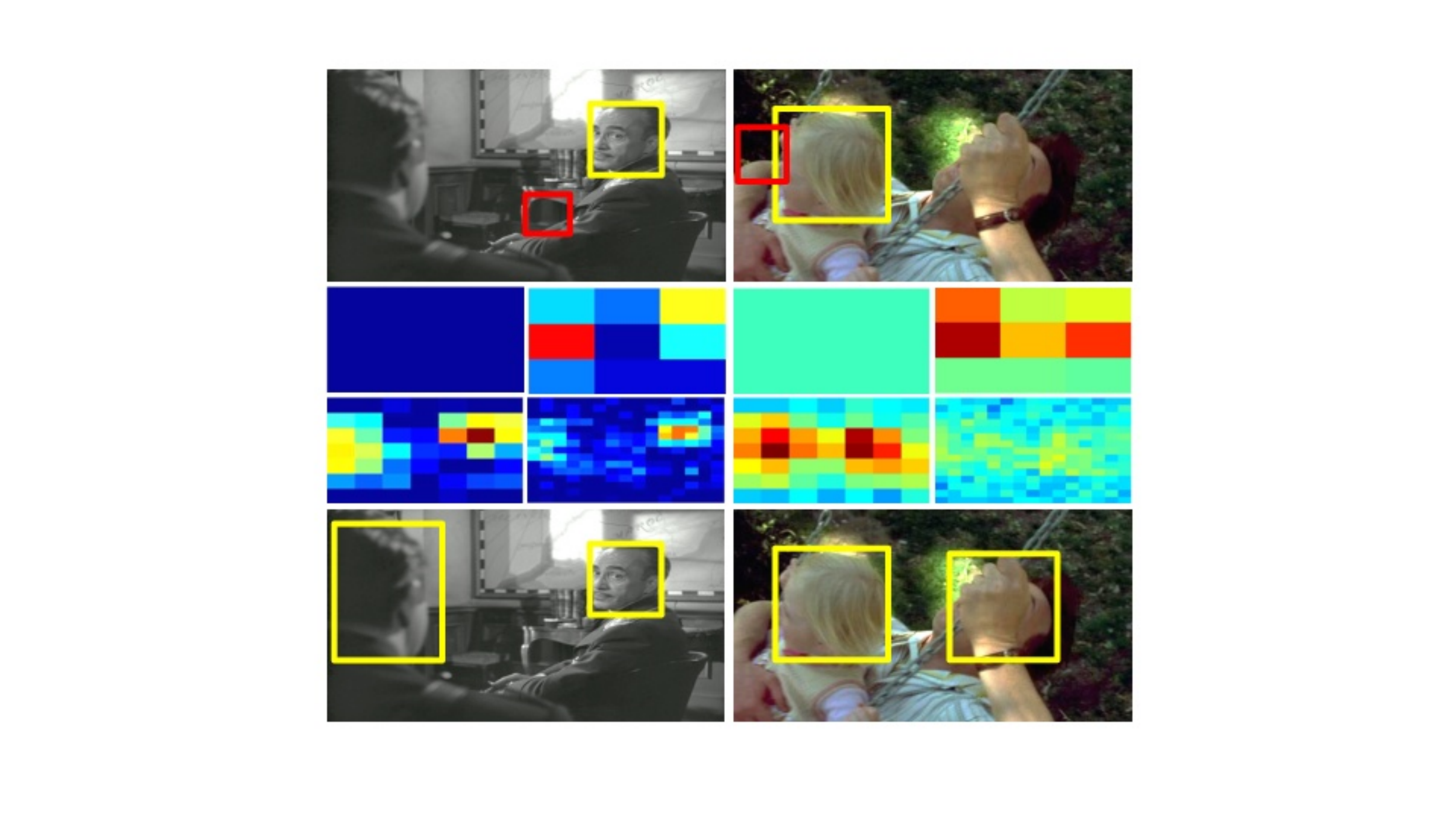}
\end{center}\vspace{-0.6cm}
\caption{Qualitative results for the Global model. The top row shows detections produced by the Local model. The middle row illustrates the multi-scale score map produced by our Global model.
Red color correspond to high score values for the ``head'' class, blue color~-- to low score values.
The bottom row demonstrates detections by the combination of the Local and Global models.
\label{fig:globalModel_examples}}
\vspace{-0.4cm}
\end{figure}

\begin{figure}[!h]
\begin{center}
\begin{tabular}{@{}c@{\,}c@{}}
\includegraphics[trim = 0mm 0mm 0mm 0mm, clip, width=0.48\linewidth]{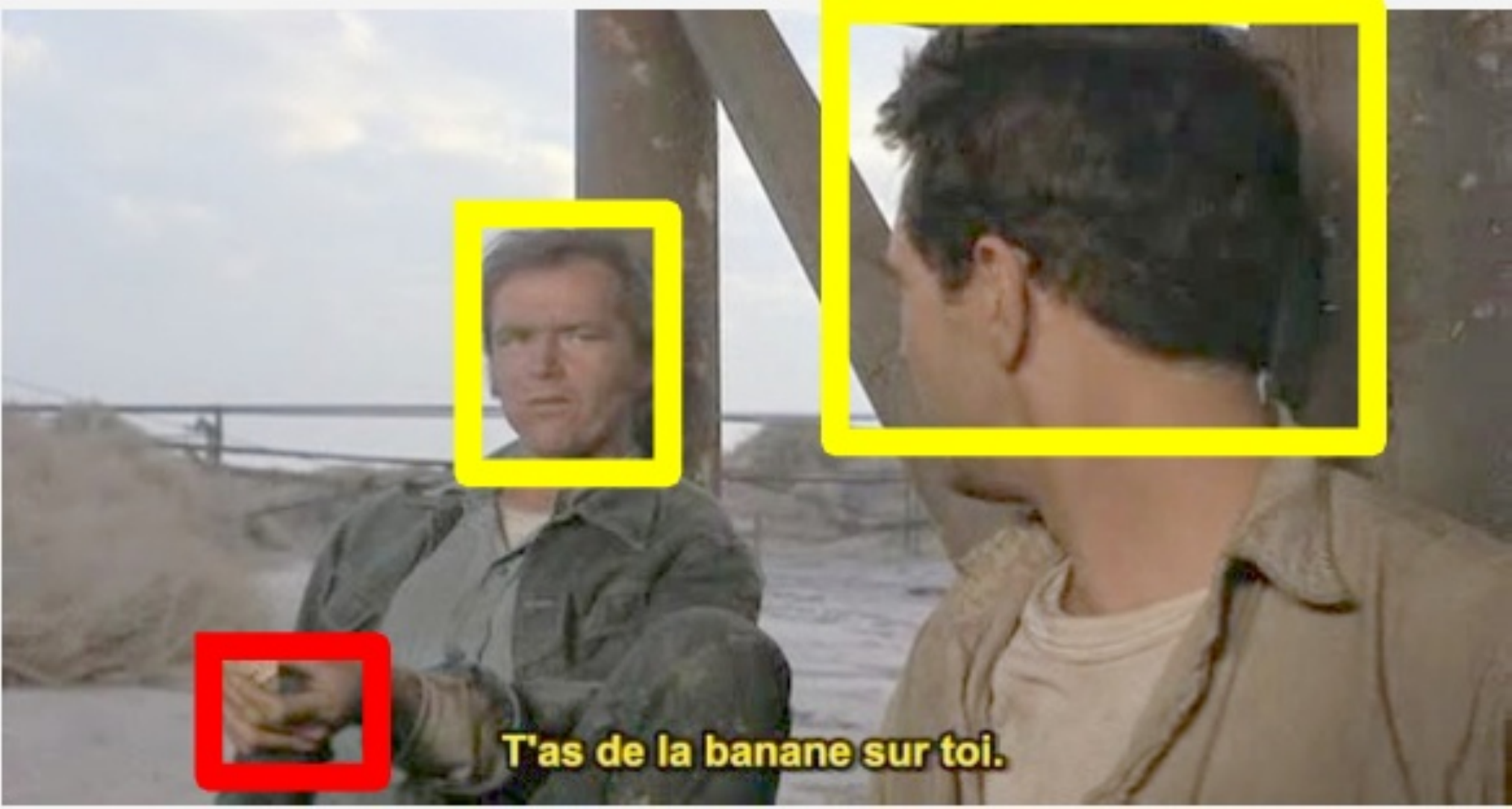}
&
\includegraphics[trim = 0mm 0mm 0mm 0mm, clip, width=0.48\linewidth]{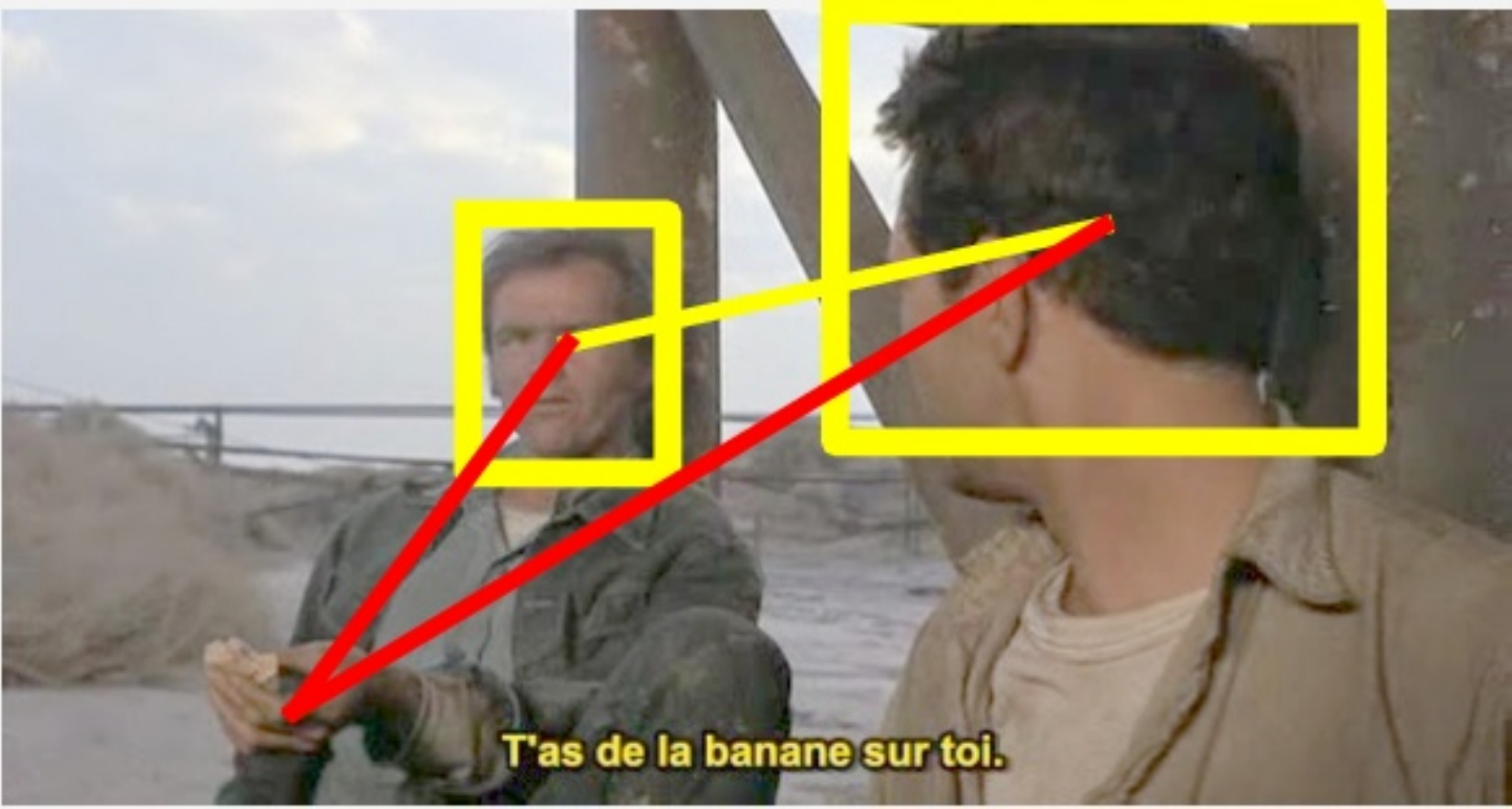} \\[-0.1cm]
\includegraphics[trim = 0mm 0mm 0mm 0mm, clip, width=0.48\linewidth]{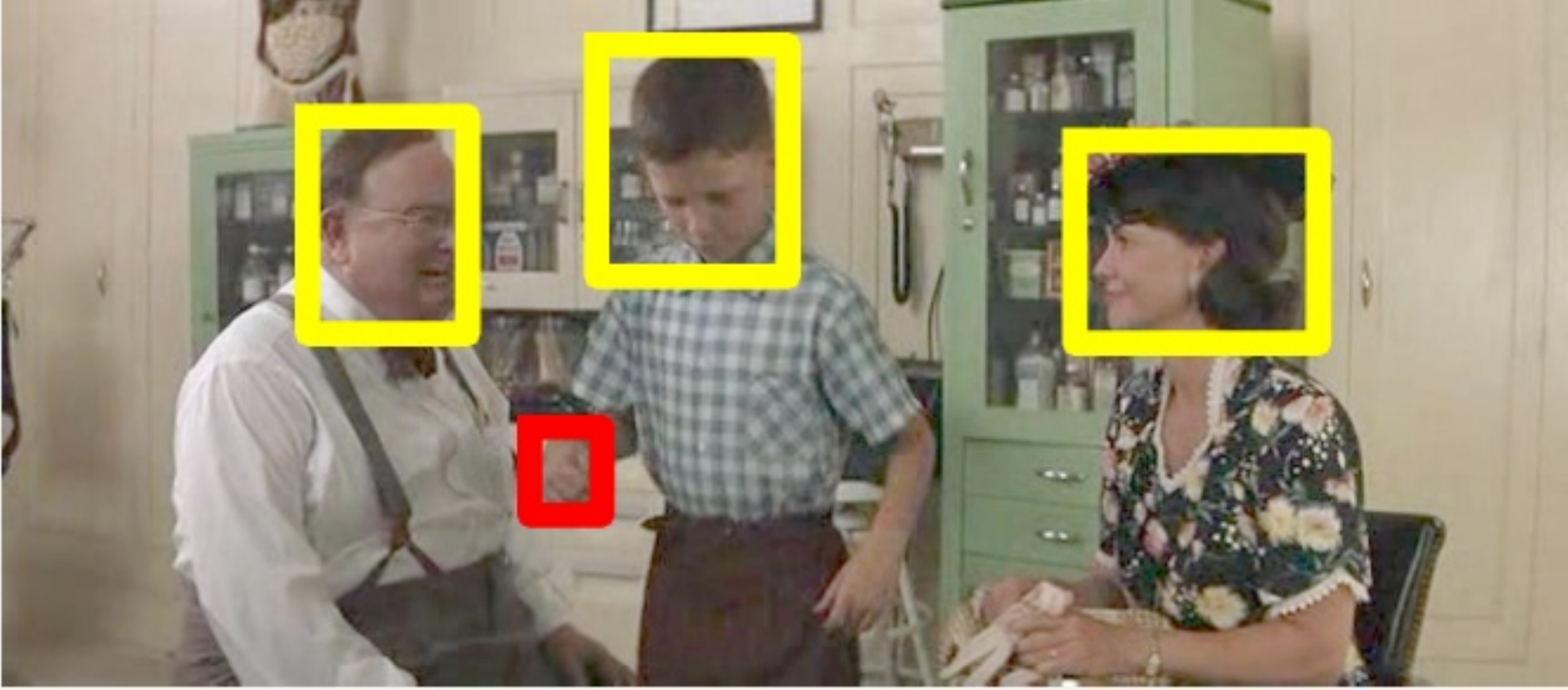}
&
\includegraphics[trim = 0mm 0mm 0mm 0mm, clip, width=0.48\linewidth]{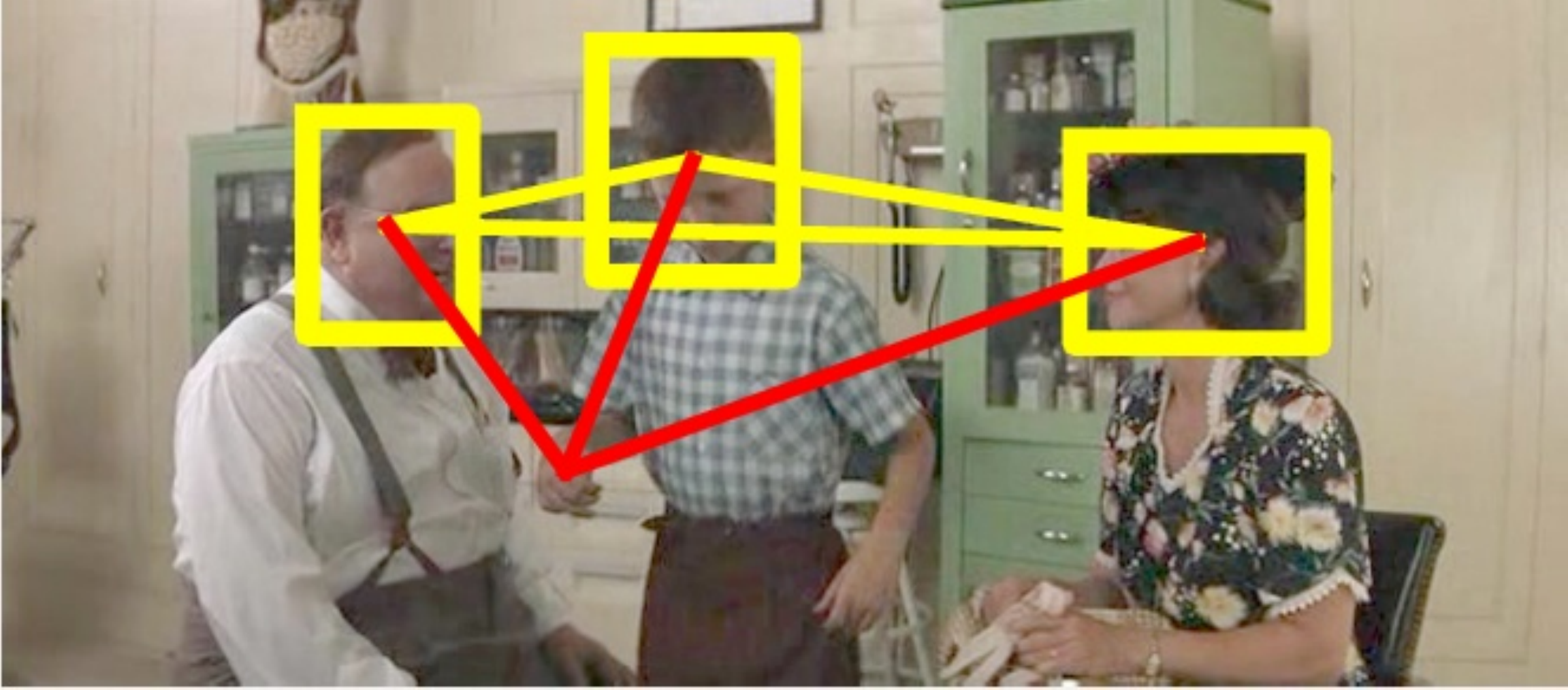} \\[-0.1cm]
\includegraphics[trim = 0mm 0mm 0mm 0mm, clip, width=0.48\linewidth]{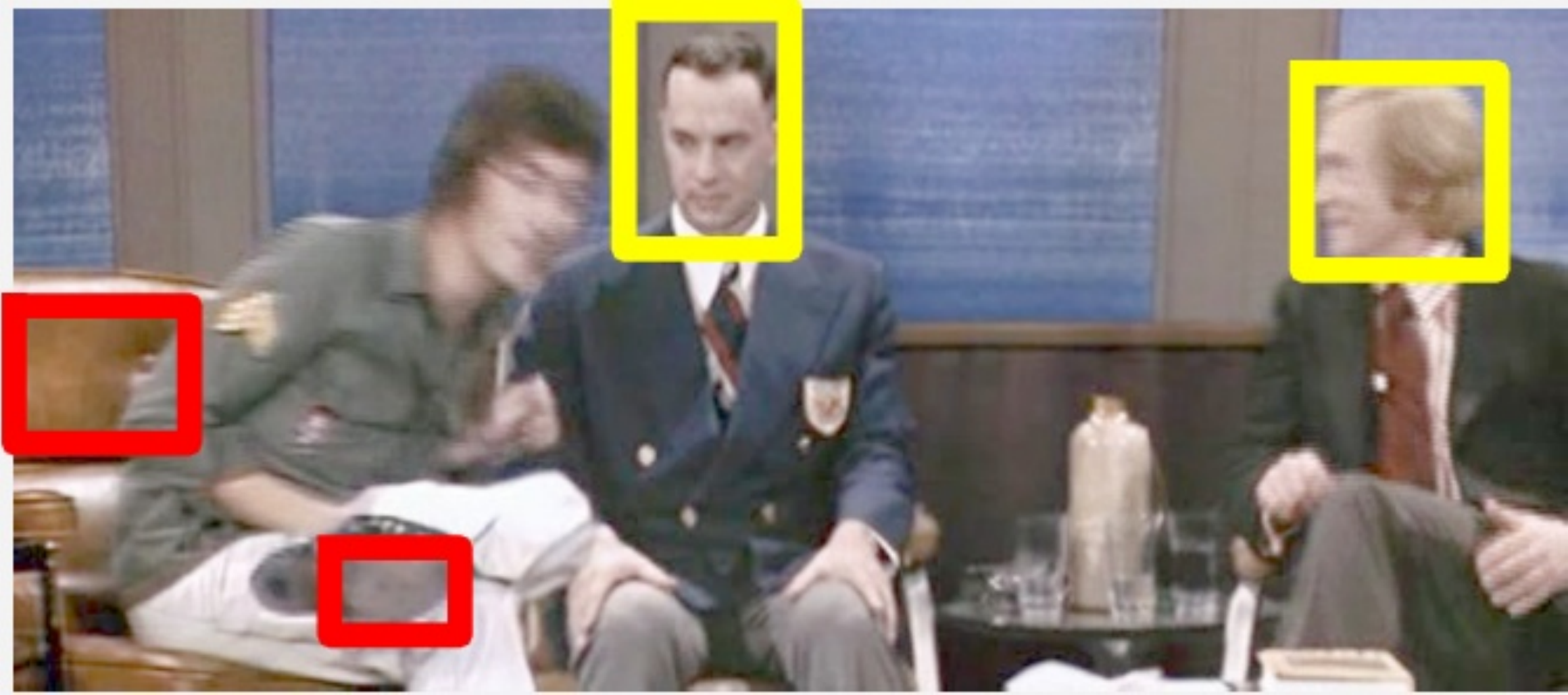}
&
\includegraphics[trim = 0mm 0mm 0mm 0mm, clip, width=0.48\linewidth]{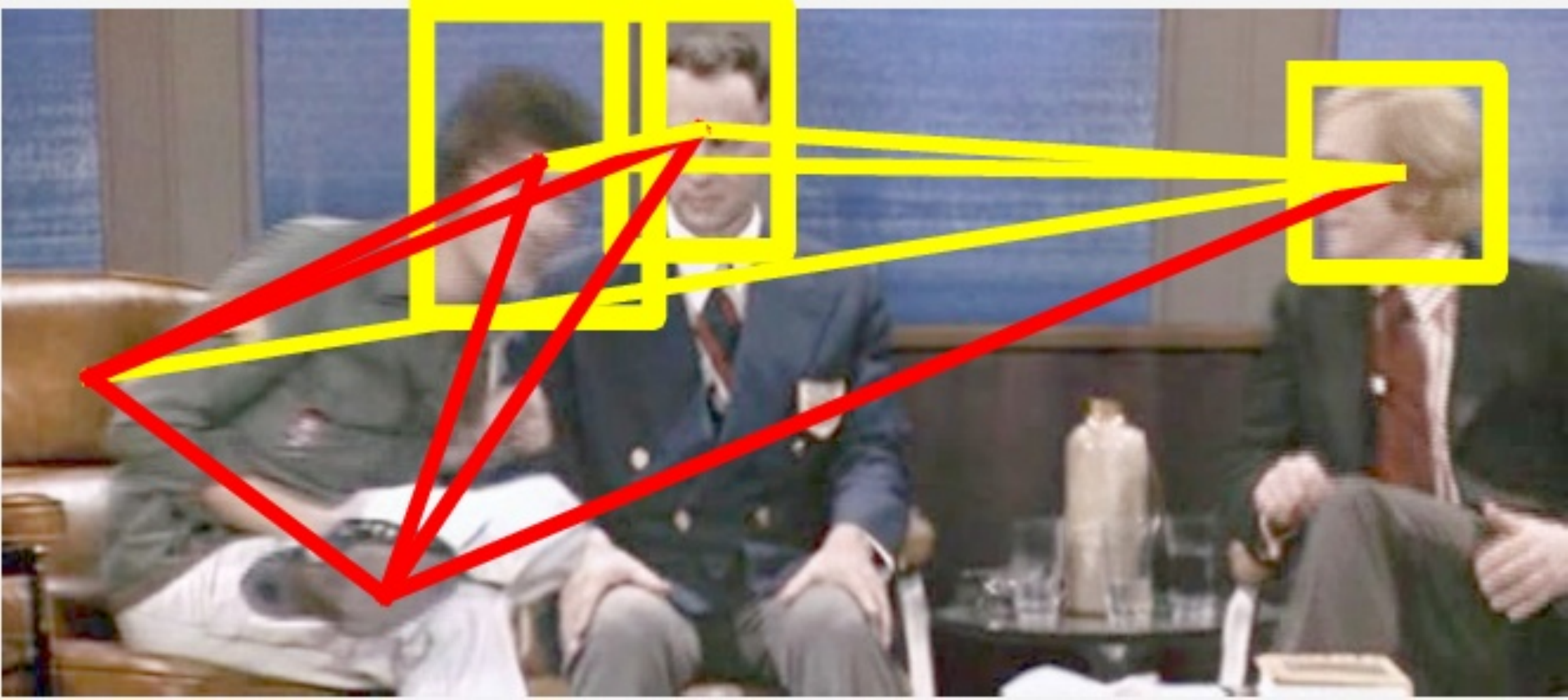} \\[-0.1cm]
\includegraphics[trim = 2mm 16mm 2mm 16mm, clip, width=0.48\linewidth]{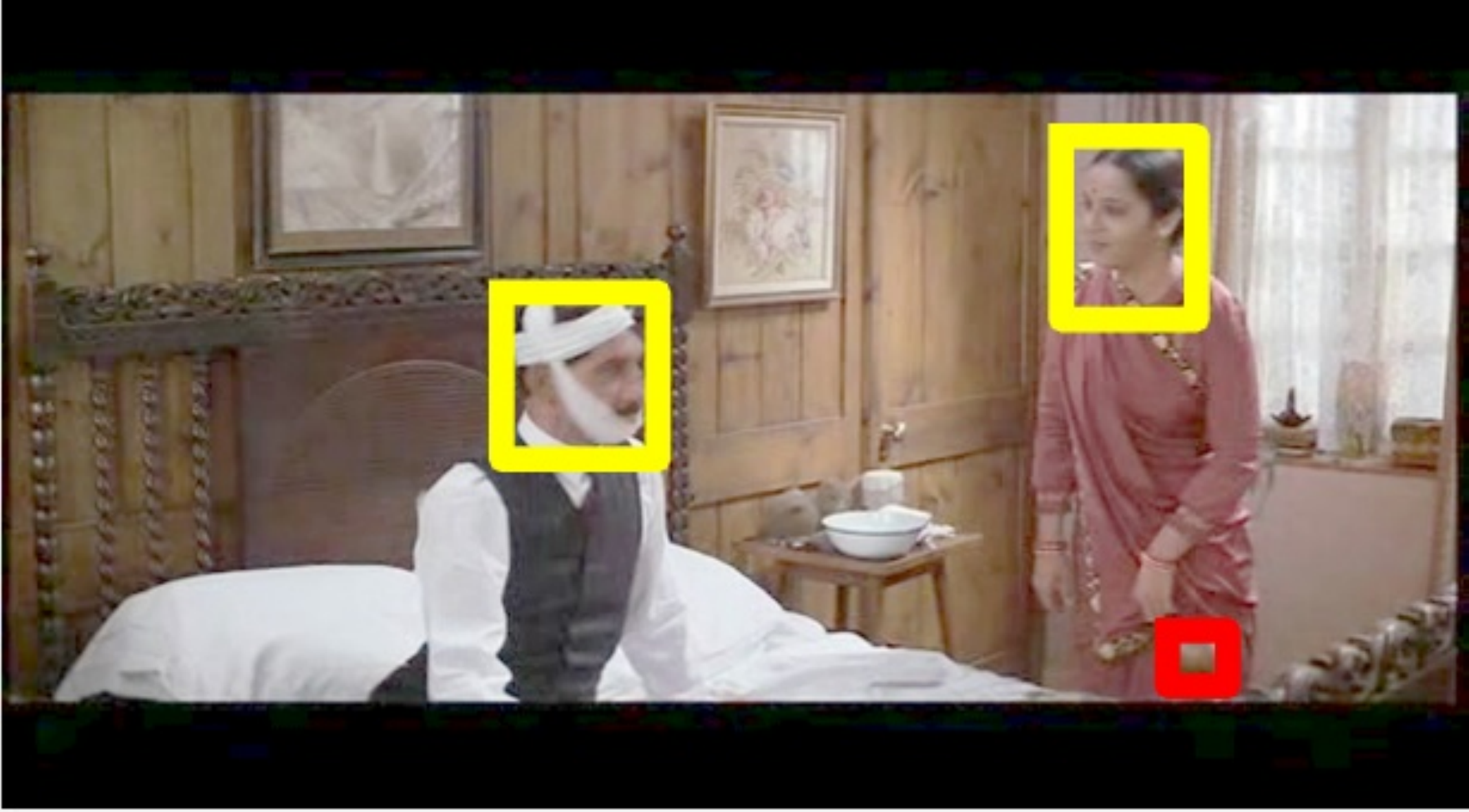}
&
\includegraphics[trim = 2mm 16mm 2mm 16mm, clip, width=0.48\linewidth]{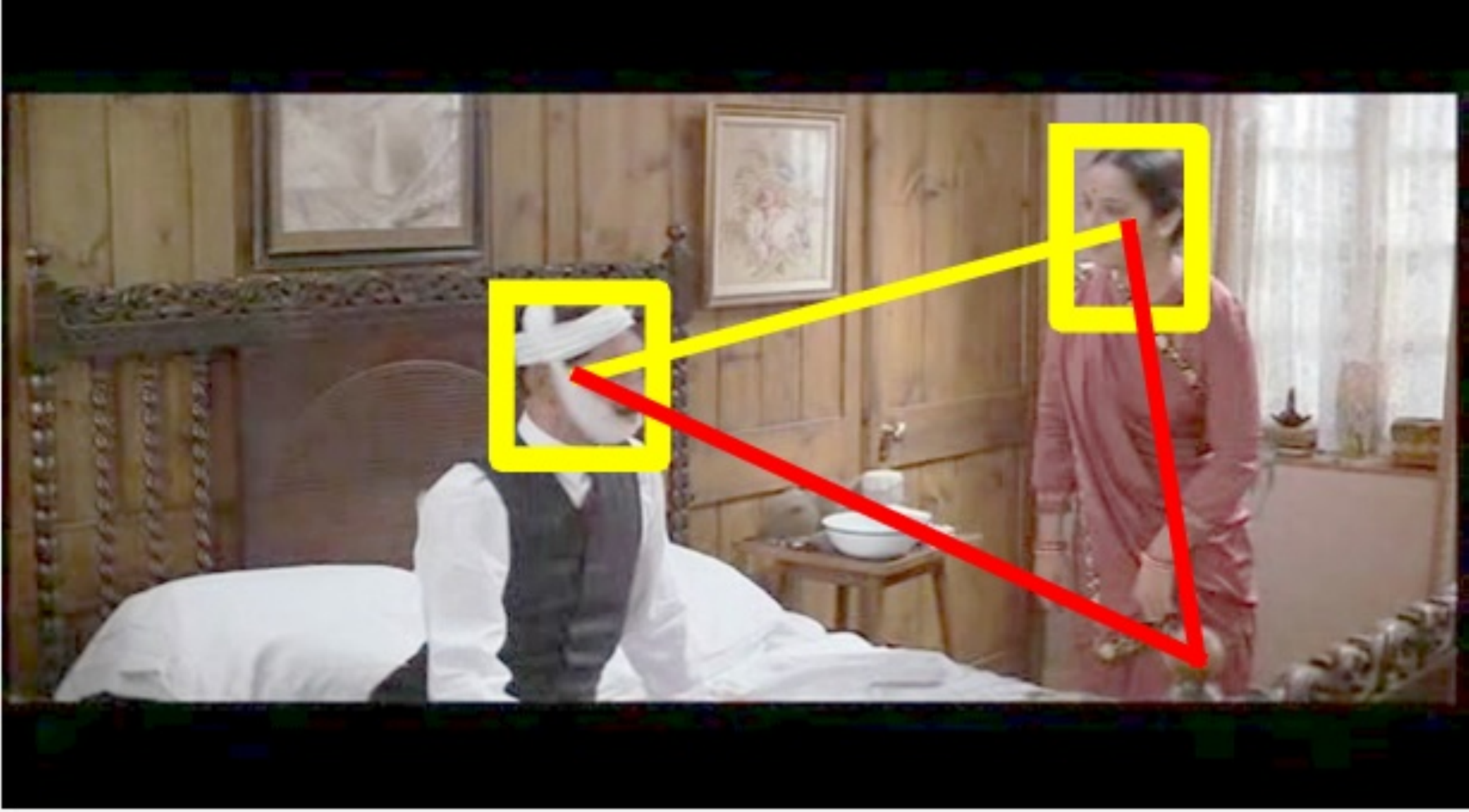} \\[-0.1cm]
\includegraphics[trim = 2mm 16mm 2mm 16mm, clip, width=0.48\linewidth]{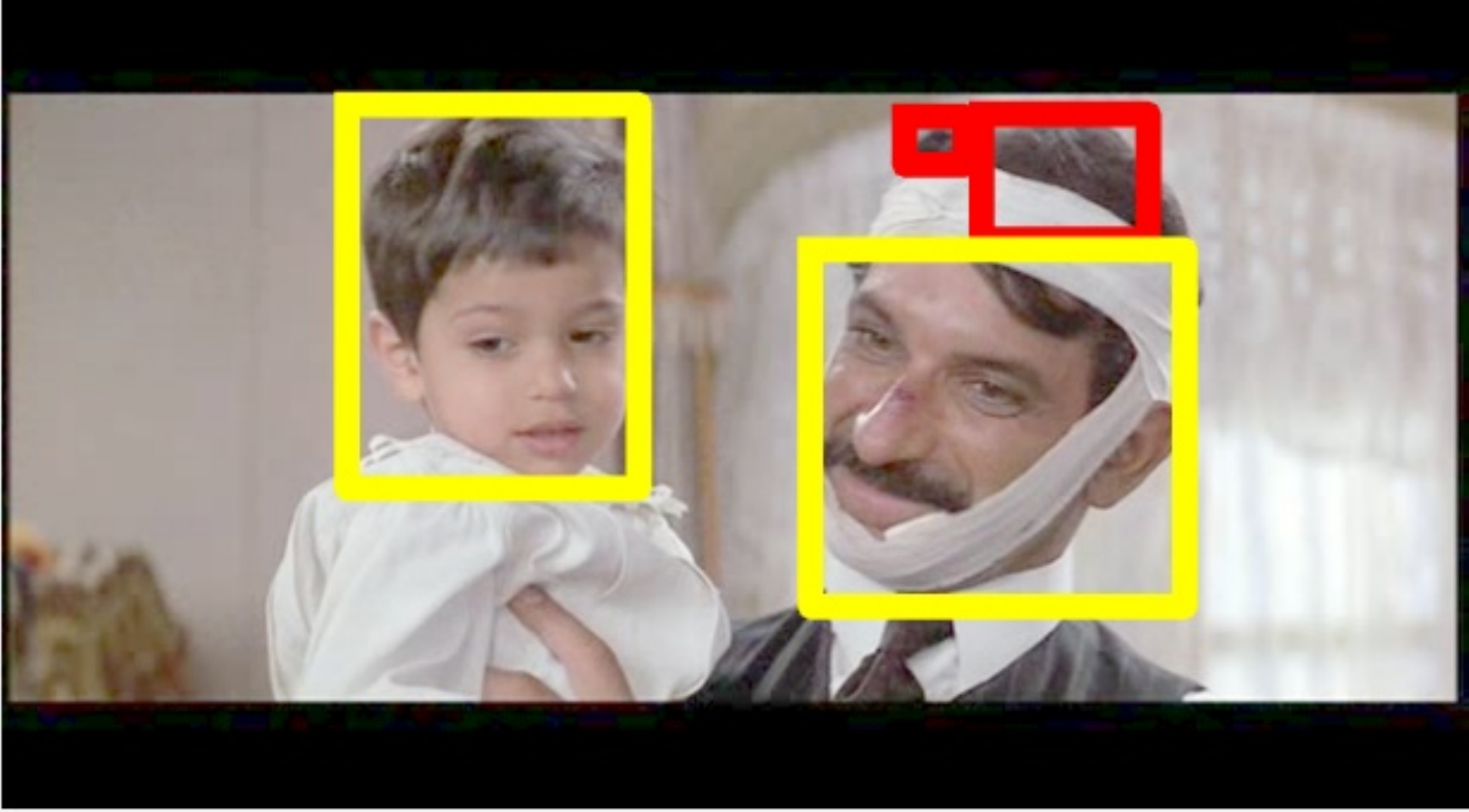}
&
\includegraphics[trim = 2mm 16mm 2mm 16mm, clip, width=0.48\linewidth]{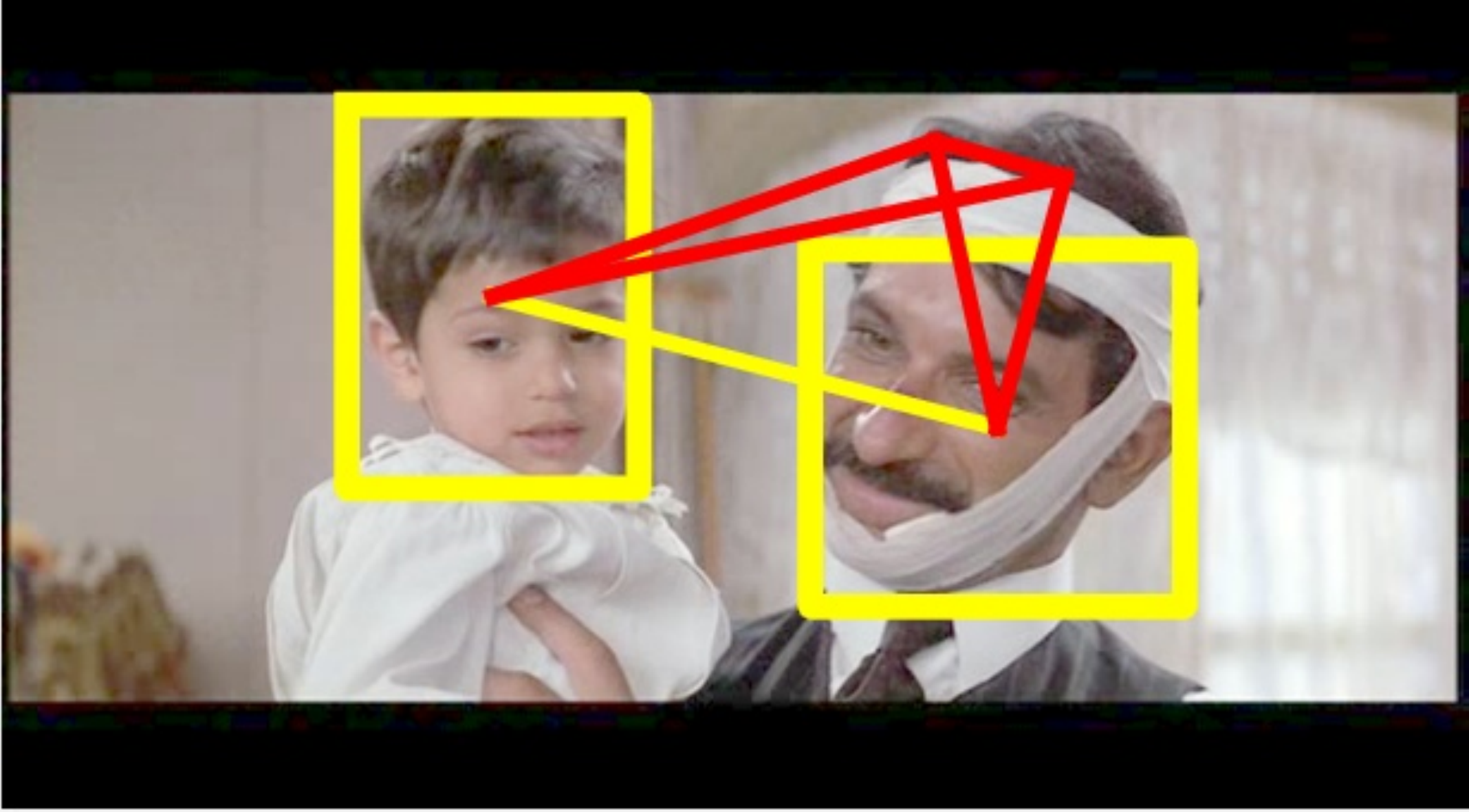} \\[-0.1cm]
\includegraphics[trim = 2mm 24mm 4mm 23mm, clip, width=0.48\linewidth]{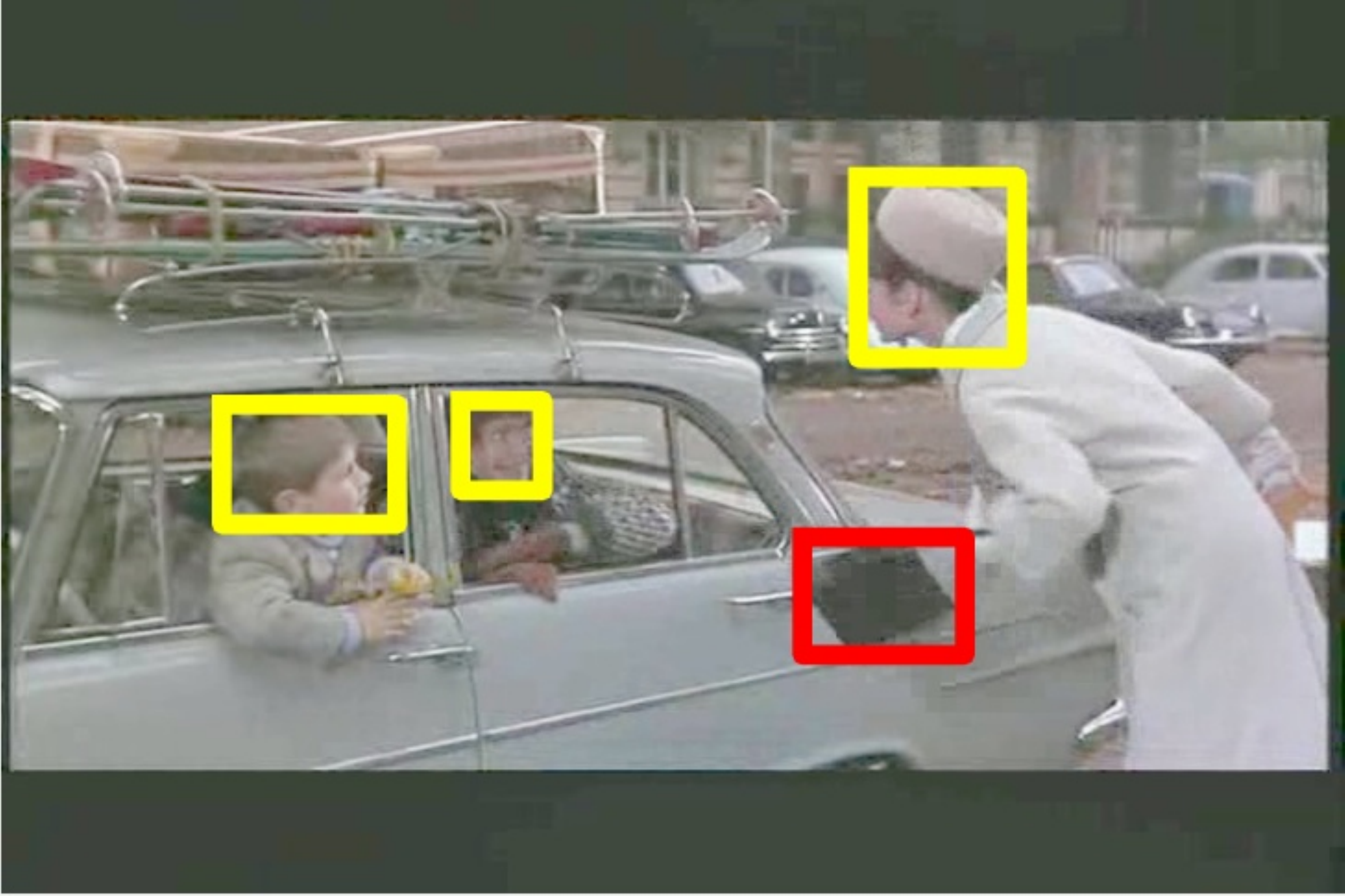}
&
\includegraphics[trim = 2mm 24mm 4mm 23mm, clip, width=0.48\linewidth]{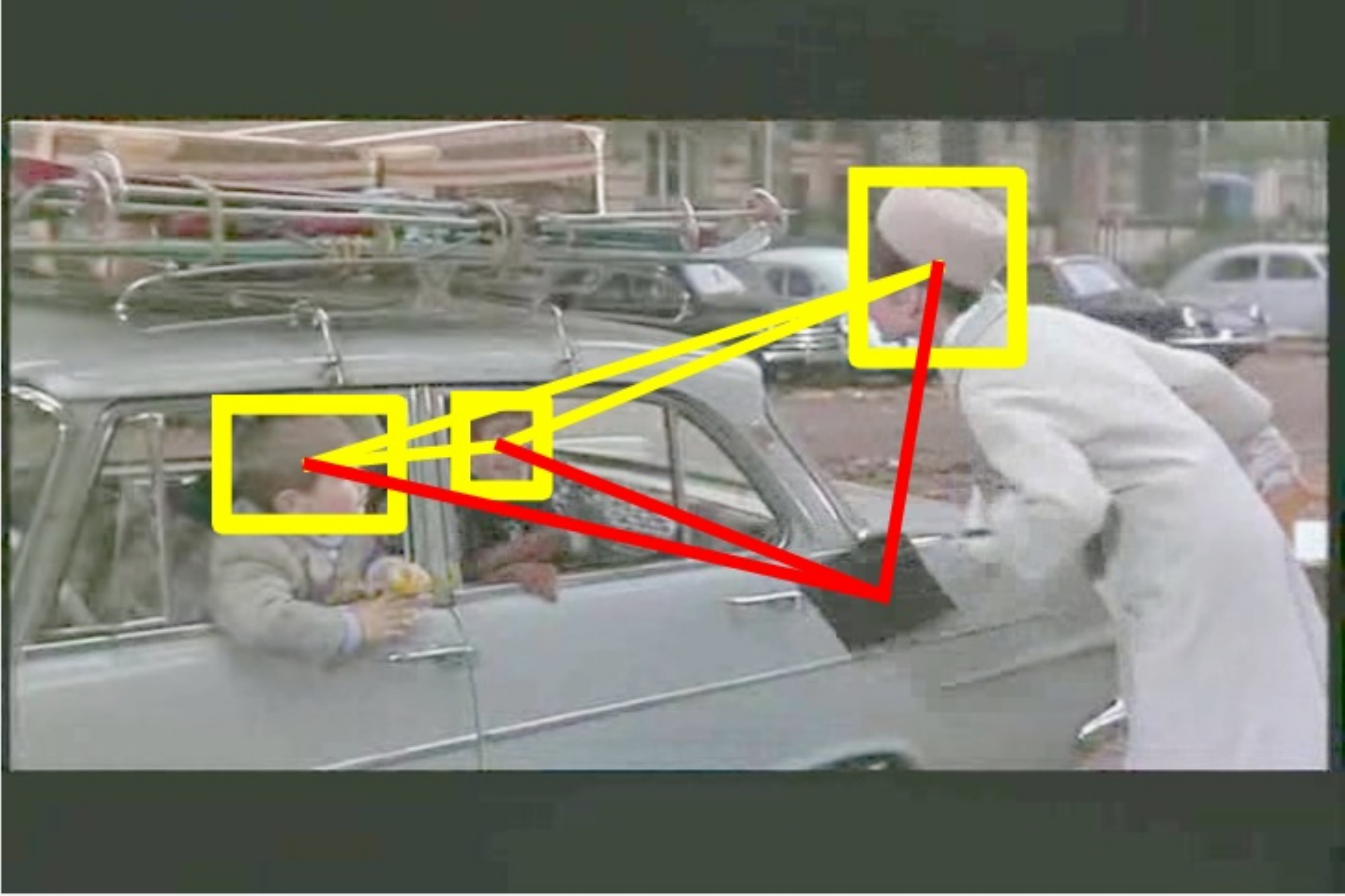} \\[-0.1cm]
\includegraphics[trim = 2mm 24mm 4mm 23mm, clip, width=0.48\linewidth]{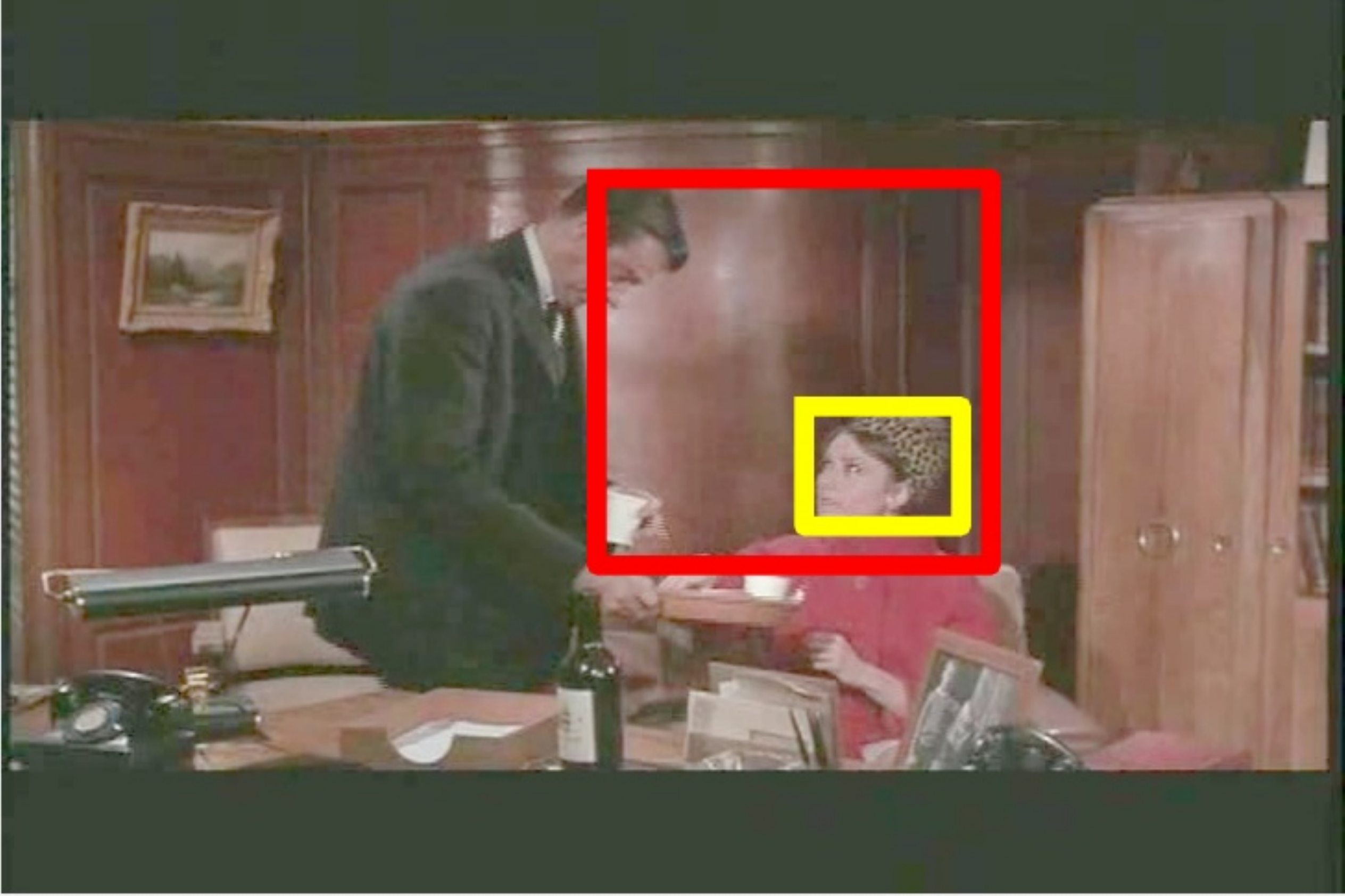}
&
\includegraphics[trim = 2mm 24mm 4mm 23mm, clip, width=0.48\linewidth]{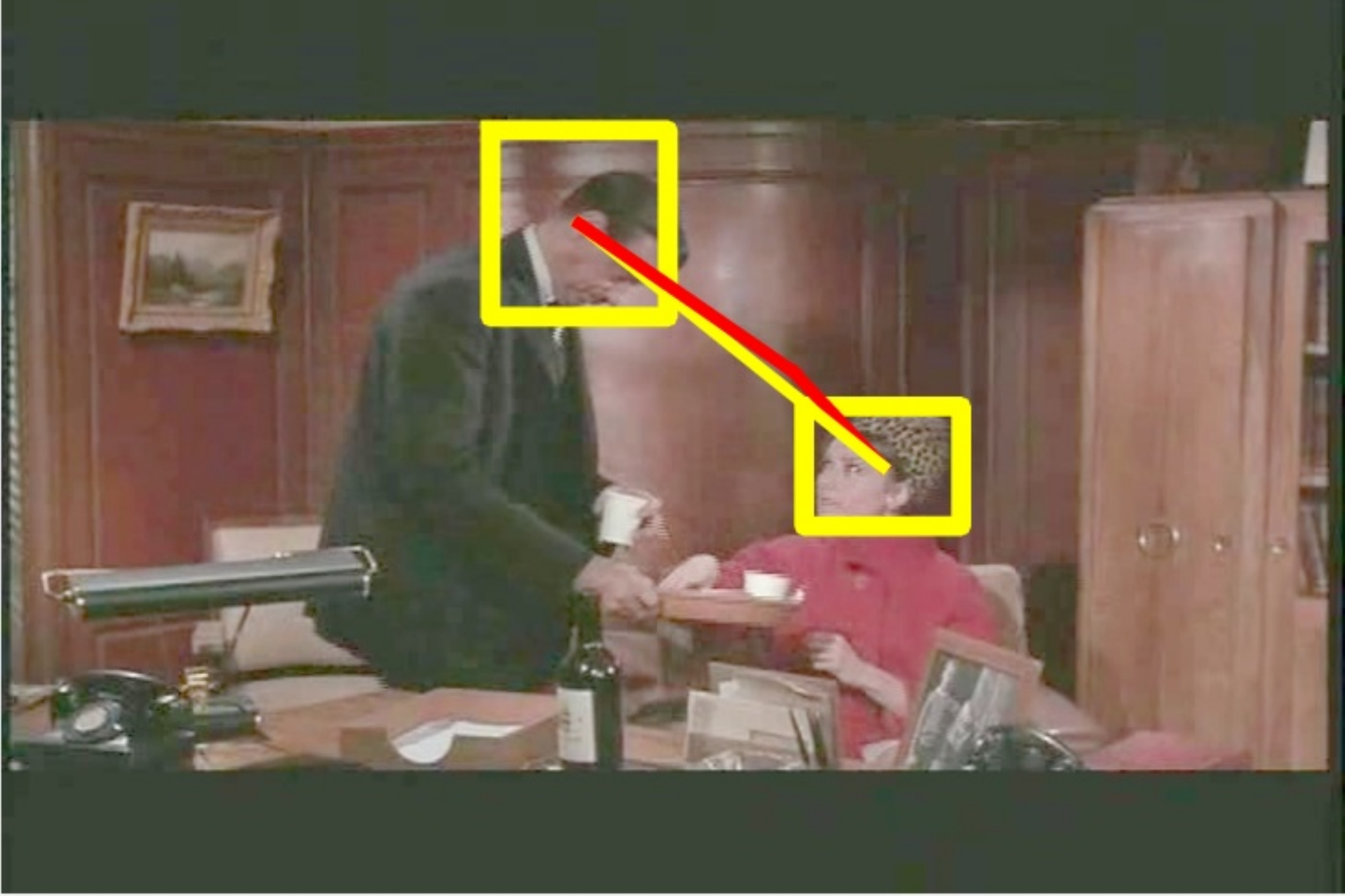}
\\[-0.1cm]
\includegraphics[trim = 2mm 24mm 4mm 23mm, clip, width=0.48\linewidth]{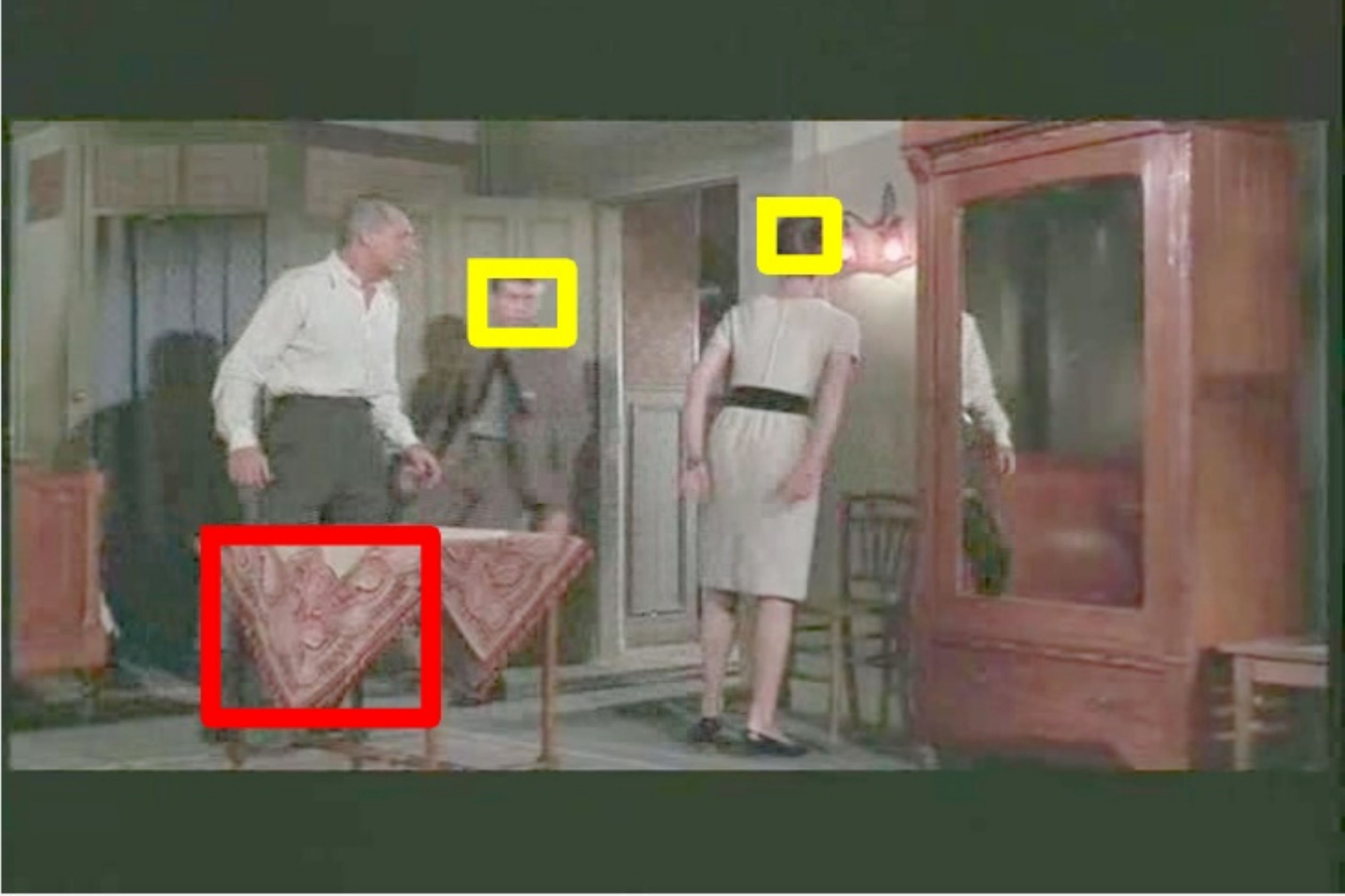}
&
\includegraphics[trim = 2mm 24mm 4mm 23mm, clip, width=0.48\linewidth]{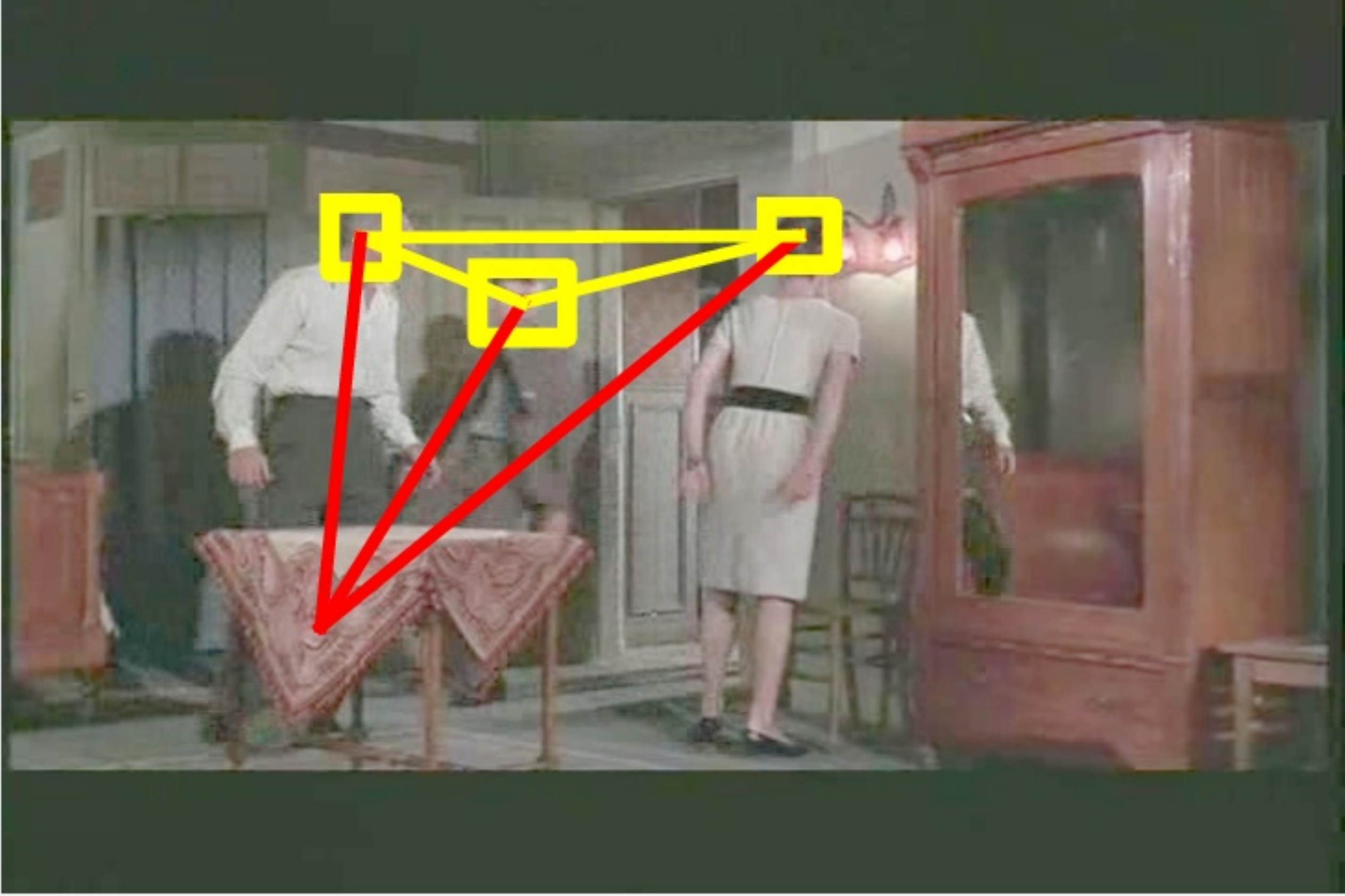}
\\[-0.1cm]
\includegraphics[trim = 0mm 0mm 0mm 0mm, clip, width=0.48\linewidth]{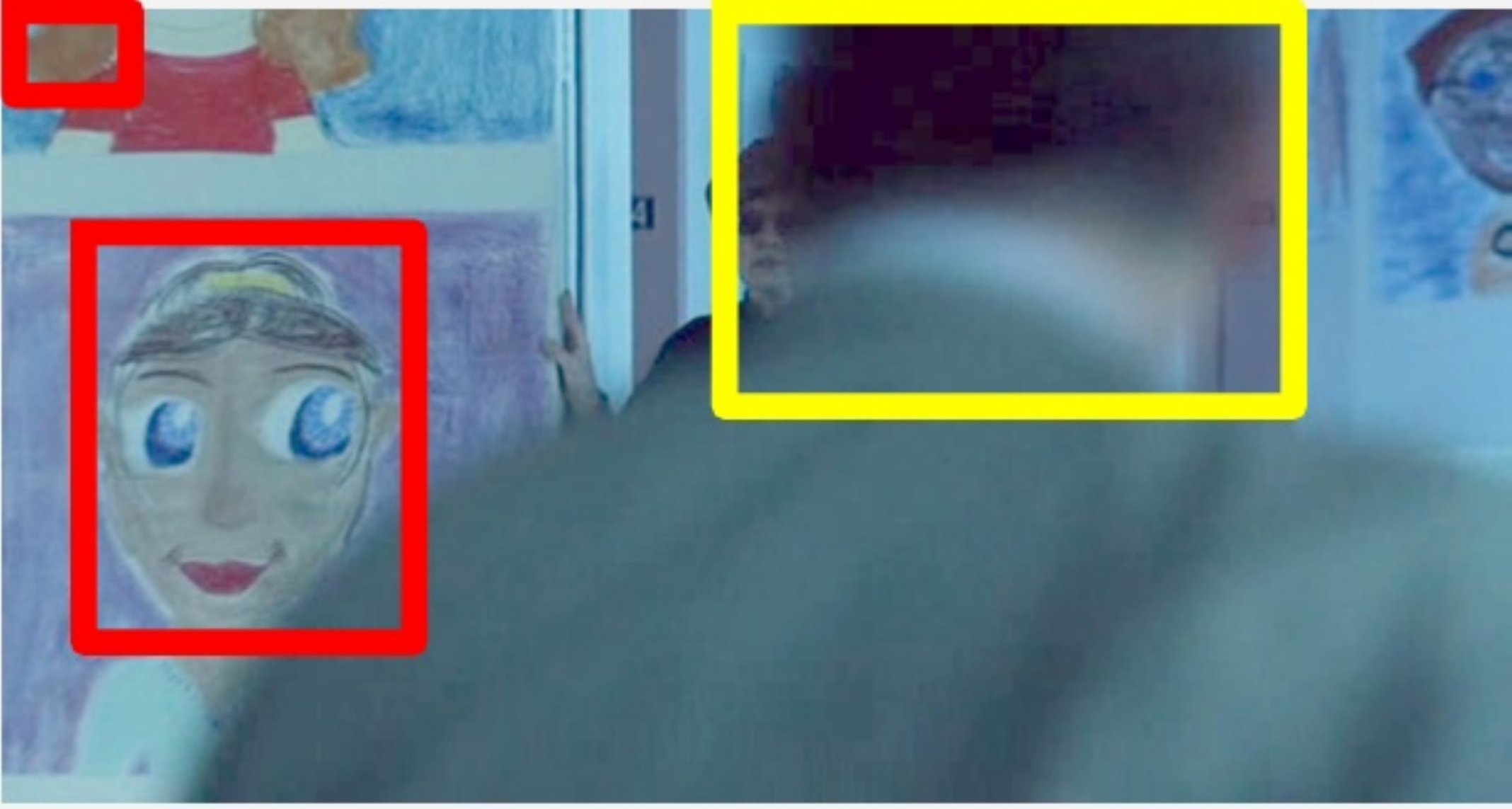}
&
\includegraphics[trim = 0mm 0mm 0mm 0mm, clip, width=0.48\linewidth]{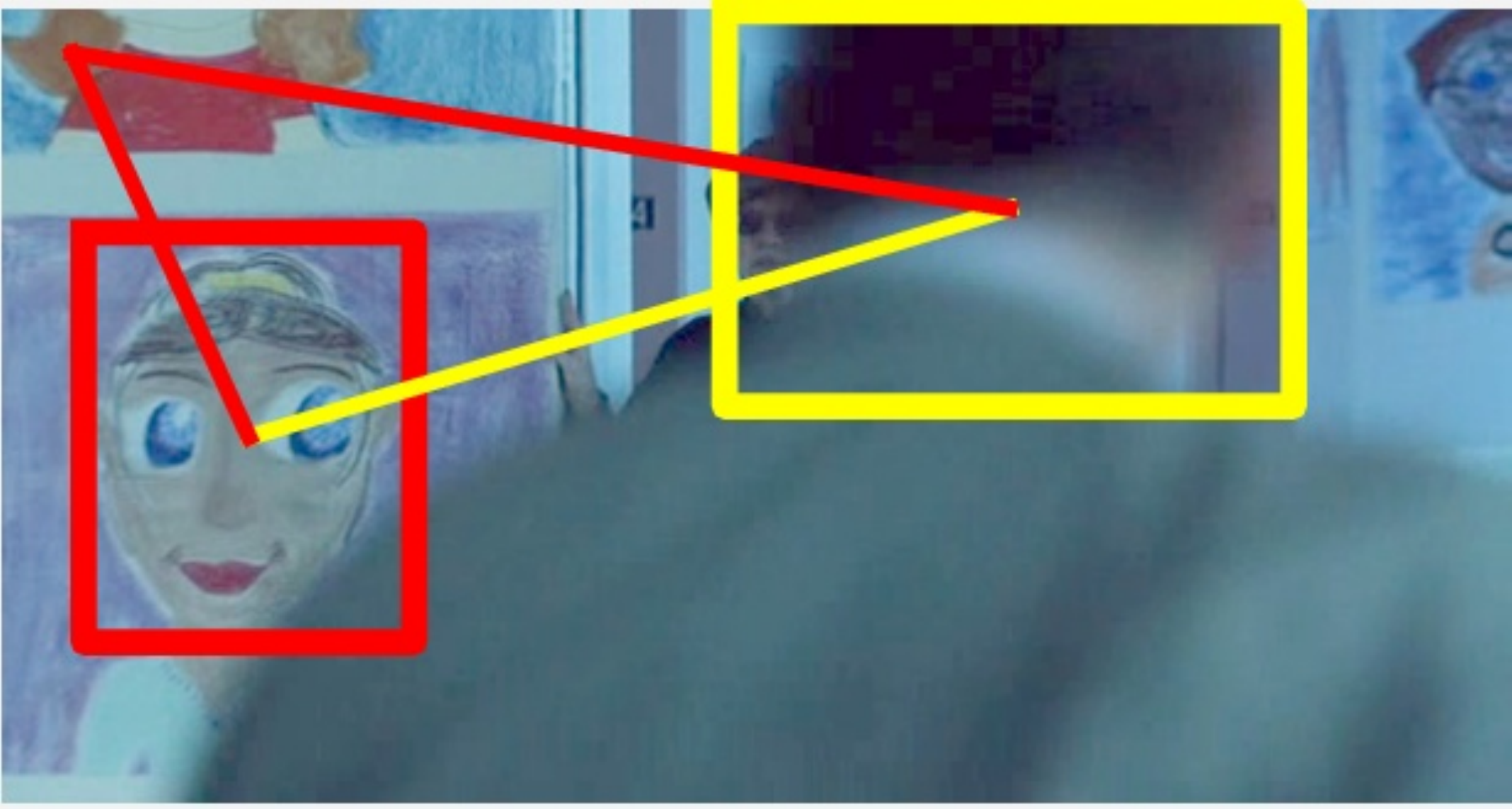}
\\[0.2cm]
\end{tabular}
\end{center}
\caption{Qualitative results for the Pairwise model. For each video frame we show results of the Local model (left) and the Pairwise model (right). For both methods we choose the score thresholds such that the precision equals the recall on the validation set. The plotted bounding boxes show the detections with the scores above the selected thresholds. Yellow boxes correspond to correct detections, red~-- to false positives. For the Pairwise model we show the strength of links between the candidates detected by at least one method. Links above a strength threshold (attractive) are plotted yellow and others~-- red (repulsive).
}\vspace{-.2cm}
\label{fig:resultsPairwiseTerm}
\end{figure}

We compare performance of the following four models: the Local model (Sec~.\ref{sec:localmodel}), the combination of the Local and Global models (Section~\ref{sec:globalmodel}), the combination of the Local and Pairwise models~(Section~\ref{sec:pairwisemodel}) and the combination of all the three proposed models.
The performance of head detection is evaluated on HollywoodHeads, Casablanca and TVHI datasets.
Qualitative results of the Global and Pairwise models are illustrated in Figures~\ref{fig:globalModel_examples} and~\ref{fig:resultsPairwiseTerm} respectively.
Table~\ref{tbl:our_performance_exp} presents quantitative results for all models.
We observe that the Global and Pairwise models consistently improve the performance of the baseline Local model. The combination of all three models demonstrates the best performance on all three datasets.

\begin{table}[t]
\begin{center}
\begin{tabular}{l|cccc}
Test set & \parbox{1.0cm}{\centering Local} & \parbox{1.2cm}{\centering Local\\\vspace{-.10cm}Global\vspace{.05cm}} & \parbox{1.2cm}{\centering Local\\\vspace{-.10cm}Pairwise\vspace{.05cm}} & \parbox{1.2cm}{\centering Local\\\vspace{-.10cm}Pairwise\vspace{.05cm}\\\vspace{-.10cm}Global\vspace{.05cm}} \\
\hline
Casablanca& 71.8 & 72.1 & 72.5 & \textbf{72.7} \\
HH & 71.8 & 72.5 & 71.9 & \textbf{72.7} \\
TVHI & 87.8 & 89.5 & 89.2 & \textbf{89.8}\\[-0.2cm]
\end{tabular}
\end{center}
\caption{Performance (\% AP) of different context-aware models on three datasets: Casablanca, HollywoodHeads (HH) and TVHI.}
\label{tbl:our_performance_exp}
\vspace{-0.5cm}
\end{table}

\subsection{Comparison with the state-of-the-art methods \label{sec:exp:rcnnComparison}}
We compare our approach against several baselines: the CNN-based object detector~\cite{girshick14} (R-CNN), DPM-based face detector~\cite{mathias2014face} (DPM Face) as well as other methods reporting results on TVHI~\cite{hoai14} (UBC+S) and Casablanca~\cite{ren2008finding} (VJ-CRF). We have trained R-CNN\footnote{\url{https://github.com/rbgirshick/rcnn}} object detector on human heads using the training subset of HollywoodHeads dataset. The CNN model was first fine-tuned on all region proposals used to train our Local model. Given memory limitations, the SVM phase of R-CNN training was done on a subset of training images.
For the DPM-based face detector we have used the vanilla DPM model provided by~\cite{mathias2014face}. Results of other methods were taken from original papers.

\begin{figure*}[!th]
\centering	
\begin{tabular}{ccc}	
\textit{HollywoodHeads} & \textit{TVHI} & \textit{Casablanca}\\[-0.0cm]	
\includegraphics[width=0.32\linewidth]{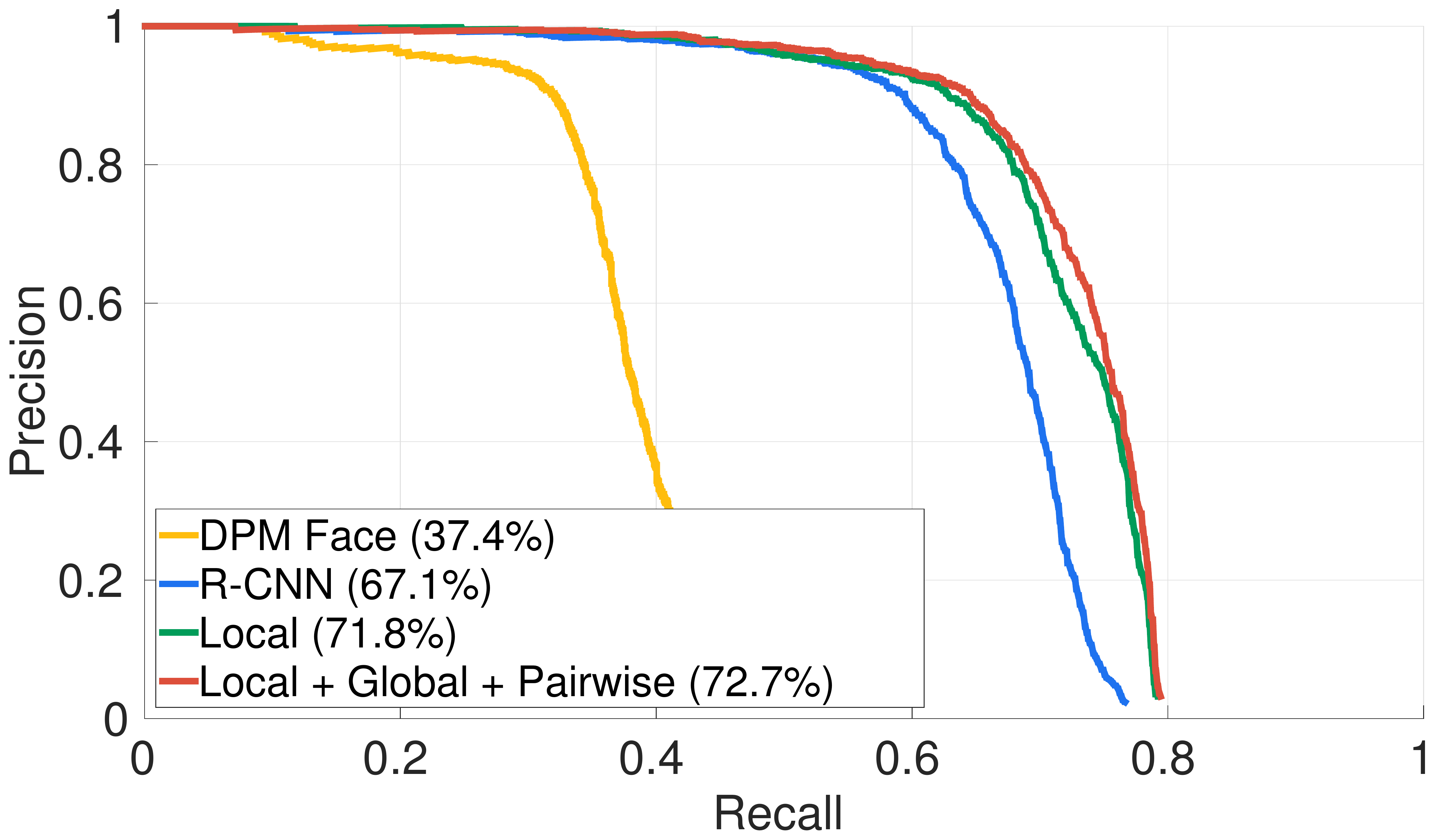}
&
\includegraphics[width=0.32\linewidth]{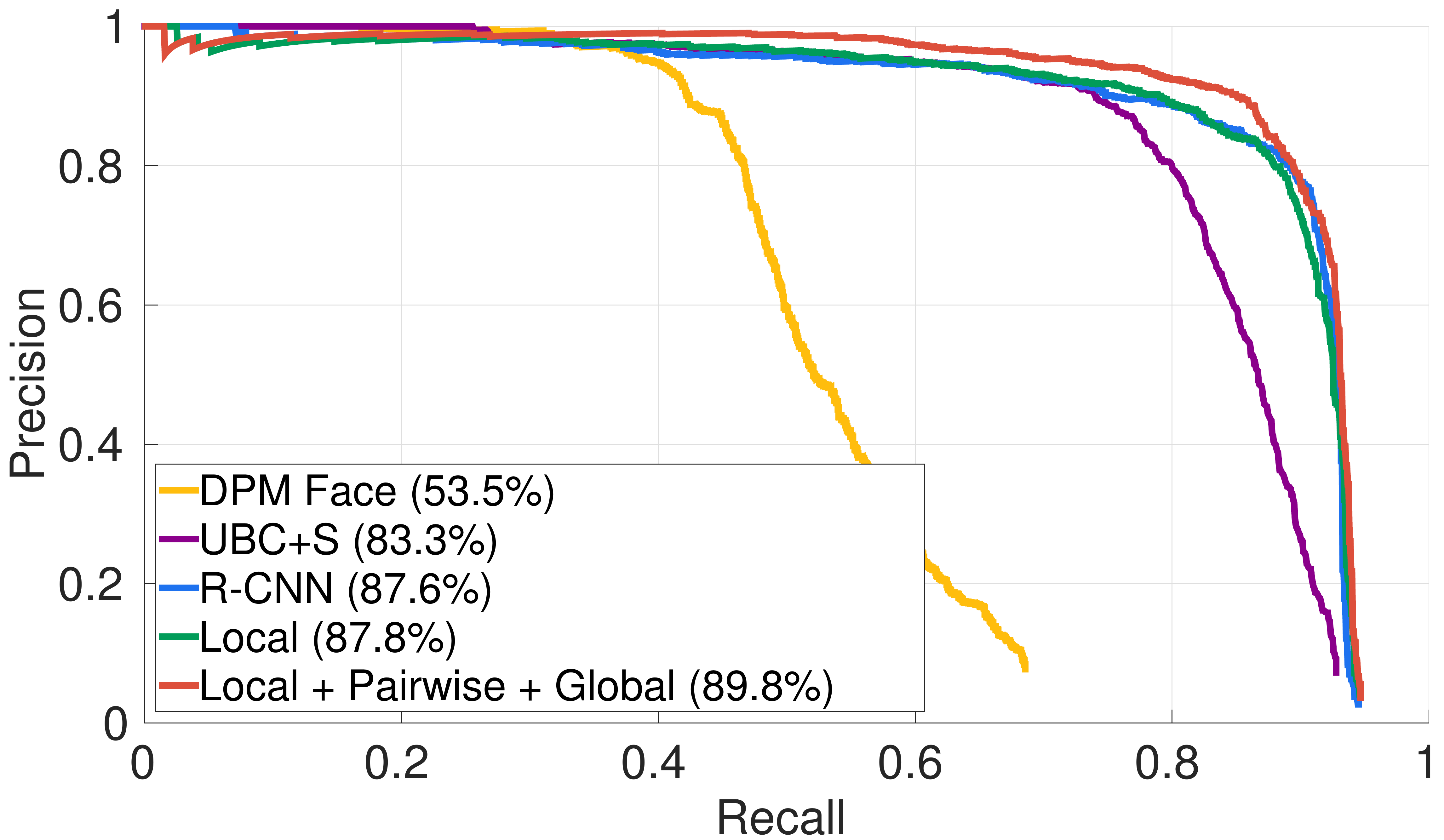}
&
\includegraphics[width=0.32\linewidth]{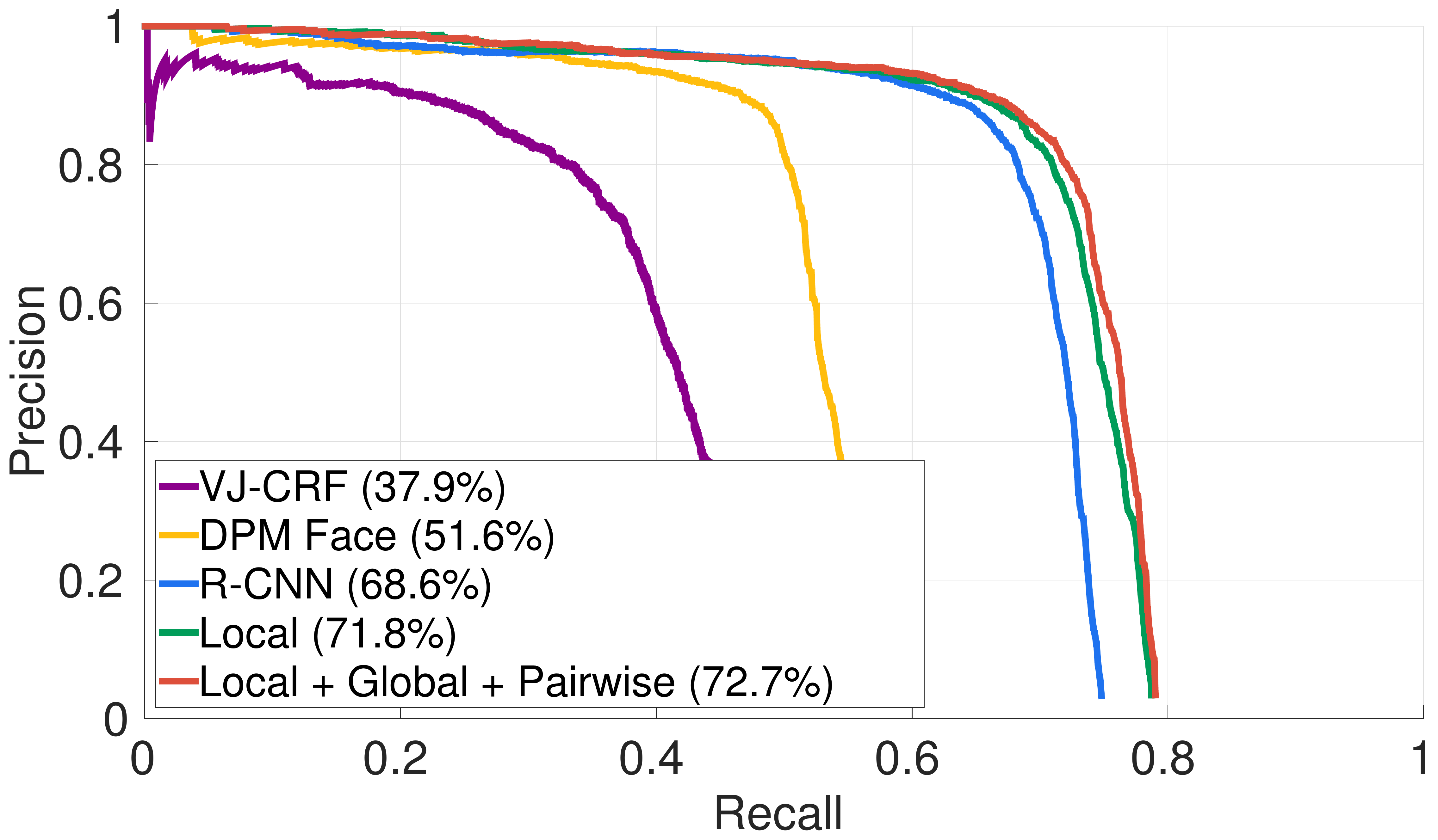} 
\end{tabular}	
\caption{Results of our method compared to the state-of-the-art on the three datasets.}	
\label{fig:performance_comp}	
\vspace{-0.3cm}
\end{figure*}

Results of all compared methods are presented in Figure~\ref{fig:performance_comp}.
Our joint model outperforms other methods on all three datasets.
Consistently with other recent evaluations, we observe the advantage of CNN-based methods compared to other baselines.
As expected, methods trained to detect faces achieve lower recall on the head detection task given the large variation of view points in natural images.
Our method significantly outperforms R-CNN on two out of three datasets and performs slightly better than R-CNN on the TVHI dataset.

Note that our evaluation on the Casablanca dataset differs from~\cite{ren2008finding} due to the improved annotation and the use of VOC evaluation procedure. Our results using the original evaluation setup by~\citet{ren2008finding} are reported in Appendix~\ref{sec:casa_supp}. Additional results of our method are available from the project web-page~\cite{projectwebpage} and in Appendix~\ref{sec:qual_res}.

\subsection{Architectures of the Local model.} \label{sec:exp:localArch}
In this section we compare performance and speed of different architectures of the Local model. We consider AlexNet architecture~\cite{krizhevsky12}, VGG-S~\cite{chatfield2014}, VGG-verydeep-16~\cite{simonyan14} provided with the MatConvNet framework~\cite{vedaldi2014matconnet}\footnote{\url{http://www.vlfeat.org/matconvnet/pretrained/}} and the model of~\citet{Oquab14}. All models were pre-trained on the ImageNet dataset~\cite{deng2009imagenet} and fine-tuned on the training set of the HollywoodHeads dataset as the Local model.
In Table~\ref{tbl:compare_pretrained_net} we report values of AP produced by different models together with the train/test speed. We measure the speed as the number of image patches processed per second. For each model we choose the size of the training batch such that the training speed is maximal. In all cases it happens to be the maximum batch size that fits into the GPU memory. Experiments of this section were run on NVIDIA TITAN X with 12G RAM.

\begin{table}
\begin{center}
\begin{tabular}{@{\:}l|@{\;\:}c@{\;\:}@{\;\:}c@{\;\:}@{\;\:}c@{\;\:}@{\;\:}c@{\;\:}}
 & AlexNet & Oquab & VGG-S & verydeep-16 \\
\hline
AP&76.3&76.7&77.2&\textbf{78.5}\\
\hline
Train speed &\textbf{445}&284&147&30\\
Test speed &\textbf{1490}&980&510&74 
\end{tabular}
\end{center}
\caption{Performance (\% AP) of Local models of different architectures on the HollywoodHeads validation set. Bottom lines report training and testing speed, measured by the number of image patches processed per second.}
\label{tbl:compare_pretrained_net}
\vspace{-0.1cm}
\end{table}

\begin{table}
\begin{center}
\begin{tabular}{l|ccc}
 Test set & 4 movies & 8 movies & 15 movies  \\
 \hline
 Casablanca& 51.2 & 62.5 & \textbf{72.7}  \\
 HollywoodHeads& 63.3  & 67.7 & \textbf{72.7} \\
 TVHI& 88.6 & 88.8 & \textbf{89.8} 
\end{tabular}
\end{center}
\caption{Performance of models trained on the training sets of different sizes. We report \% AP for each test set.}
\label{tbl:less_mov_exp}
\vspace{-0.1cm}
\end{table}

\subsection{Size of the training set \label{sec:exp:trainingSize}}
In this experiment we analyze the amount of training data required to train our models.
Our full training set is constructed from $15$ movies.
We also examine the use of smaller training sets corresponding to the first 8 movies and the first 4 movies of the full training set respectively.
We use each training set to train parameters of our full model and evaluate it on three datasets.
Corresponding results are reported in Table~\ref{tbl:less_mov_exp}. We observe that the amount of the training data and, maybe more importantly, its diversity helps to improve the performance.

\subsection{Complexity reduction with the Global model \label{sec:exp:speedUp}}
Here we show that the Global model can suppress false candidates and reduce the computational complexity of R-CNN and our Local model at test time. We achieve this by transferring scores of the Global model detection proposals. We then filter out low-score candidates and thus reduce the number of candidates that have to pass through Local CNN. We evaluate the performance of detectors with different percentage of candidates left after the filtering.
Table~\ref{tbl:speed_up_global} presents results of this experiment. We observe that detection performance remains high despite aggressive filtering by the scores of the Global model.

\begin{table}
\begin{center}
\begin{tabular}{@{\:}l|@{\;\:}c@{\;\:}@{\;\:}c@{\;\:}@{\;\:}c@{\;\:}@{\;\:}c@{\;\:}@{\;\:}c@{\;\:}@{\;\:}c@{\;\:}@{\;\:}c@{\:}}
\% left  & \centering 100 & 30 & 20 & 10 & 6 & 4 & 2 \\
\hline
R-CNN &  $67.1$ & $65.0$ & $63.9$ & $59.0$ & $53.7$ & $48.9$ & $41.3$ \\
Local &  $71.8$ & $68.3$ & $66.8$ & $60.2$ & $53.4$ & $48.8$ & $41.9$ 
\end{tabular}
\end{center}
\caption{Performance of the R-CNN method and of our Local model (\% AP) on the test set of HollywoodHeads with different percentage of candidates left after filtering using the Global model.}
\label{tbl:speed_up_global}
\vspace{-0.1cm}
\end{table}

\section{Conclusion}
\label{sec:conclusions}
In this work we have addressed the task of detecting people in still images. We proposed two context-aware CNN-based models.
To train and evaluate our method, we have collected the new large-scale HollywoodHeads dataset consisting of movie frames and human head annotations. 
The combination of our context-aware models and the CNN-based local detector achieves state-of-the-art results on our dataset and the two existing human detection datasets, TVHI and Casablanca.

We believe that our context-aware models can be extended to tackle general object classes. In particular, the Microsoft COCO dataset~\cite{Lin14coco} contains many small object classes with implied spatial constraints. Another possible direction for future work is to take into account motion information to extend our methods to perform long-term tracking.

\vspace{0.1in}
\noindent {\bf Acknowledgements} \;\,{\small
This work was partly supported by the ERC grant Activia (no. 307574) and the MSR-INRIA Joint Center.
}


\appendix
\section{Implementation details \label{sec:impl_dtl}}
\subsection{Local model \label{sec:app:local}}
To train the Local model, we assign each candidate region to the positive (head) or negative (background) class. For a given image, we make this assignment based on the intersection-over-union (IoU) overlap ratio~$o$ of the candidate bounding box with the best matching ground-truth bounding box. Specifically, candidates with $o > 0.6$ are labeled as positives and candidates with $o < 0.5$ are labeled as negatives. The remaining candidates are considered ambiguous and are not used at the training. Following~\cite{girshick14} we exploit the context padding. Each candidate is resized to $188 \times 188$ square patch which is extended with~$18$ pixels on each side filled from the original image. The input images of our CNN are of size $224 \times 224$.
For each image, we form a training batch by sampling 64 proposals such that the balance between classes is roughly maintained.

We initialize parameters of the network using the ImageNet pre-trained network of~\citet{Oquab14}. We optimize the parameters of the network by minimizing the sum of independent log-losses with a stochastic gradient descent (SGD) algorithm with momentum~$0.9$ and weight decay~$0.0005$. We initialize the learning rate at~$0.01$, and decrease it several times by a factor of~$10$ after the validation error reaches saturation.

\subsection{Global model \label{sec:app:global}}

The Global model takes the whole image (isotropically rescaled and zero-padded to size $224 \times 224$) as the input and provides a vector of~$284$ numbers as the output. Each element of the output vector is associated with a cell of our multi-scale grid. For each cell we construct a target objective: $1$ is the corresponding image patch has at least~$0.3$ IoU ratio with at least one ground-truth object bounding box. To train the Global model we optimize the sum of independent log-losses with an SGD algorithm. We initialize the model with the ImageNet pre-trained network~\cite{Oquab14}. The learning rate of SGD is set to~$0.0001$, momentum~-- to~$0.9$, weight decay~-- to~$0.0005$.

\subsection{Pairwise model \label{sec:app:pairwise}}
The number of candidates from one image that our Pairwise model can process is quite limited due to the complexity of the inference procedure. To select the ``good'' candidates out of the thousands produced by the selective search~\cite{uijlings2013selective} we use the non-maximum suppression (with threshold 0.3) on top of the scores provided by the Local model. We find that 16 candidates per image produced this way provide good balance between quality and speed.

To construct clusters of candidate pair (edges) incorporating the layout information we use the three features representing the vertical and horizontal displacements and the ratio of the candidate sizes. To be precise, if the position of each candidate is defined by a tuple $(x_i, y_i, w_i, h_i)$ we define the size of the candidate as $s_i = (w_i + h_i) / 2$, its horizontal position as $x^c_i = x_i + w_i / 2$ and its vertical position as $y^c_i = y_i + h_i / 2$.
For the two candidates sorted such that $x_i \leq x_j$ we compute the features as follows: $f_{ij}^1 = \log(s_i / s2)$, $f_{ij}^2 = \phi((x^c_j - x^c_i) / s_i) $, $f_{ij}^3 = \phi( (y^c_j - y^c_i) / s_i )$, where $\phi(x) = \sign(x) \log(|x| + 1)$.
All the three features are normalized to have zero mean and unit standard deviation on the training set. We find that increasing number of clusters beyond~$20$ does not improve the performance.

To train the Pairwise model we assign each selected candidate a target binary label based on the maximum IoU ratio with the ground-truth bounding boxes (threshold 0.5). We form a training batch from 64 candidates coming from 4 different images. The FE part of the model is initialized from the Local model. The weights of the UN and PN were initialized randomly using zero-mean Gaussians with standard deviation 0.01. The structured surrogate objective is optimized with and SGD with momentum 0.9, weight decay $0.000005$, and learning rate $0.00001$. We decreased the learning rate by a factor of~$10$ after 4 passes over the training data.

\subsection{Combining models \label{sec:app:combination}}
\paragraph{Local and Pairwise models.}


We now describe the process of computing the scores of the joint model. 
First, we compute the scores of the Local model for all candidates and perform the non-maximum suppression~\cite{felzenszwalb10dpm} using NMS threshold 0.3.
The 16 top-scoring detections produced by NMS are then used as input for the Pairwise model.
This number of candidates is sufficient on scenes with a few people, but can cause the drop of recall for crowded scenes (especially for some scenes of Casablanca dataset).
To compensate for this drop, we combine scores produced by the Local and Pairwise models~~$s_{l}$,$s_{p}$ respectively. For candidates with both scores existing, we use the affine combination $s_{lp}=\alpha s_{l} + (1-\alpha) s_{p} + \beta$. For candidates with the score of the Pairwise model non-existent, we use the score of the Local model $s_{lp}=  s_{l}$. Parameters $\alpha\in[0,1]$ and $\beta\in[-10,10]$ are selected by maximizing AP on the validation set using grid search.

\paragraph{Local, Pairwise and Global models.}
To combine scores $s_{lp}$ with the Global model, we associate each candidate with the output cell of the Global model having maximum IoU overlap-ratio. The score of the joint model $s^*$ is computed as an affine combination of the detection score $s_{lp}$ and the grid cell score $s_g$, i.e. $s^*=\gamma s_{lp} + (1-\gamma) s_g$ where  $\gamma\in[0,1]$ is obtained by maximizing AP on the validation set.

\subsection{Implementation details \label{sec:app:technical}}
All our experiments were run on NVIDIA GPUs using MATLAB-based MatConvNet~\cite{vedaldi2014matconnet} framework with the cuDNN backend~\cite{chetlur2014cudnn}. To avoid speed bottlenecks, we found it important to crop and resize image patches corresponding to object proposals using GPU which can be easily implemented using e.g.\ NVIDIA Performance Primitives (NPP) library provided in the CUDA package\footnote{\url{https://github.com/aosokin/cropRectanglesMex}}\!\!.

\paragraph{Running times.}
We report the running times of different parts of our model measured on NVIDIA TITAN X.
The forward and backward passes of the Local model on a batch of 64 proposals take~$0.08$s and~$0.18$s, respectively.
The forward and backward passes of the Global model on a batch of 32 images, take~$0.06$s and ~$0.12$s, respectively.
The Pairwise model consists of several parts: feature extractor, unary network, pairwise network, structured loss.
For a batch of 64 candidates taken from 4 images (16 candidates from each) the forward pass through a feature extractor network takes 0.07s, the unary network~-- 0.003s, the pairwise network~-- 0.003s.
The backward pass through these networks takes 0.2s, 0.004s and 0.004s, correspondingly.
The computation of the structured loss and its derivatives takes 0.01s per image.
Overall, the forward and backward passes through a joint Pairwise model take 0.36s for a batch coming from 4 images.



\section{Evaluation on the original Casablanca dataset \label{sec:casa_supp}}

\begin{figure}[h!]
\centering	
\includegraphics[width=\linewidth]{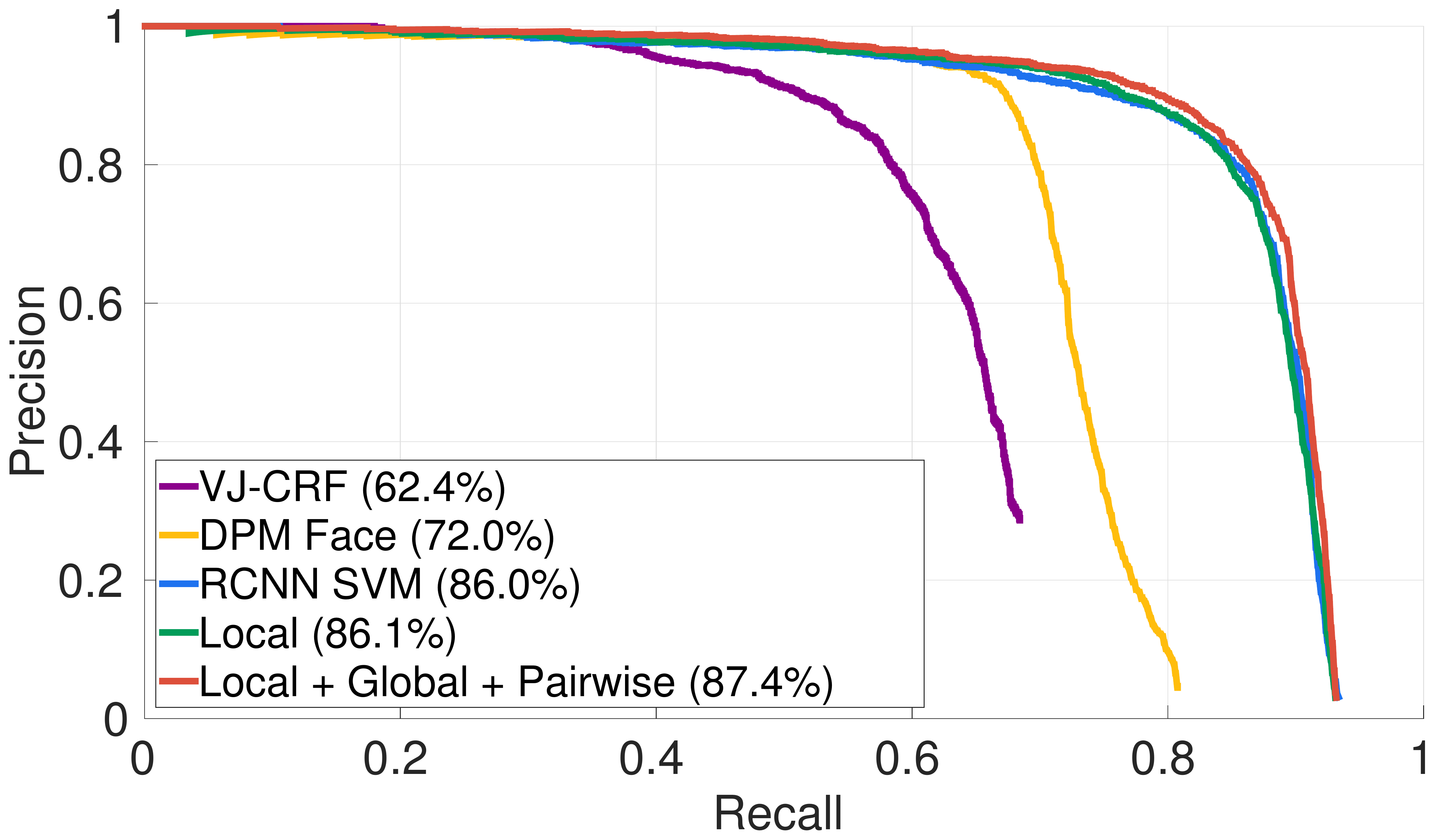}
\caption{Results of head detection in the original Casablanca dataset.}	
\label{fig:performance_comp_casa}	
\vspace{-0.2cm}
\end{figure}

To compare our results with the exact results reported in~\cite{ren2008finding}, we evaluate head detection on the Casablanca dataset using the original set of annotations and the evaluation procedure used in~\cite{ren2008finding}. Figure~\ref{fig:performance_comp_casa} demonstrates corresponding precision-recall curves. Our method significantly outperforms VJ-CRF~\cite{ren2008finding} as well as other baselines.

\begin{figure}[h!]
\centering	
\begin{tabular}{cc}	
Original annotation & {Improved annotation}\\[-0.0cm]	
\includegraphics[trim=30mm 30mm 30mm 30mm, clip, width=0.48\linewidth]{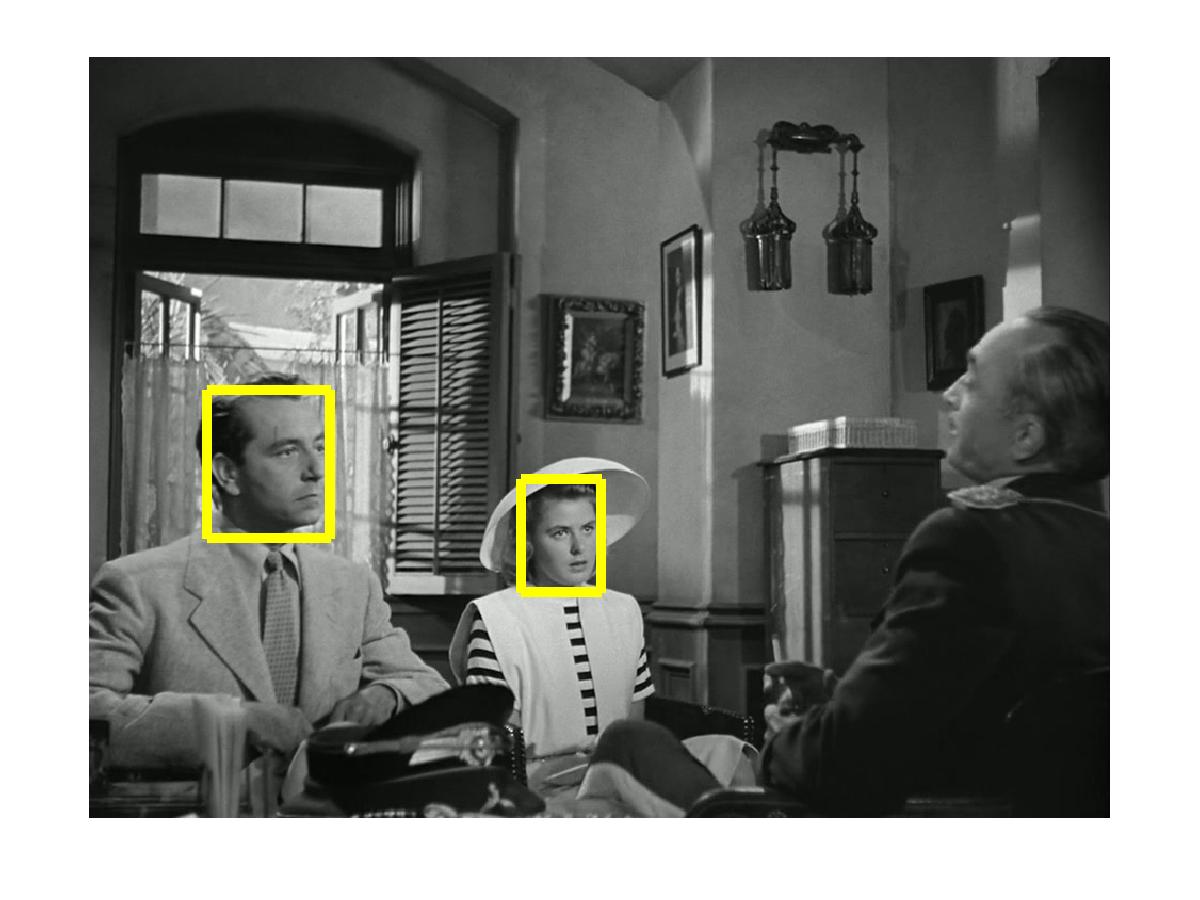}
&
\includegraphics[trim=30mm 30mm 30mm 30mm, clip, width=0.48\linewidth]{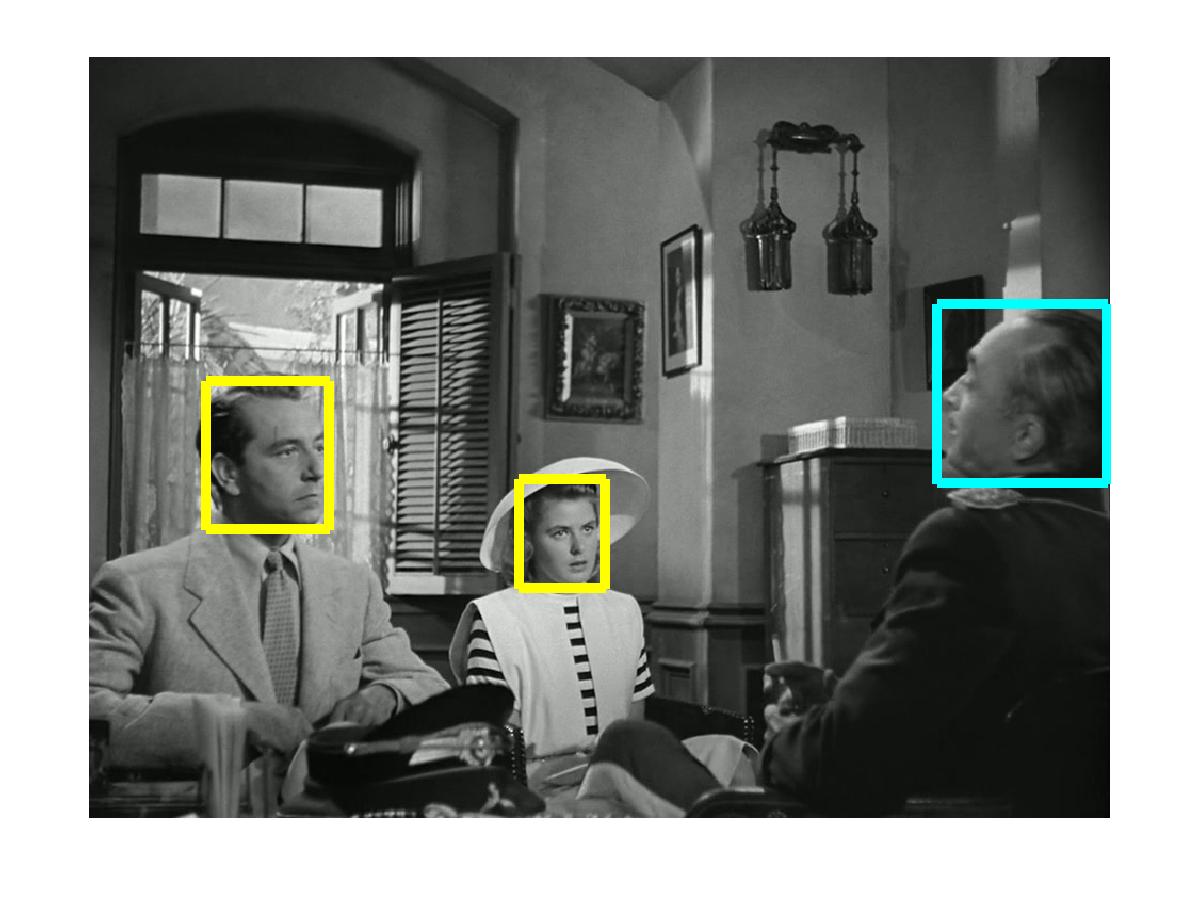}\\
\includegraphics[trim=30mm 30mm 30mm 30mm, clip, width=0.48\linewidth]{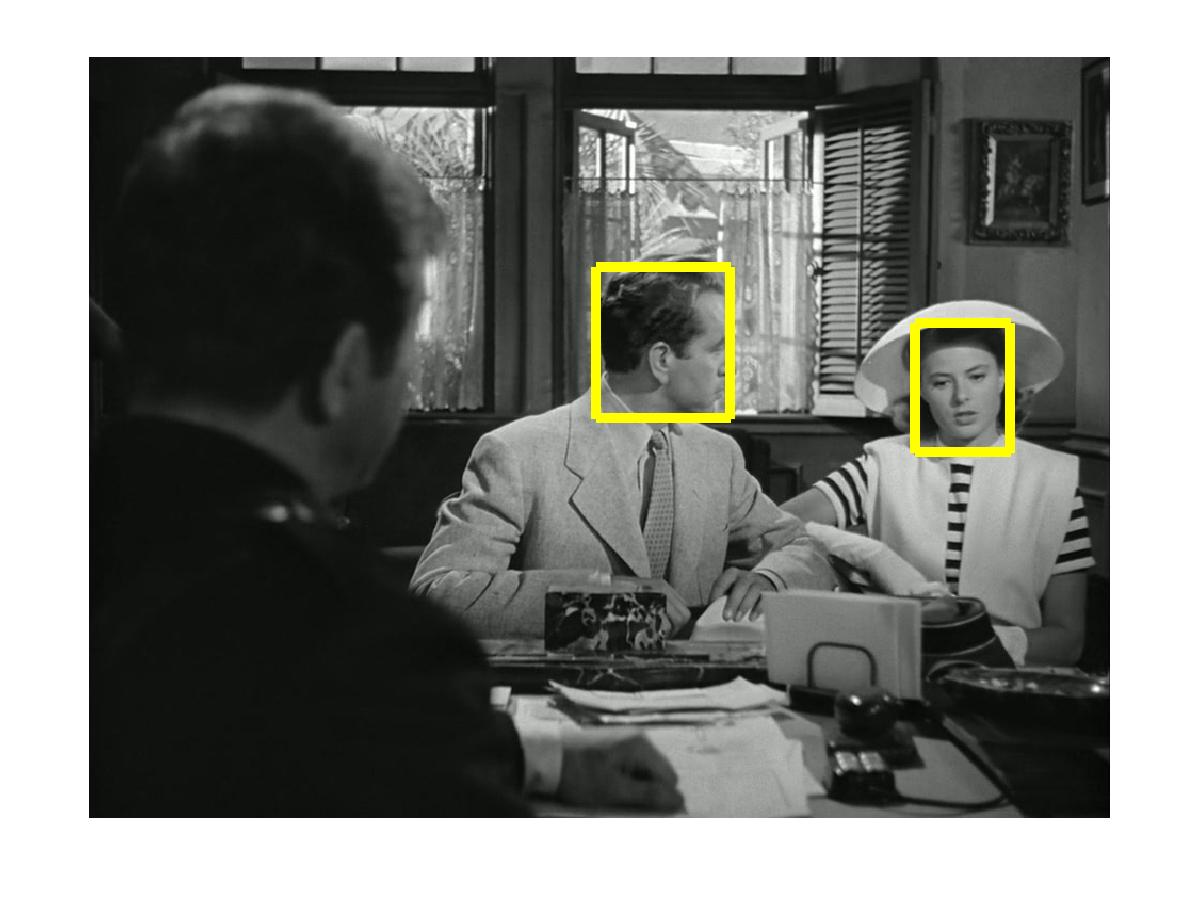}
&
\includegraphics[trim=30mm 30mm 30mm 30mm, clip, width=0.48\linewidth]{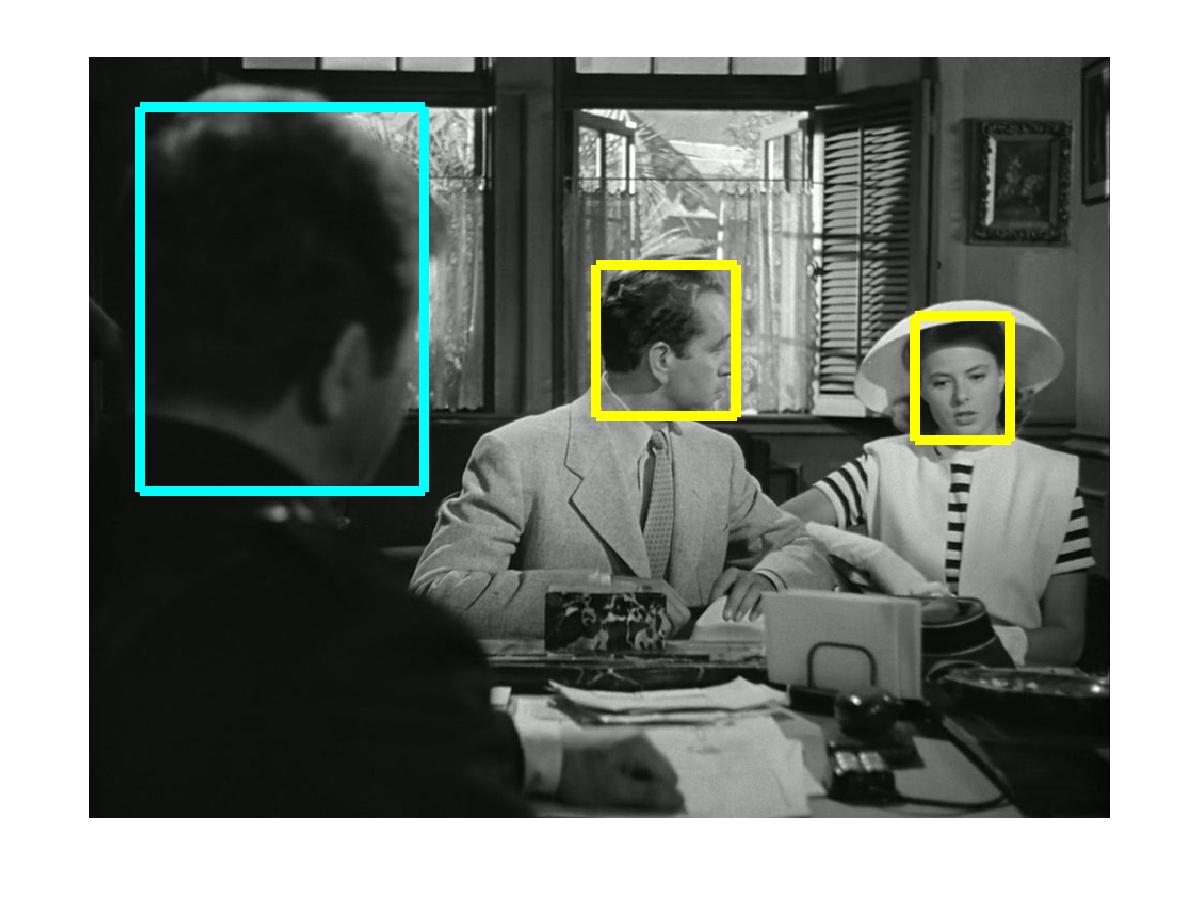}\\
\includegraphics[trim=30mm 30mm 30mm 30mm, clip, width=0.48\linewidth]{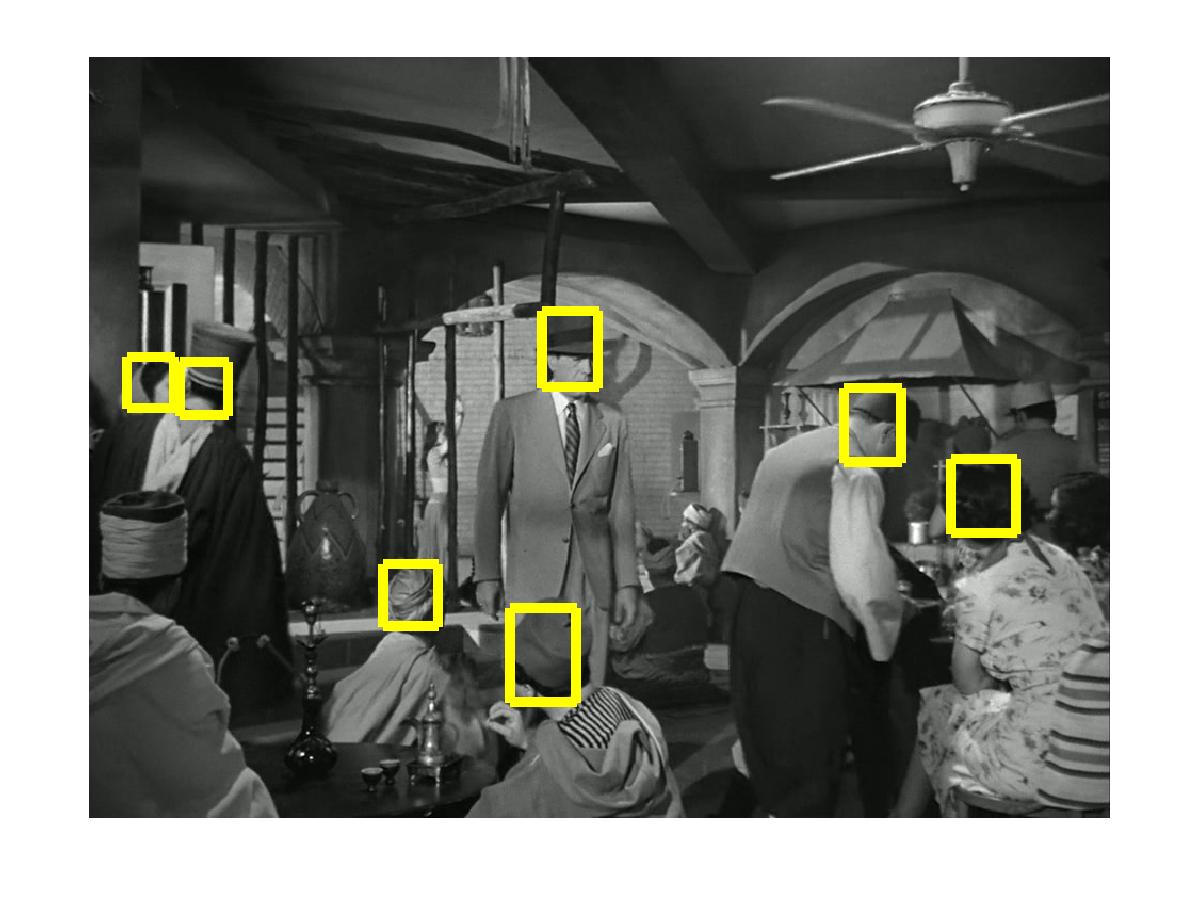}
&
\includegraphics[trim=30mm 30mm 30mm 30mm, clip, width=0.48\linewidth]{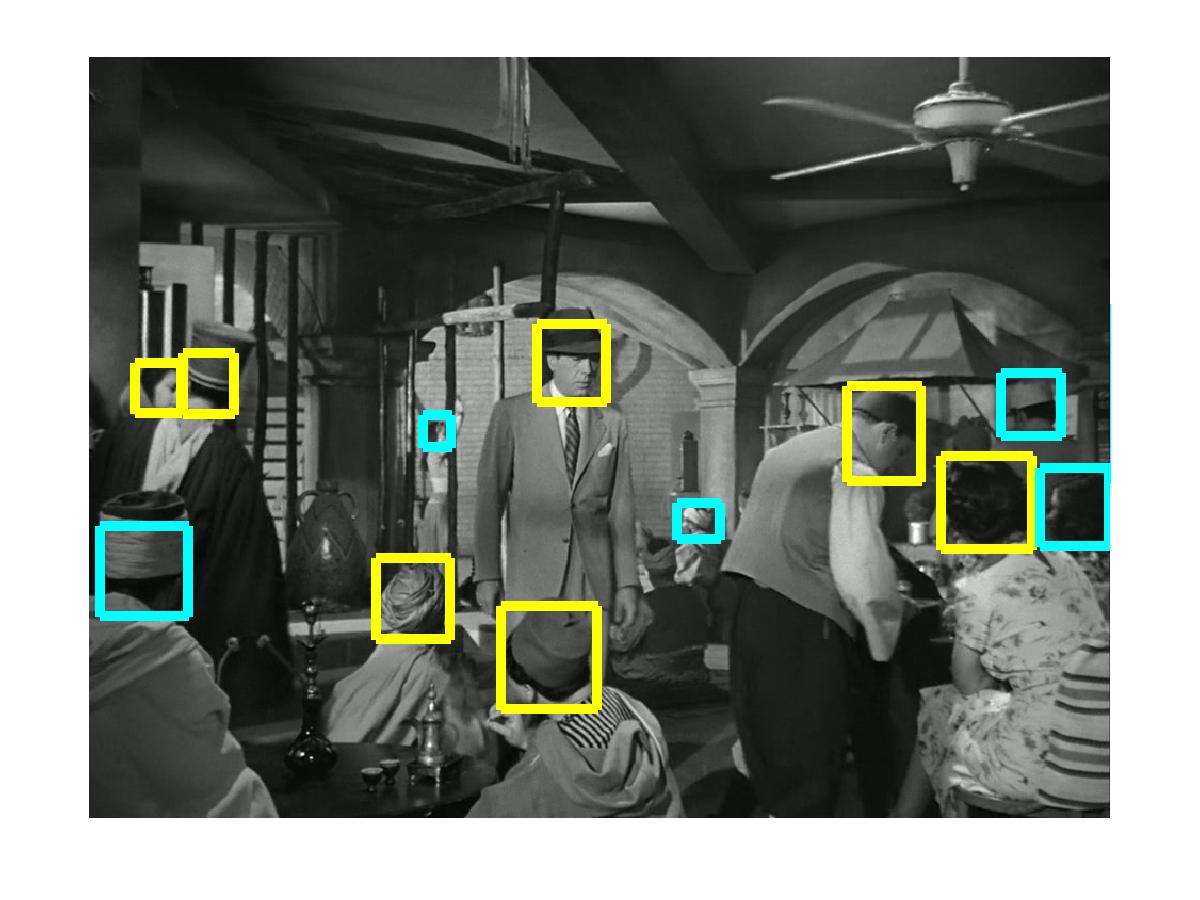}\\
\end{tabular}
\caption{(Left): Examples of original annotations in Casablanca dataset~\cite{ren2008finding}. (Right): Our corrected annotation with added head bounding boxes marked with cyan.}
\label{fig:fix_anno}	
\vspace{-0.2cm}
\end{figure}

As mentioned in Section~\ref{sec:casadataset}, the original Casablanca dataset~\cite{ren2008finding} contains many cases of missing and imprecisely localized head annotations. Figure~\ref{fig:fix_anno} (left) depicts some examples with missing annotations. To provide more conclusive results in Section~\ref{sec:experiments}, we have improved original annotation by adding missing and correcting existing annotations on all test frames defined in~\cite{ren2008finding}, see Figure~\ref{fig:fix_anno} (right). Despite our effort, some crowded scenes may still contain missing annotations of very small heads.

\begin{figure*}[ht]
\begin{center}
\begin{tabular}{c@{}c@{}c@{}c@{}c}
\includegraphics[trim=25mm 17mm 17mm 17mm, clip, width=0.2\textwidth]{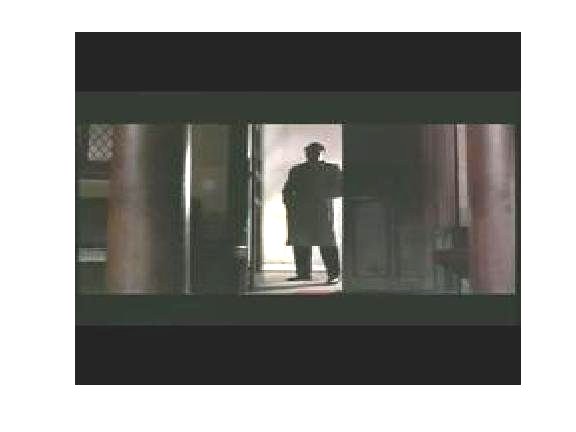}
&
\includegraphics[trim=25mm 17mm 17mm 17mm, clip, width=0.2\textwidth]{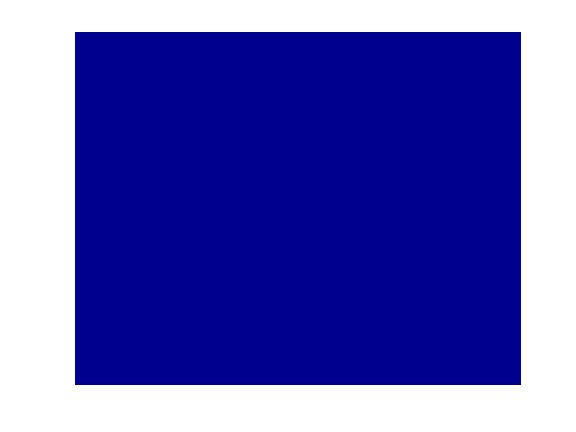}
&
\includegraphics[trim=25mm 17mm 17mm 17mm, clip, width=0.2\textwidth]{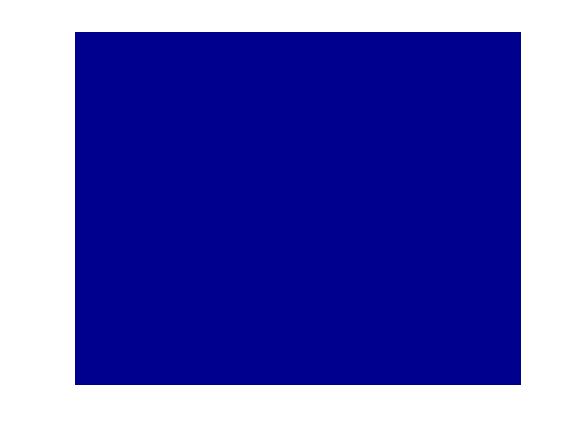}
&
\includegraphics[trim=25mm 17mm 17mm 17mm, clip, width=0.2\textwidth]{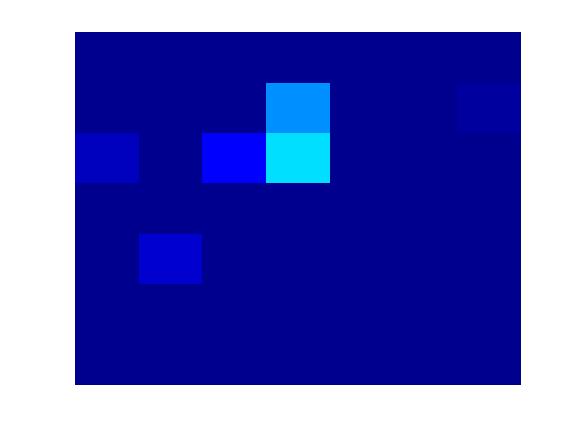}
&
\includegraphics[trim=25mm 17mm 17mm 17mm, clip, width=0.2\textwidth]{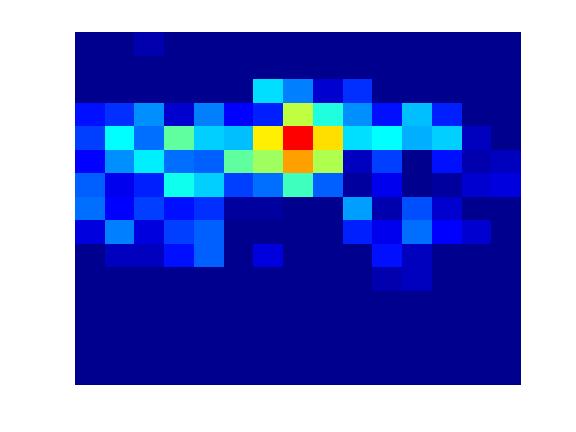}\\
\includegraphics[trim=25mm 17mm 17mm 17mm, clip, width=0.2\textwidth]{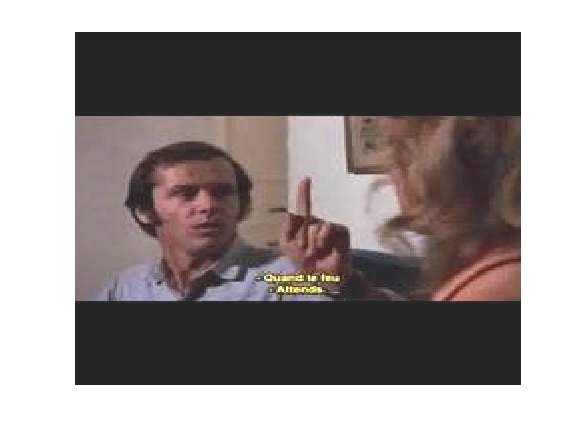}
&
\includegraphics[trim=25mm 17mm 17mm 17mm, clip, width=0.2\textwidth]{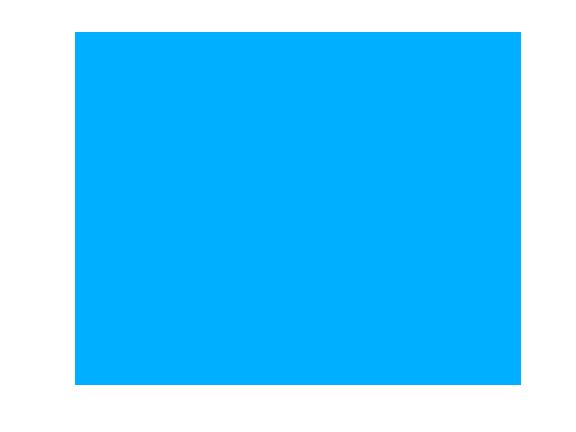}
&
\includegraphics[trim=25mm 17mm 17mm 17mm, clip, width=0.2\textwidth]{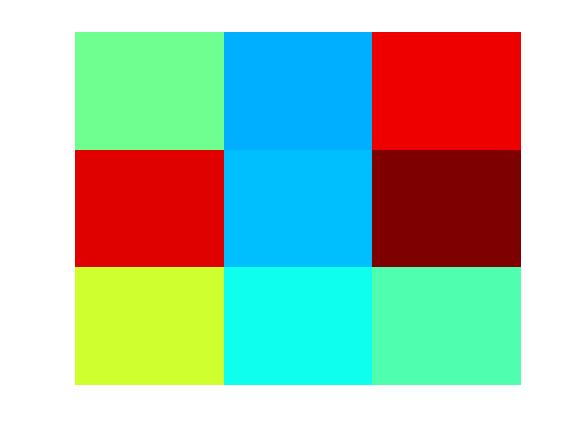}
&
\includegraphics[trim=25mm 17mm 17mm 17mm, clip, width=0.2\textwidth]{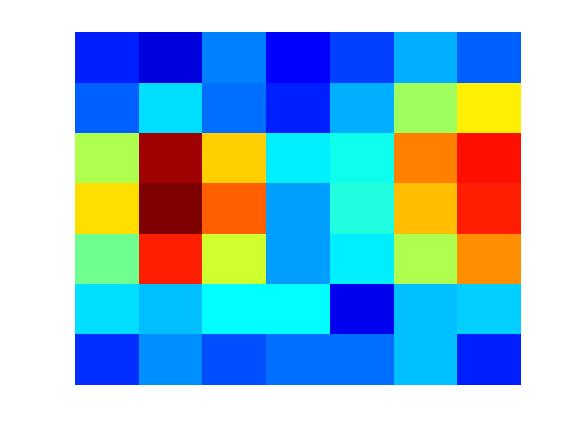}
&
\includegraphics[trim=25mm 17mm 17mm 17mm, clip, width=0.2\textwidth]{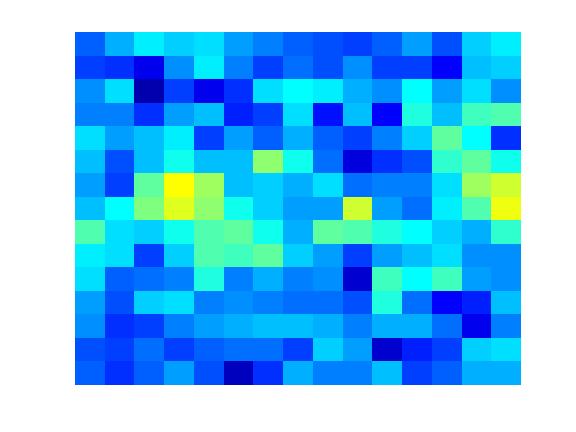}\\
\includegraphics[trim=25mm 17mm 17mm 17mm, clip, width=0.2\textwidth]{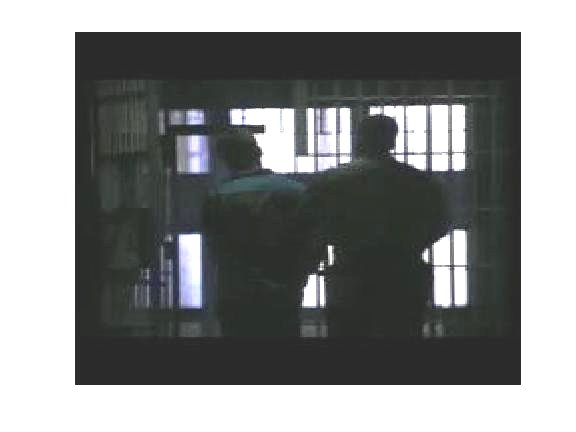}
&
\includegraphics[trim=25mm 17mm 17mm 17mm, clip, width=0.2\textwidth]{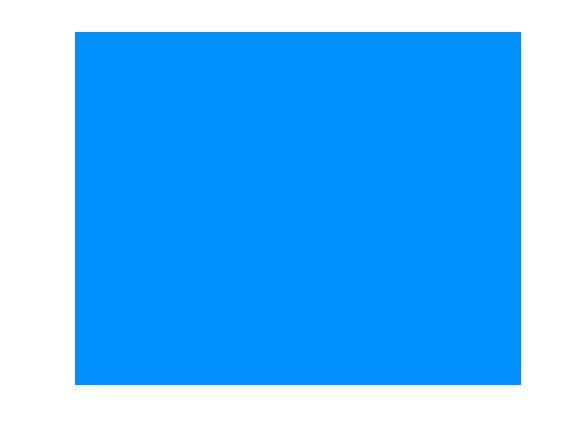}
&
\includegraphics[trim=25mm 17mm 17mm 17mm, clip, width=0.2\textwidth]{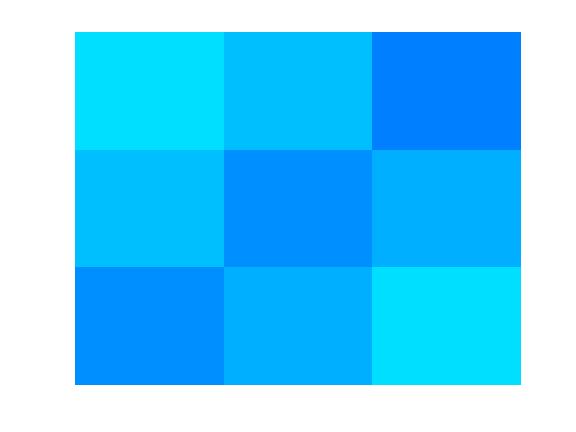}
&
\includegraphics[trim=25mm 17mm 17mm 17mm, clip, width=0.2\textwidth]{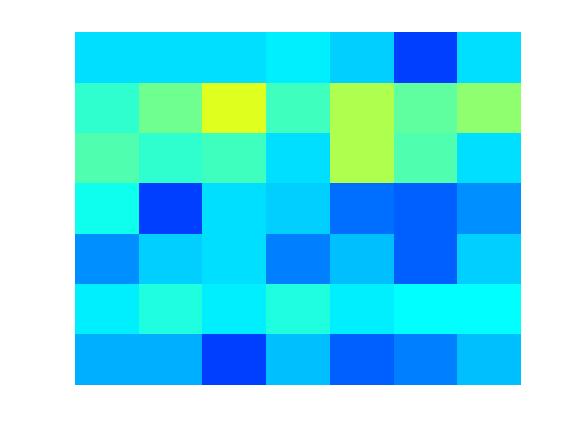}
&
\includegraphics[trim=25mm 17mm 17mm 17mm, clip, width=0.2\textwidth]{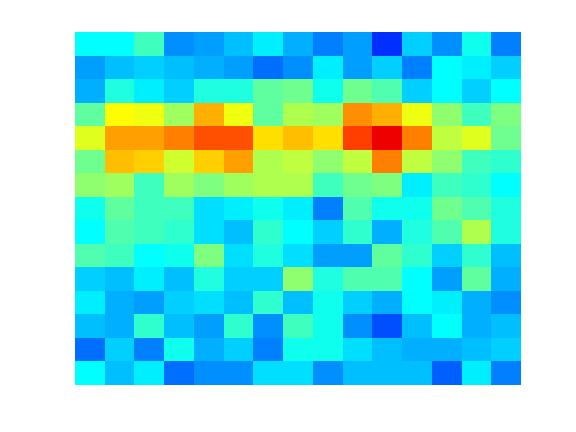}\\
\includegraphics[trim=25mm 17mm 17mm 17mm, clip, width=0.2\textwidth]{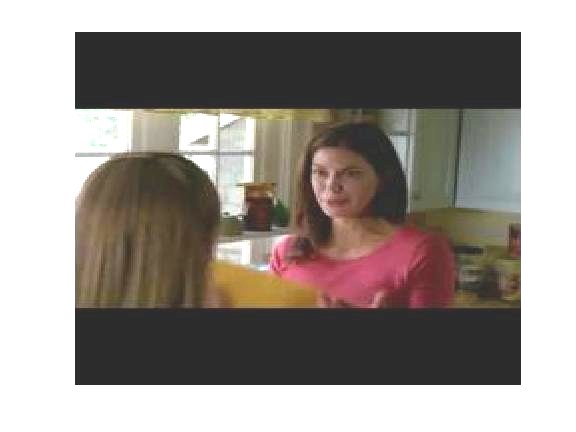}
&
\includegraphics[trim=25mm 17mm 17mm 17mm, clip, width=0.2\textwidth]{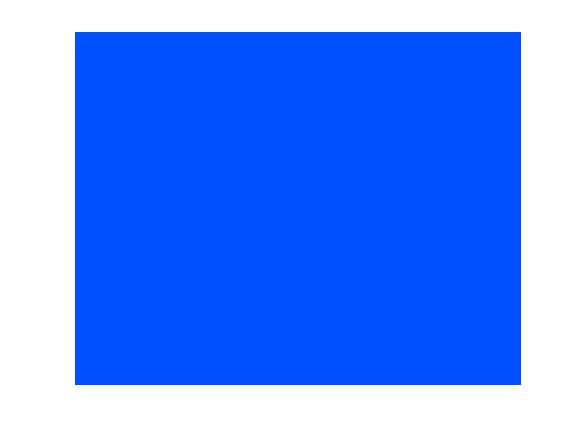}
&
\includegraphics[trim=25mm 17mm 17mm 17mm, clip, width=0.2\textwidth]{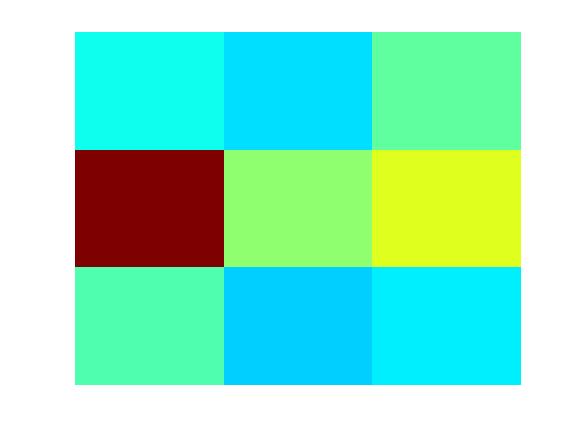}
&
\includegraphics[trim=25mm 17mm 17mm 17mm, clip, width=0.2\textwidth]{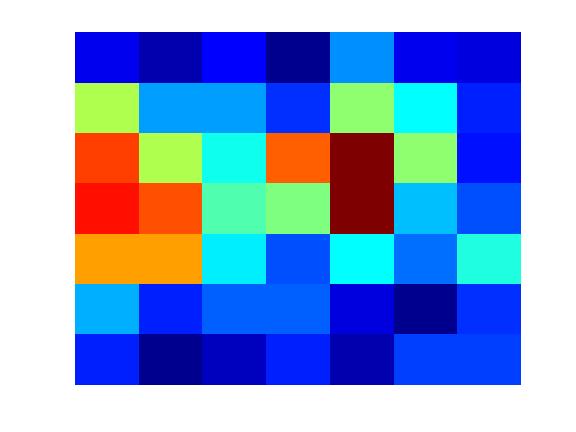}
&
\includegraphics[trim=25mm 17mm 17mm 17mm, clip, width=0.2\textwidth]{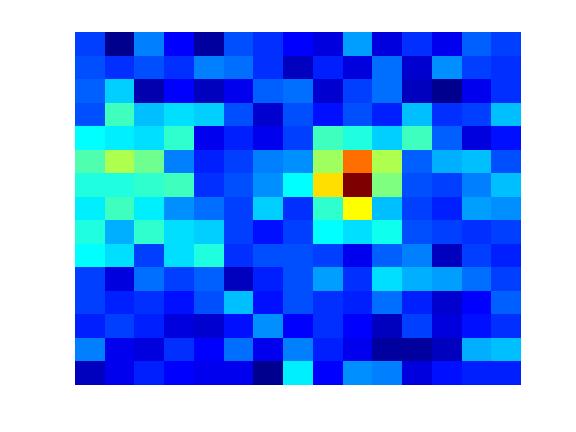}\\
\includegraphics[trim=25mm 17mm 17mm 17mm, clip, width=0.2\textwidth]{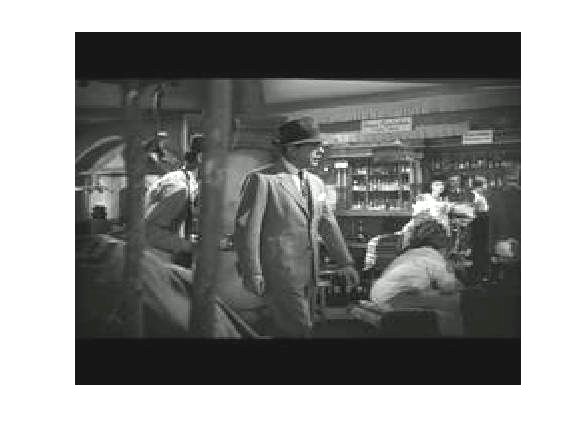}
&
\includegraphics[trim=25mm 17mm 17mm 17mm, clip, width=0.2\textwidth]{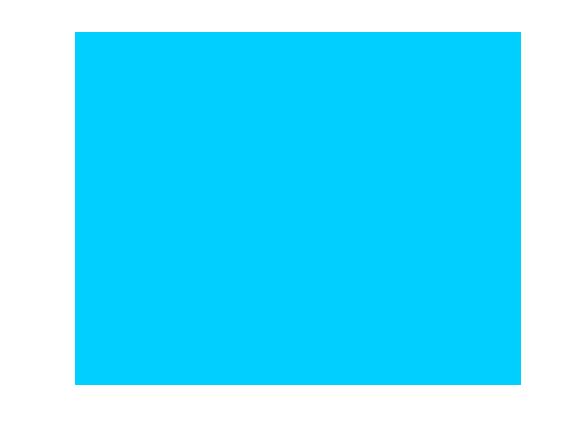}
&
\includegraphics[trim=25mm 17mm 17mm 17mm, clip, width=0.2\textwidth]{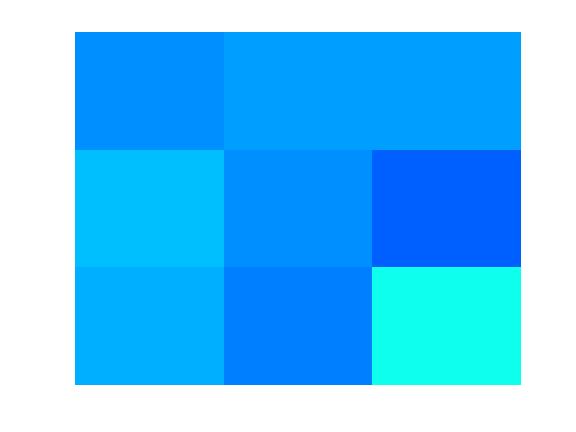}
&
\includegraphics[trim=25mm 17mm 17mm 17mm, clip, width=0.2\textwidth]{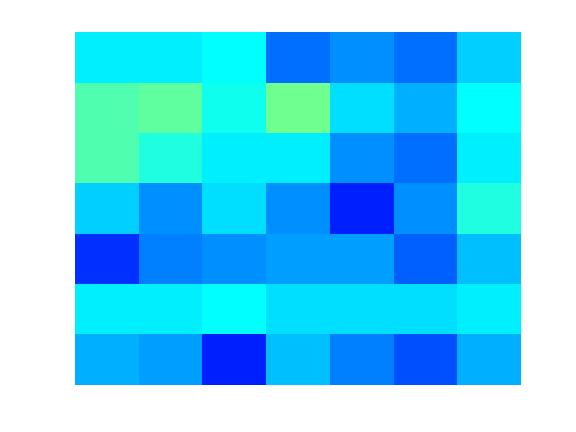}
&
\includegraphics[trim=25mm 17mm 17mm 17mm, clip, width=0.2\textwidth]{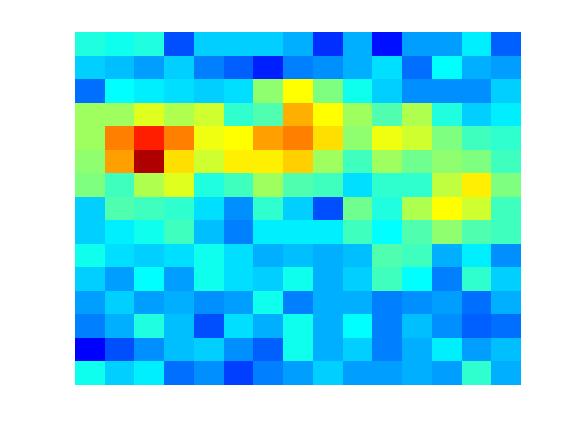}\\
\includegraphics[trim=25mm 17mm 17mm 17mm, clip, width=0.2\textwidth]{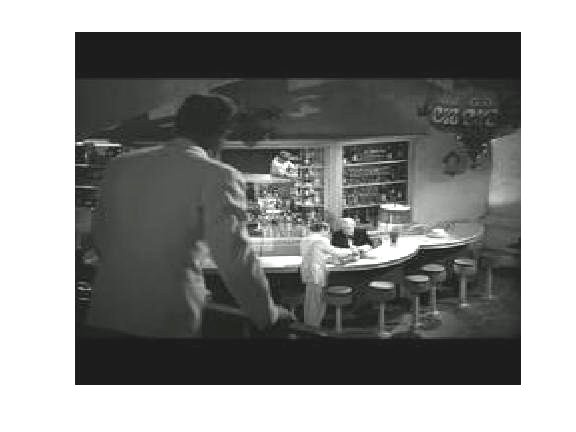}
&
\includegraphics[trim=25mm 17mm 17mm 17mm, clip, width=0.2\textwidth]{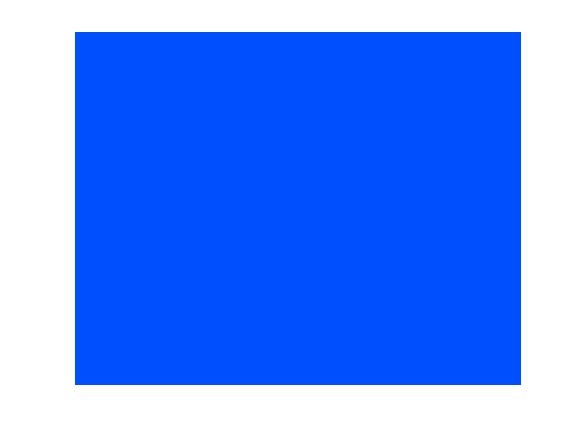}
&
\includegraphics[trim=25mm 17mm 17mm 17mm, clip, width=0.2\textwidth]{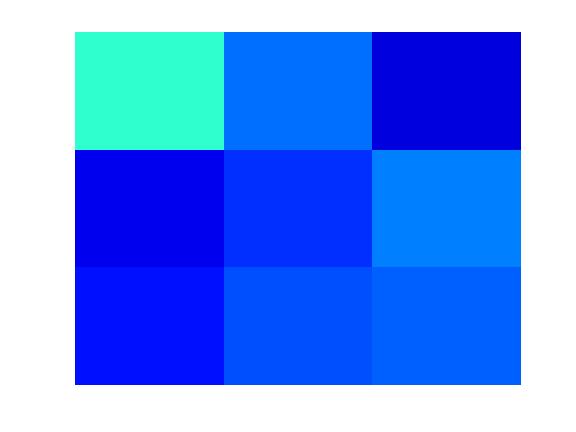}
&
\includegraphics[trim=25mm 17mm 17mm 17mm, clip, width=0.2\textwidth]{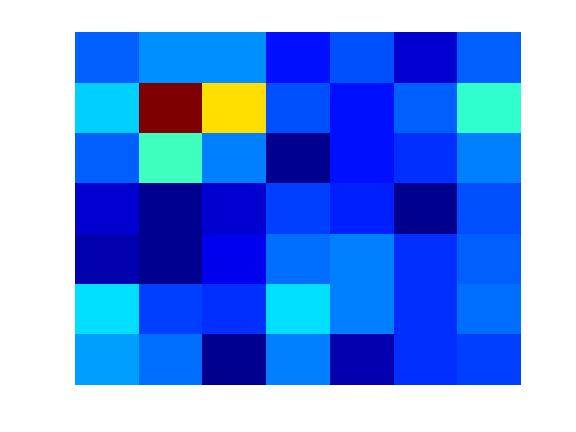}
&
\includegraphics[trim=25mm 17mm 17mm 17mm, clip, width=0.2\textwidth]{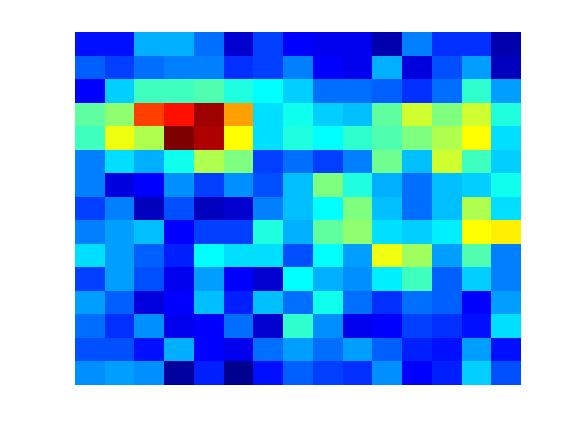}\\
\includegraphics[trim=25mm 17mm 17mm 17mm, clip, width=0.2\textwidth]{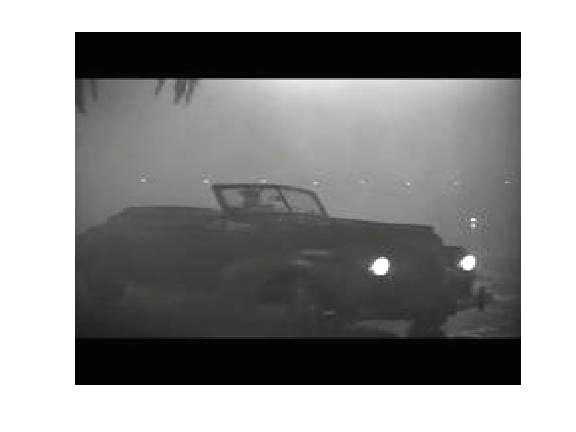}
&
\includegraphics[trim=25mm 17mm 17mm 17mm, clip, width=0.2\textwidth]{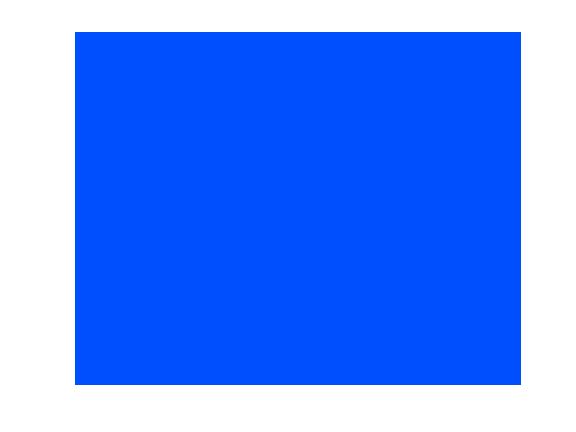}
&
\includegraphics[trim=25mm 17mm 17mm 17mm, clip, width=0.2\textwidth]{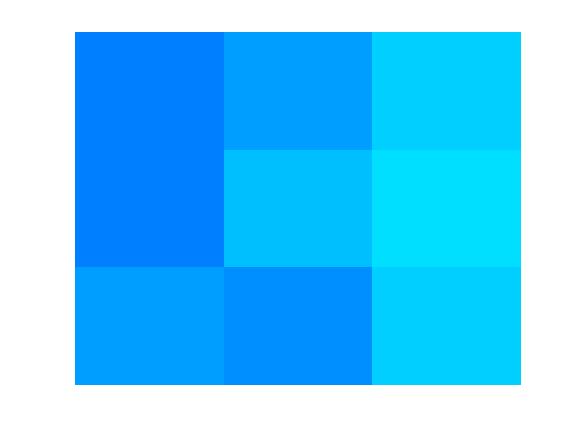}
&
\includegraphics[trim=25mm 17mm 17mm 17mm, clip, width=0.2\textwidth]{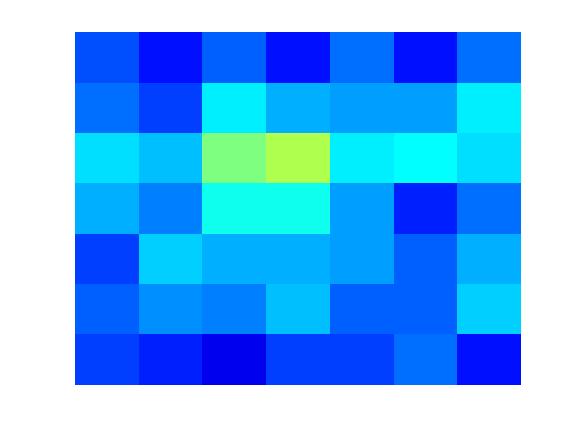}
&
\includegraphics[trim=25mm 17mm 17mm 17mm, clip, width=0.2\textwidth]{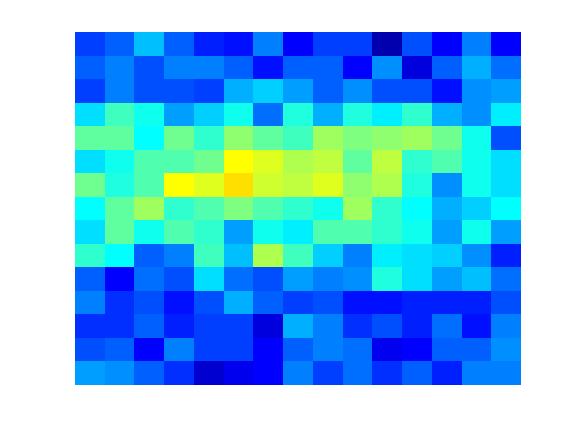}\\
\includegraphics[trim=25mm 17mm 17mm 17mm, clip, width=0.2\textwidth]{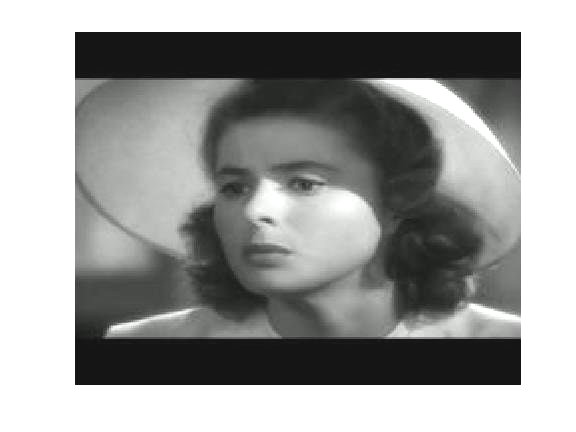}
&
\includegraphics[trim=25mm 17mm 17mm 17mm, clip, width=0.2\textwidth]{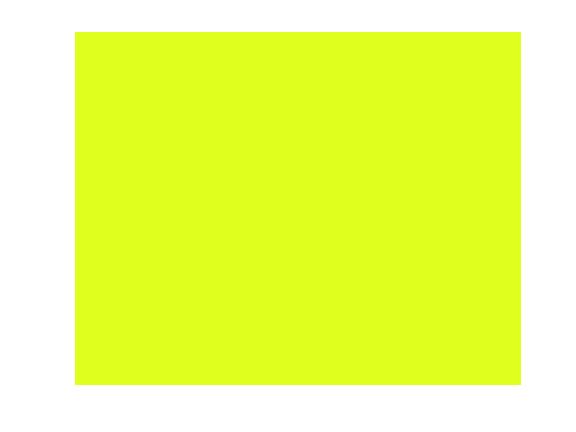}
&
\includegraphics[trim=25mm 17mm 17mm 17mm, clip, width=0.2\textwidth]{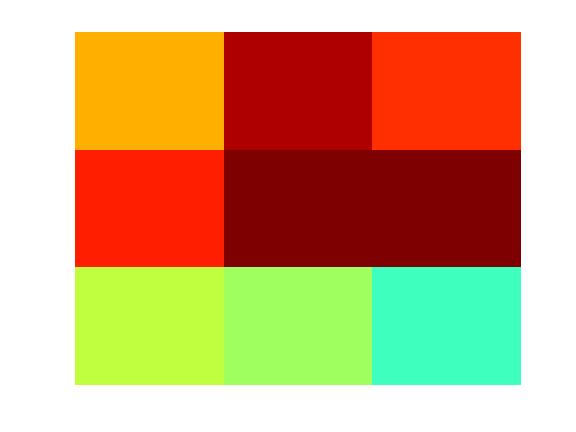}
&
\includegraphics[trim=25mm 17mm 17mm 17mm, clip, width=0.2\textwidth]{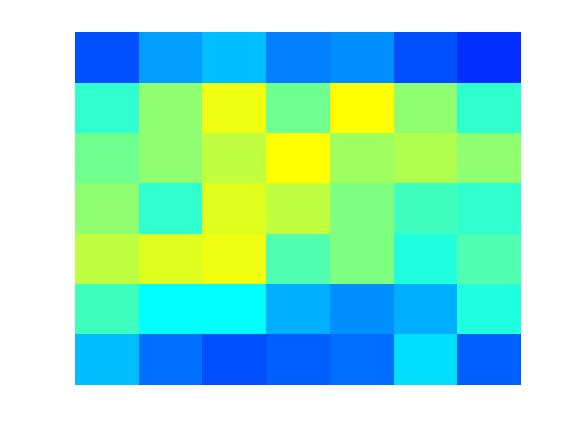}
&
\includegraphics[trim=25mm 17mm 17mm 17mm, clip, width=0.2\textwidth]{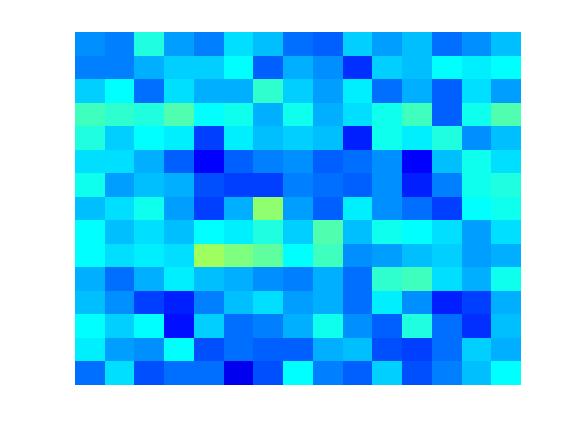}\\
\end{tabular}
\end{center}
\caption{Qualitative results for our Global model. Columns from the left to the right correspond to original images and the output of the Global model at different resolutions. Red color corresponds to cells with high scores for the ``head'' class, blue color indicates cells with low scores for the ``head'' class.}
\label{fig:globalModel_qual_examples}
\end{figure*}

\begin{figure*}
\begin{tabular}{@{}c@{\,}c@{\;\;\;\;}c@{\,}c}
Local & Pairwise & Local & Pairwise \\[-0.0cm]
\includegraphics[trim = 0mm 0mm 6mm 0mm, clip, width=0.24\linewidth]{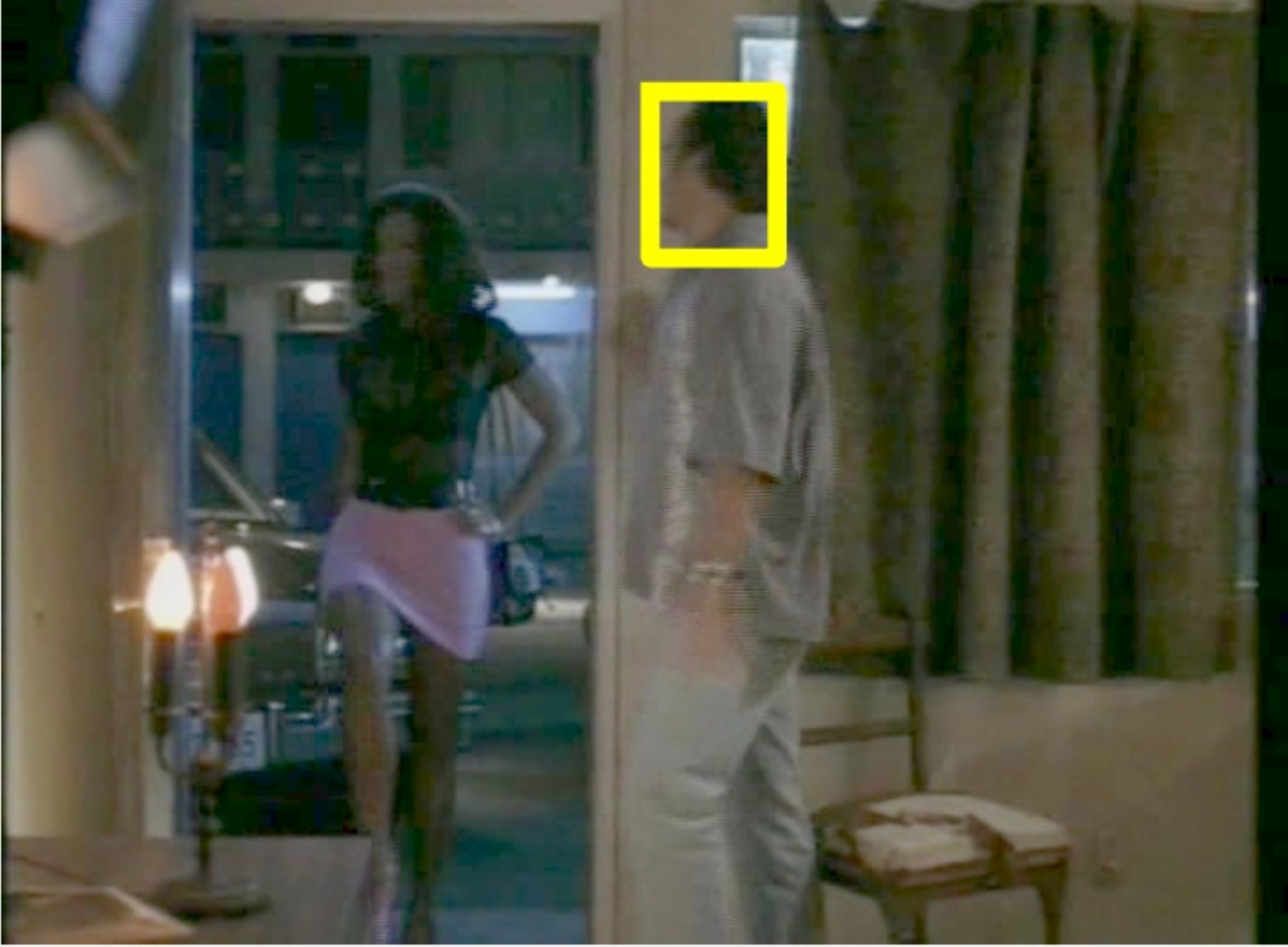}
&
\includegraphics[trim = 0mm 0mm 6mm 0mm, clip, width=0.24\linewidth]{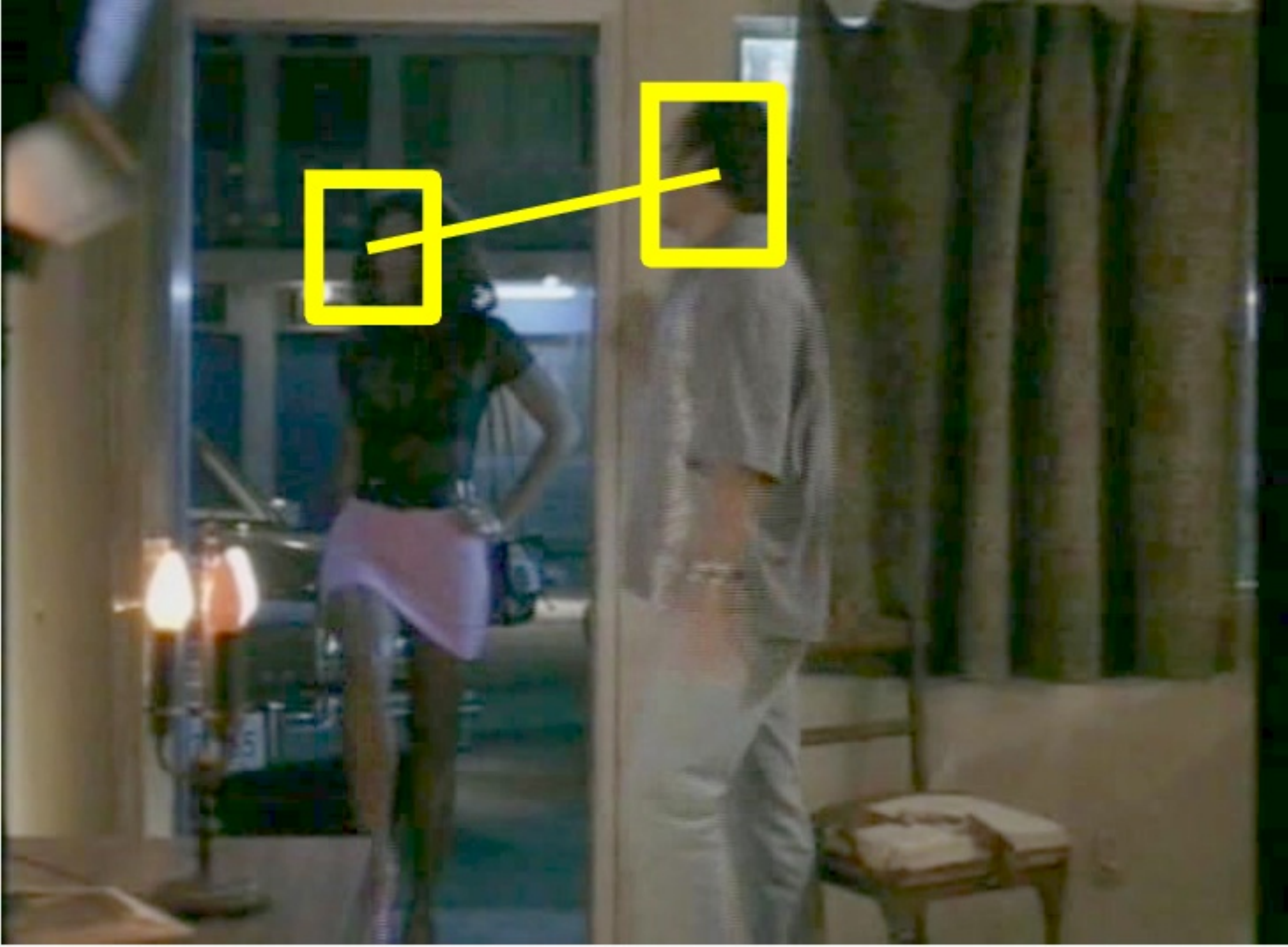}
&
\includegraphics[trim = 0mm 0mm 6mm 0mm, clip, width=0.24\linewidth]{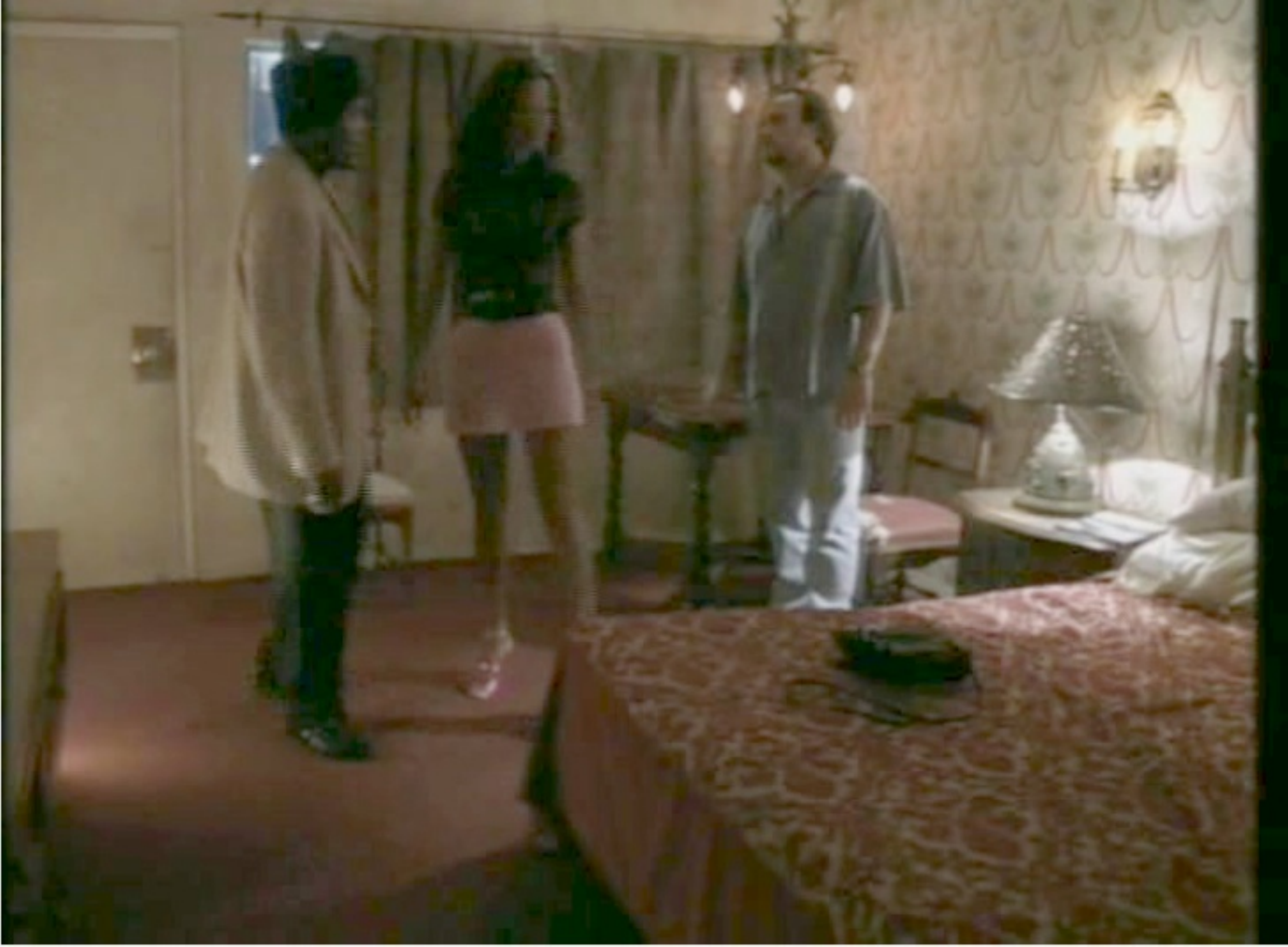}
&
\includegraphics[trim = 0mm 0mm 6mm 0mm, clip, width=0.24\linewidth]{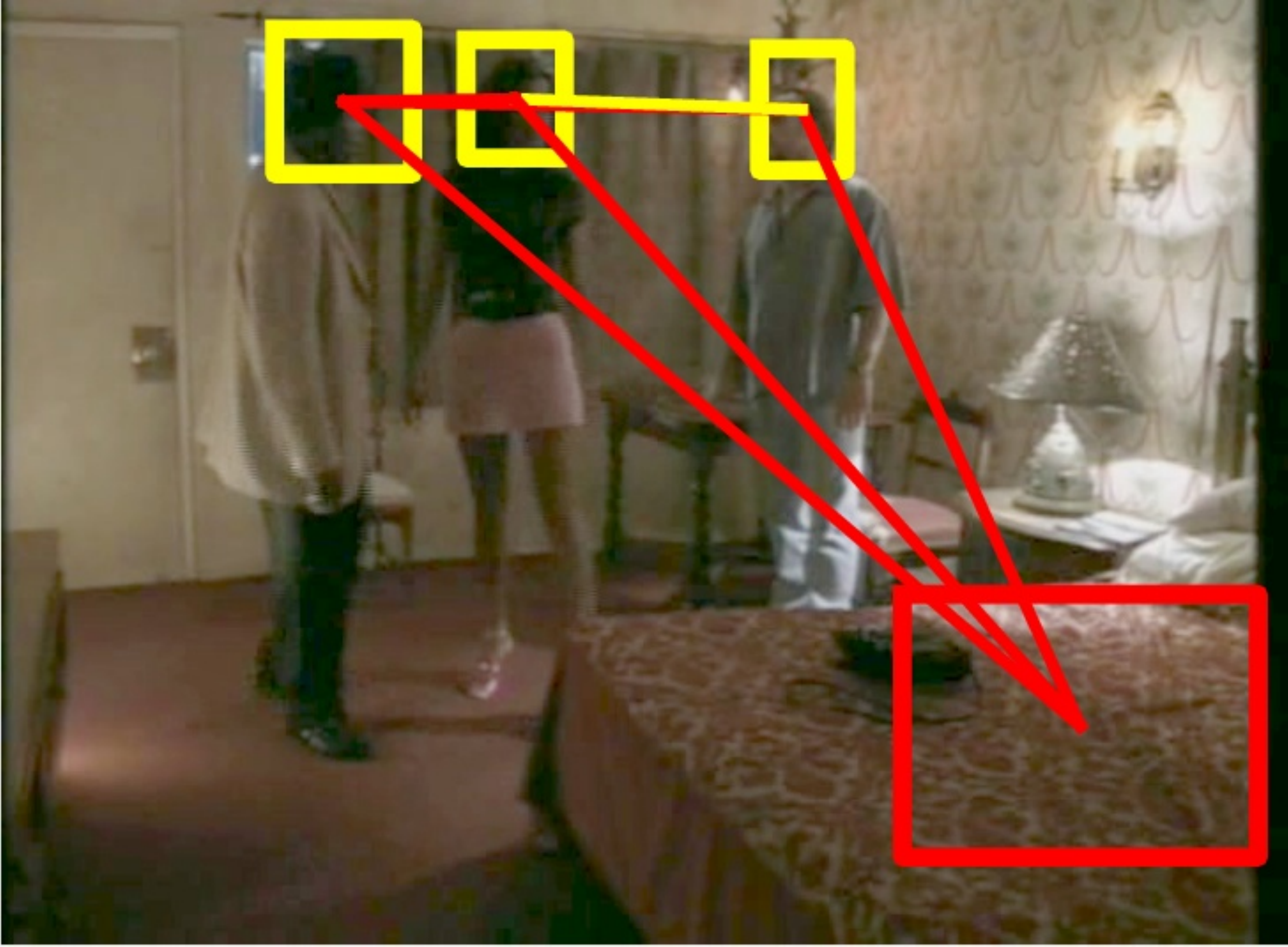} \\[-0.05cm]
\includegraphics[trim = 0mm 0mm 6mm 0mm, clip, width=0.24\linewidth]{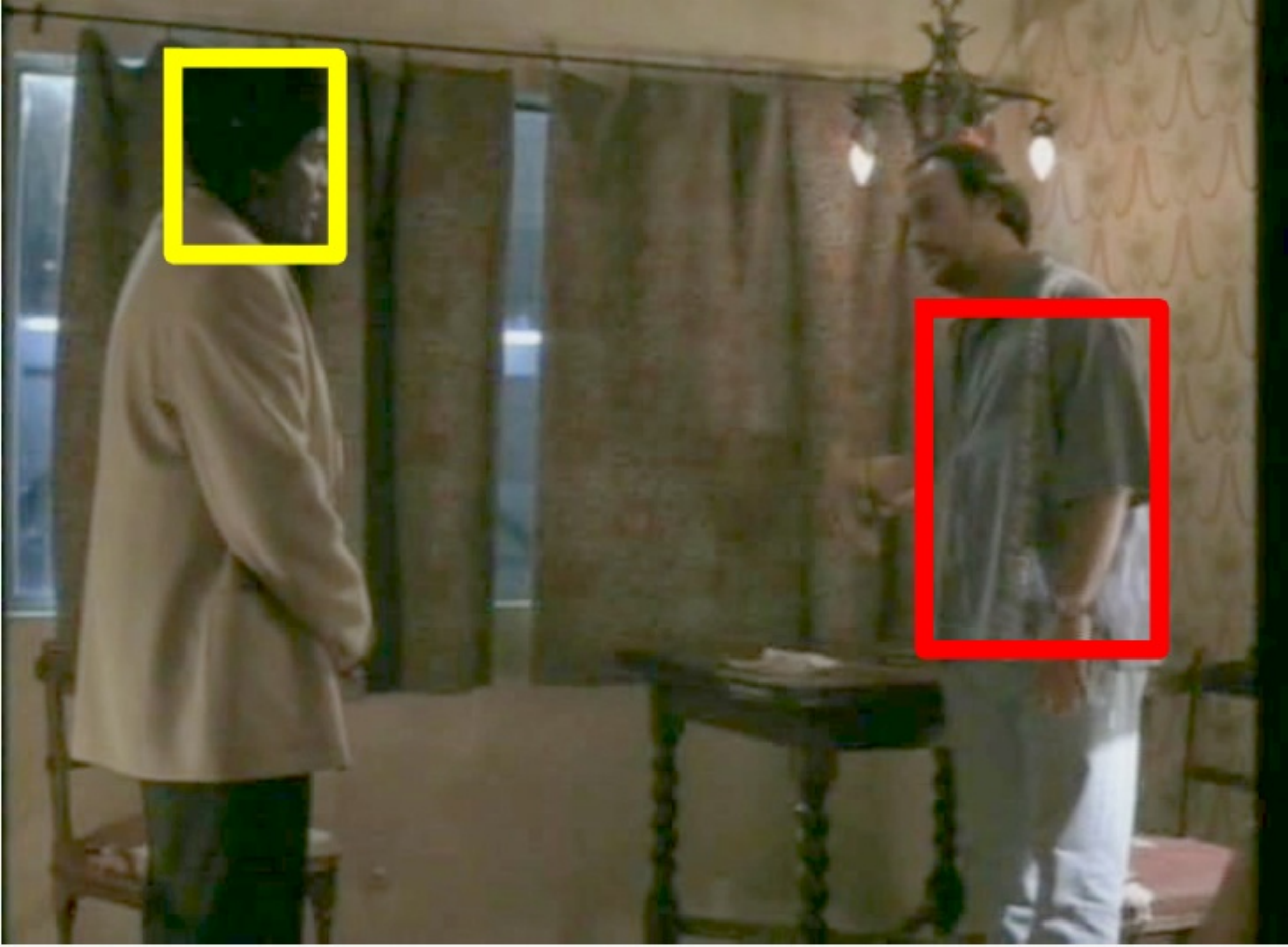}
&
\includegraphics[trim = 0mm 0mm 6mm 0mm, clip, width=0.24\linewidth]{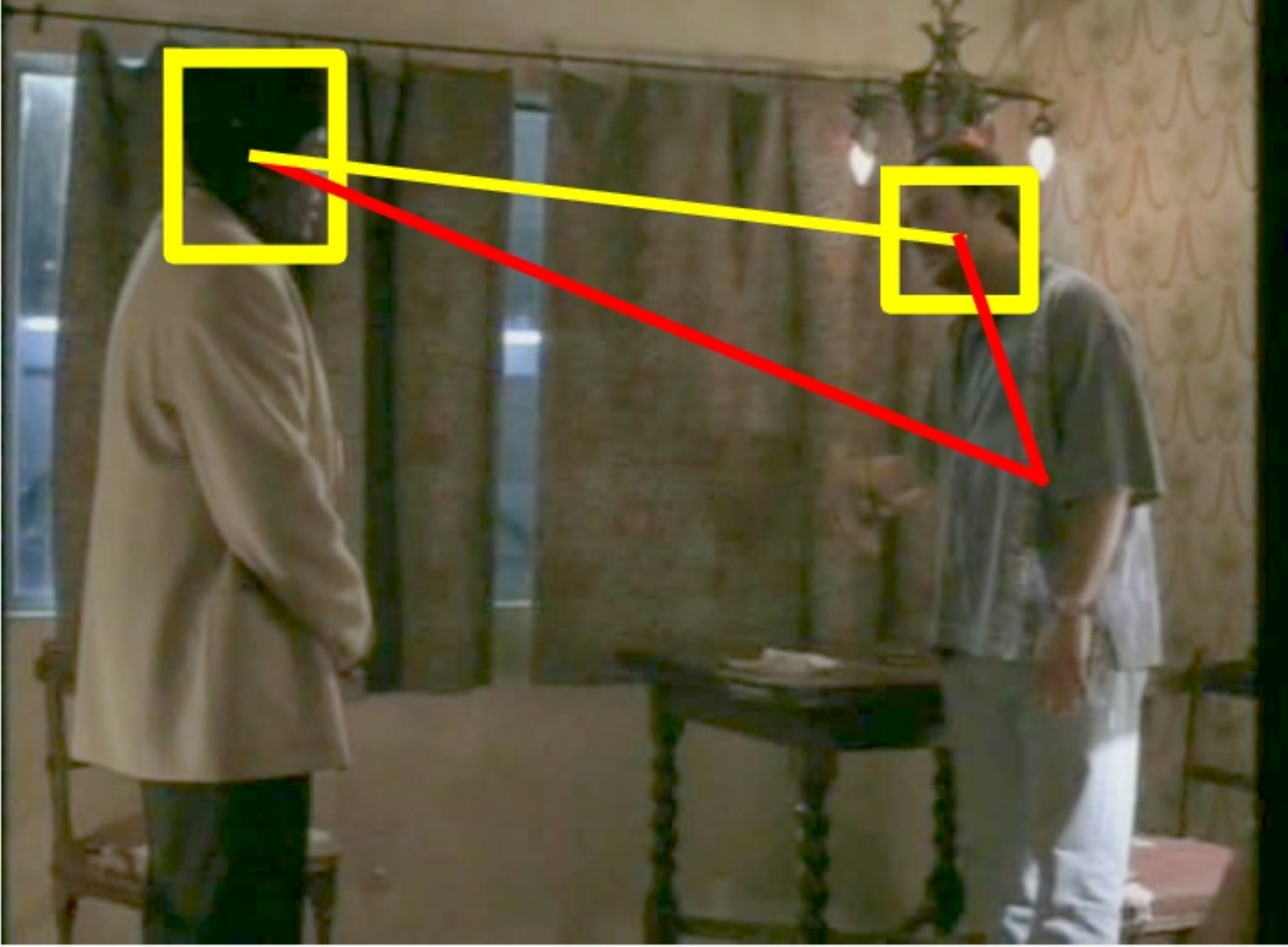}
&
\includegraphics[trim = 0mm 0mm 6mm 0mm, clip, width=0.24\linewidth]{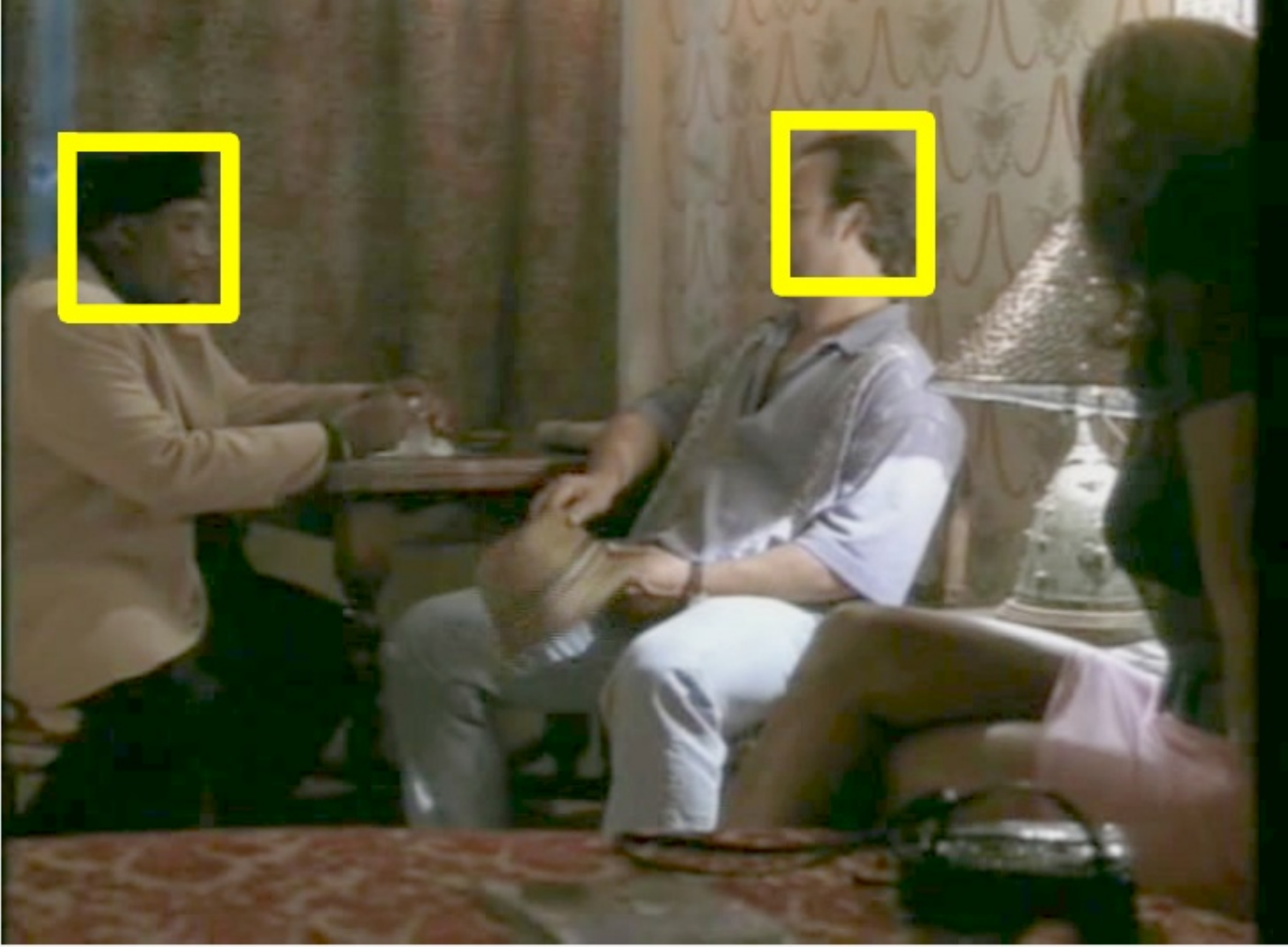}
&
\includegraphics[trim = 0mm 0mm 6mm 0mm, clip, width=0.24\linewidth]{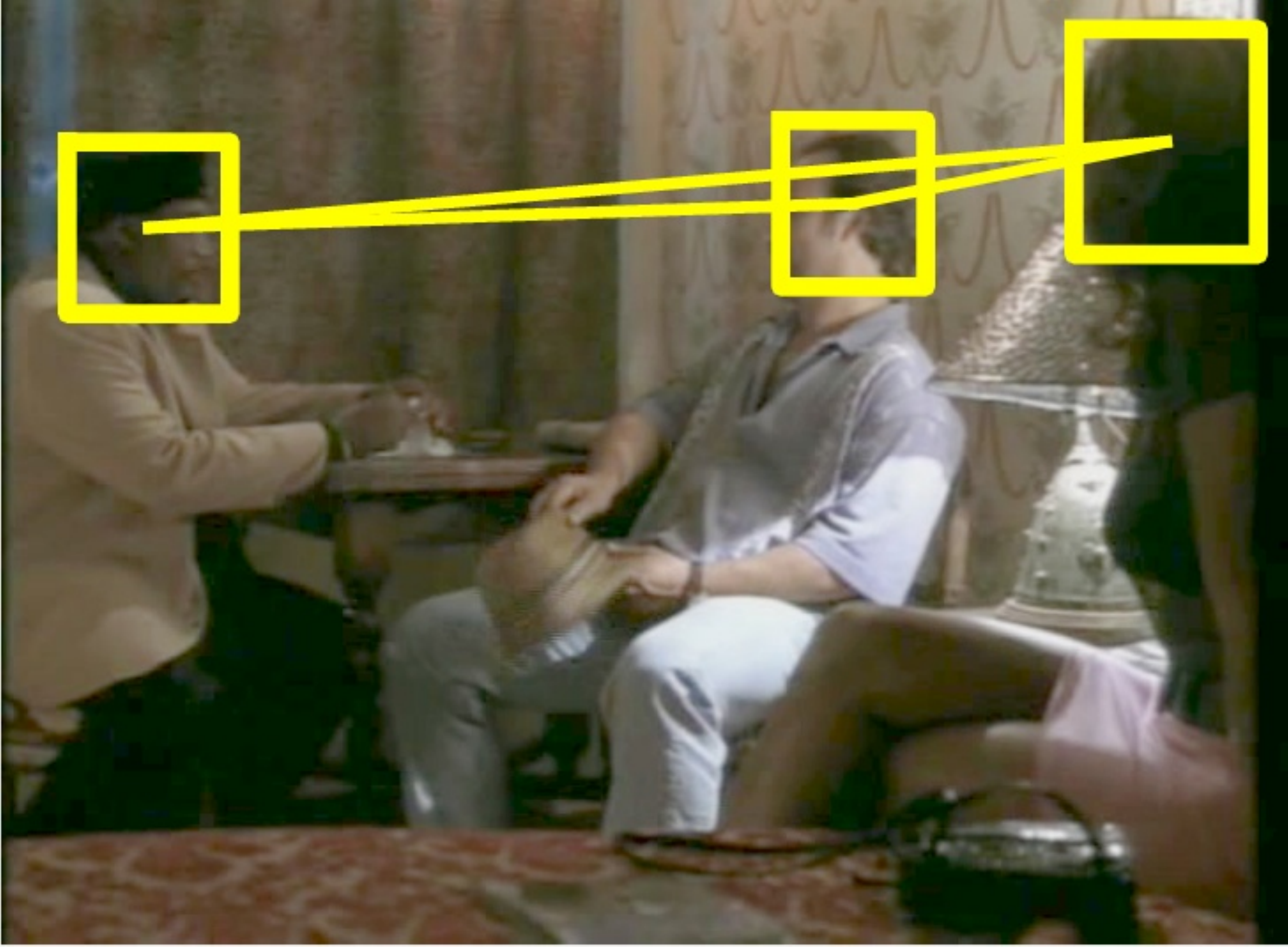} \\[-0.05cm]
\includegraphics[trim = 0mm 0mm 6mm 0mm, clip, width=0.24\linewidth]{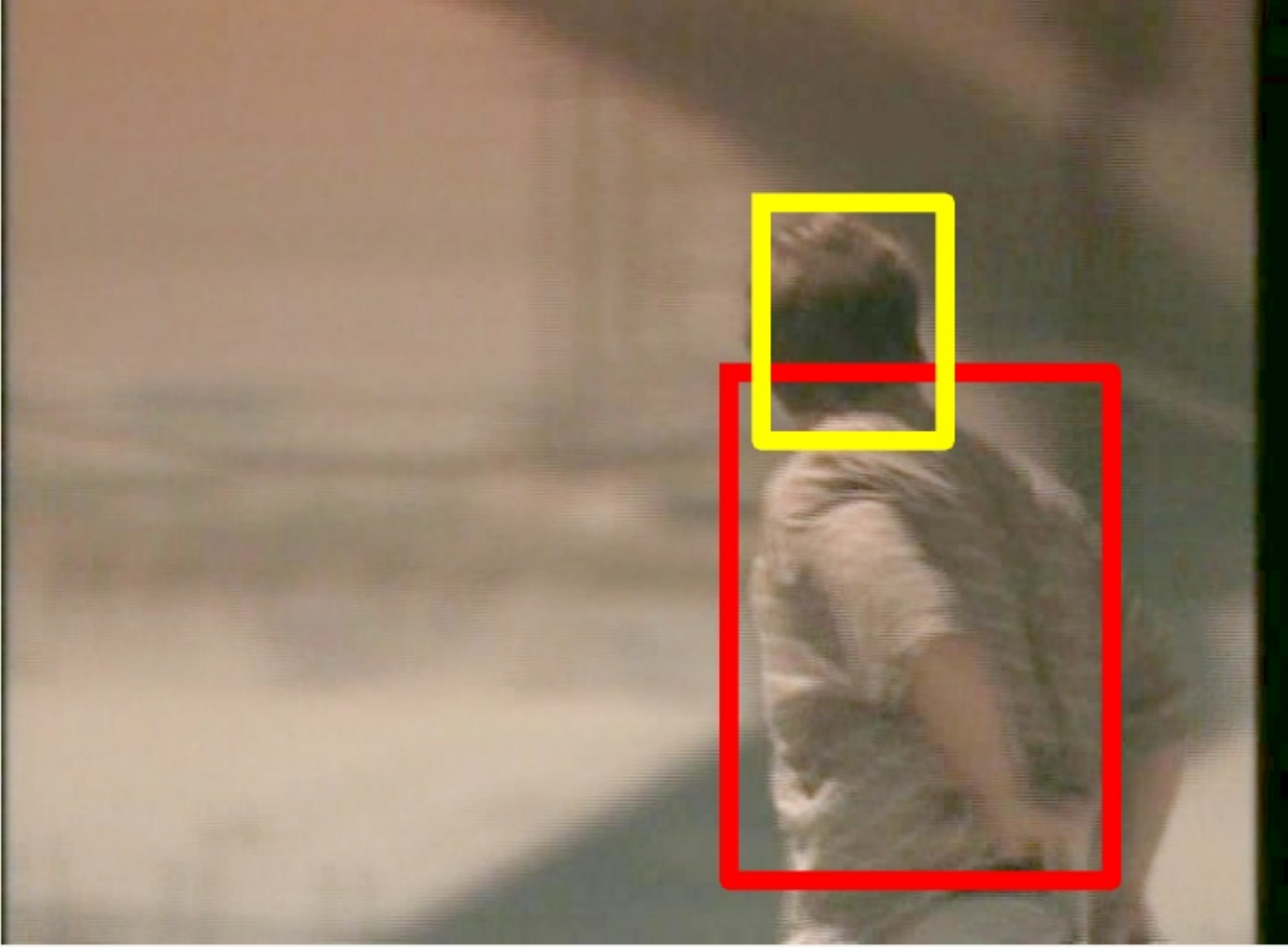}
&
\includegraphics[trim = 0mm 0mm 6mm 0mm, clip, width=0.24\linewidth]{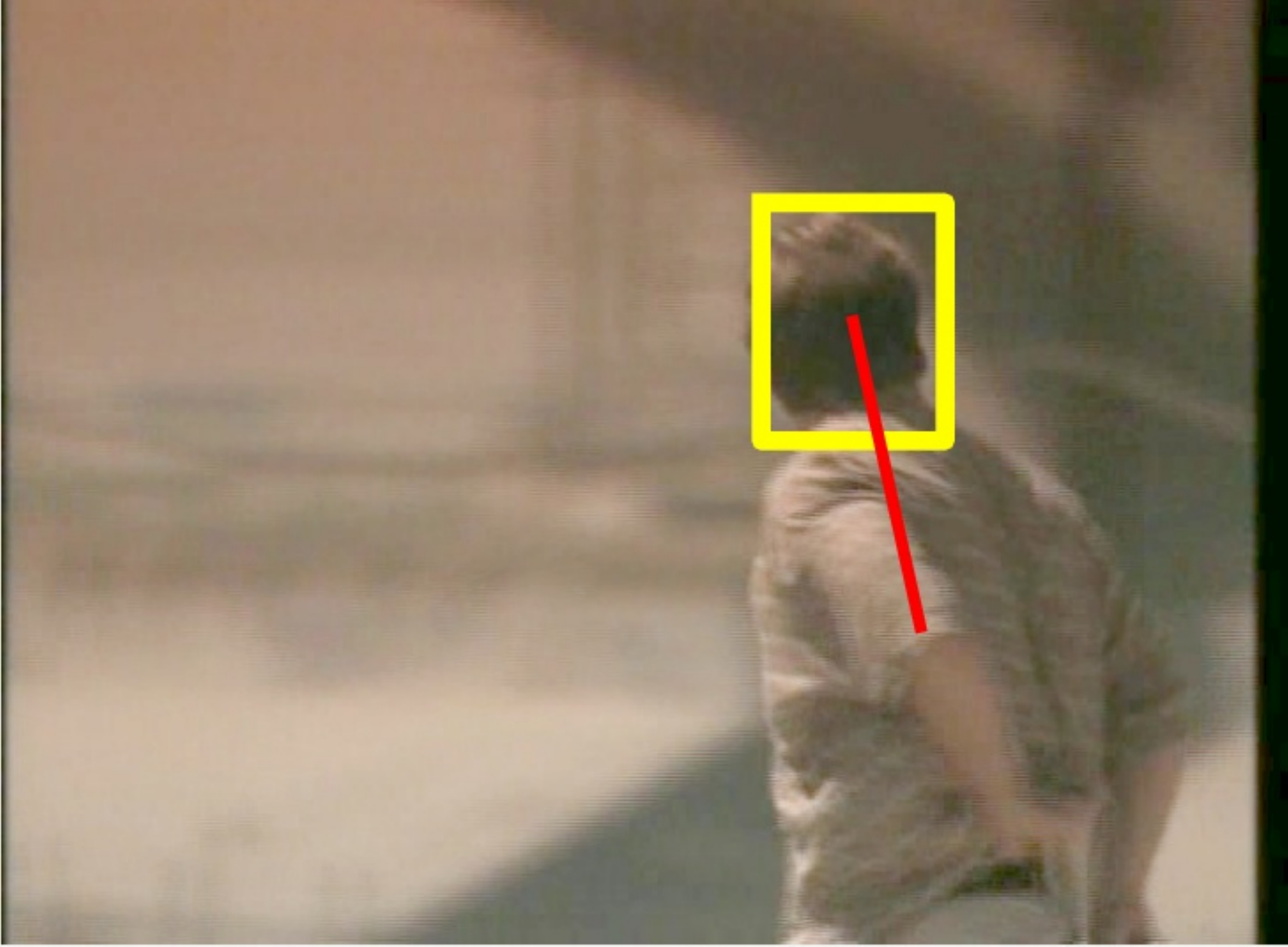}
&
\includegraphics[trim = 0mm 0mm 6mm 0mm, clip, width=0.24\linewidth]{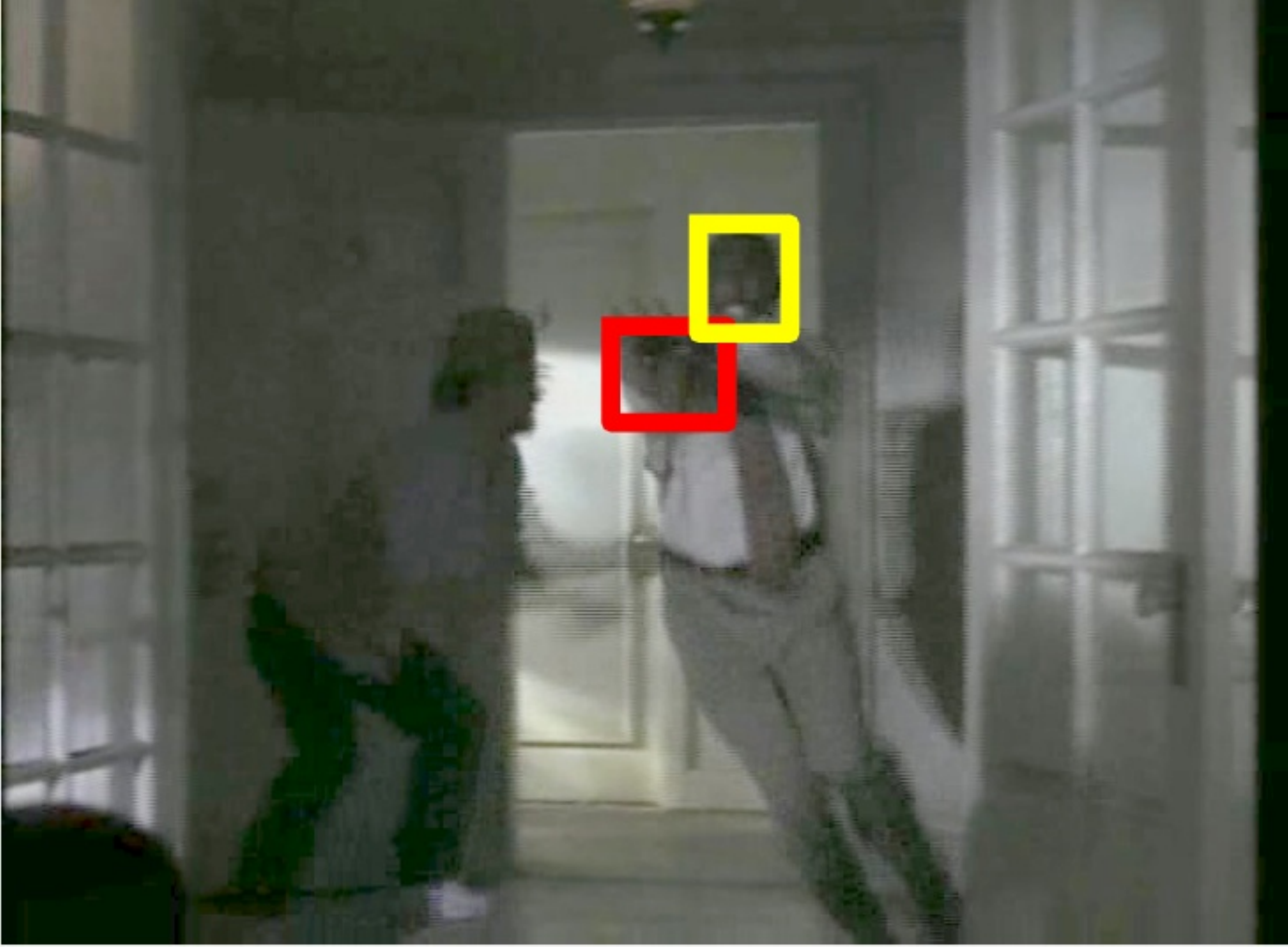}
&
\includegraphics[trim = 0mm 0mm 6mm 0mm, clip, width=0.24\linewidth]{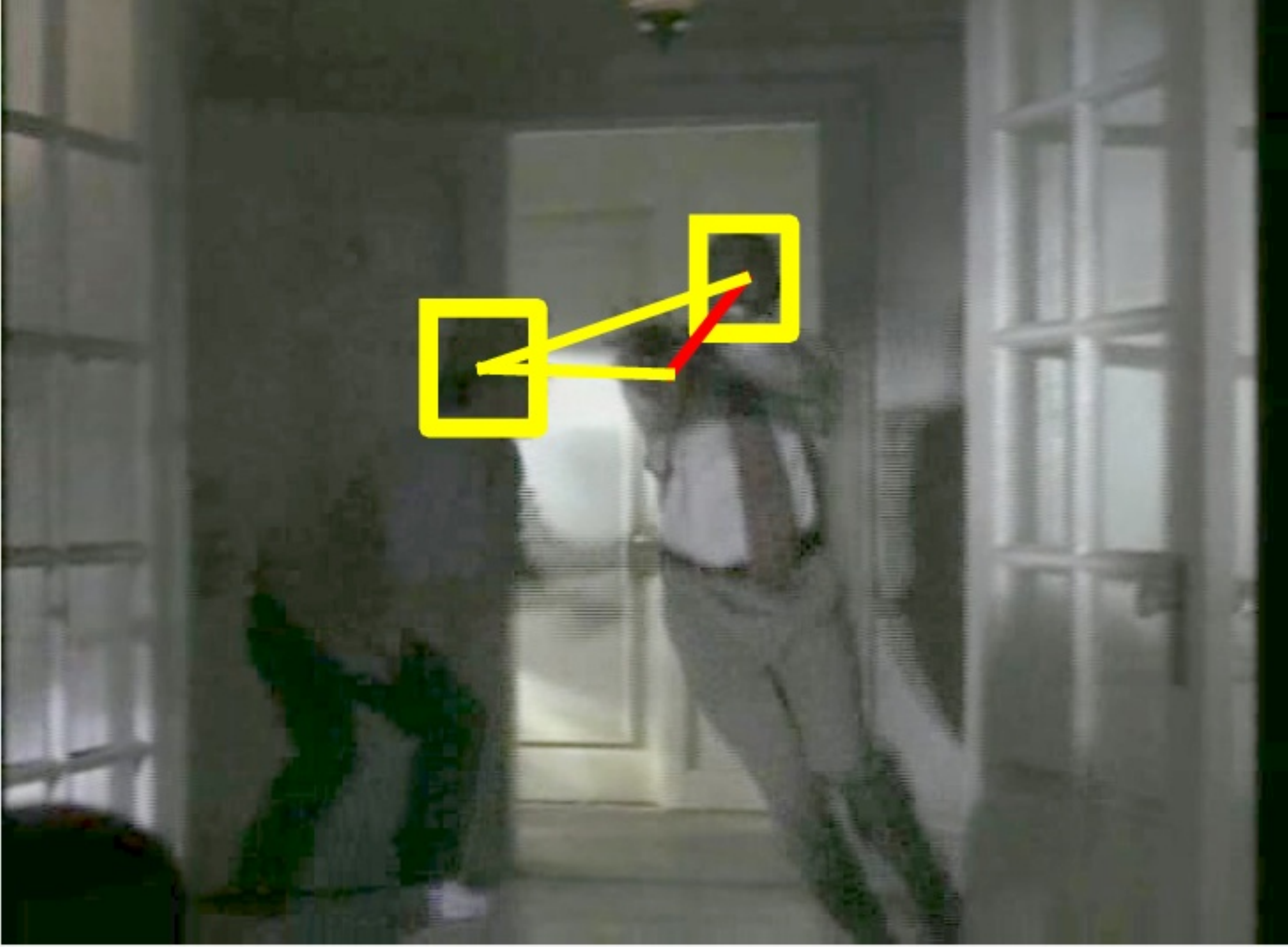} \\[-0.05cm]
\includegraphics[trim = 0mm 0mm 0mm 0mm, clip, width=0.24\linewidth]{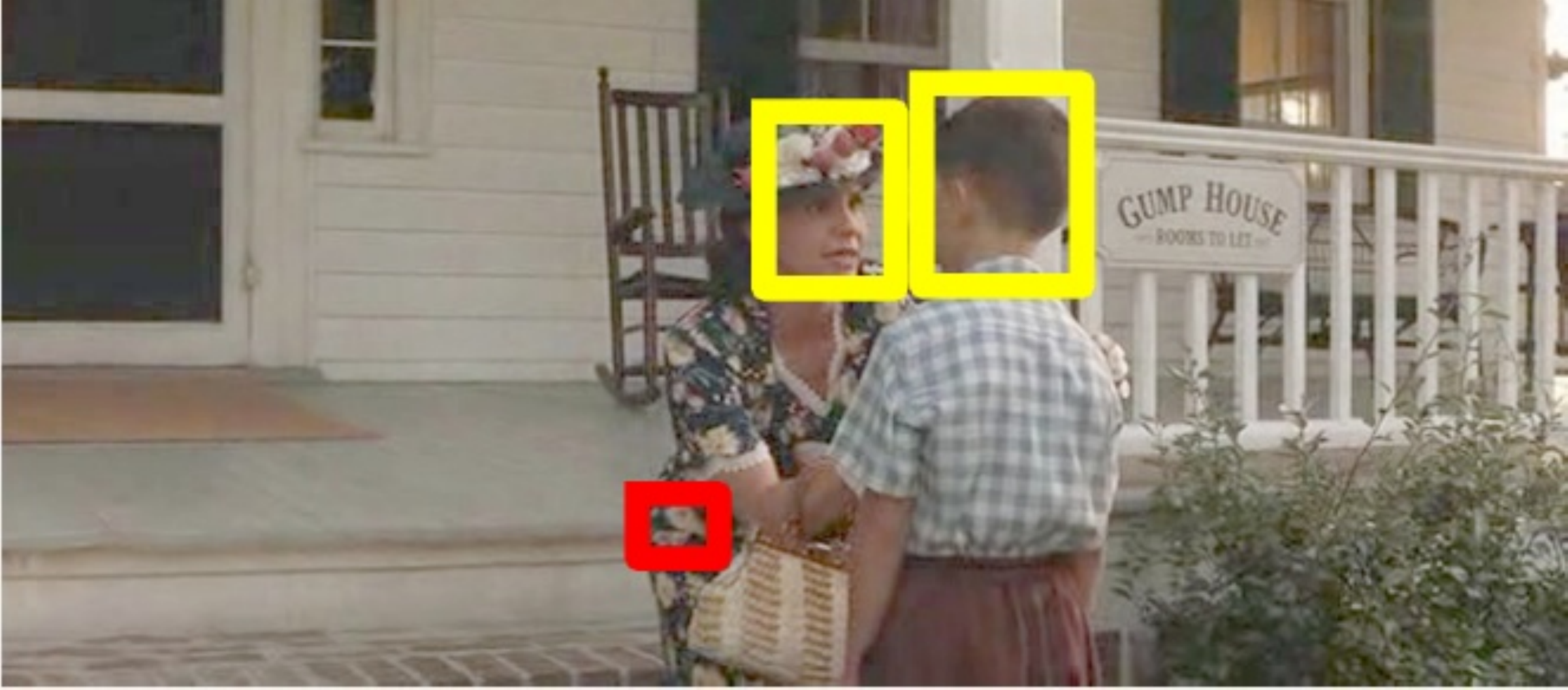}
&
\includegraphics[trim = 0mm 0mm 0mm 0mm, clip, width=0.24\linewidth]{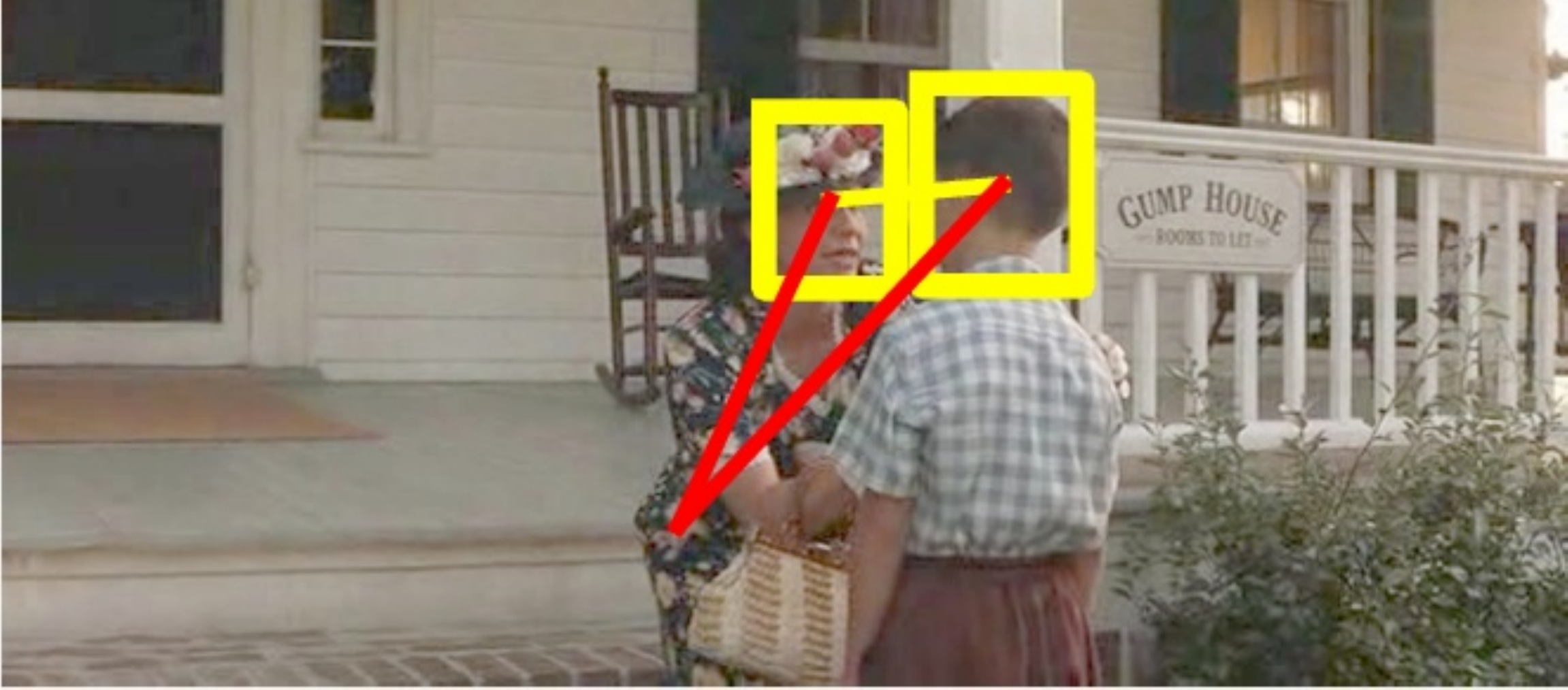}
&
\includegraphics[trim = 0mm 0mm 0mm 0mm, clip, width=0.24\linewidth]{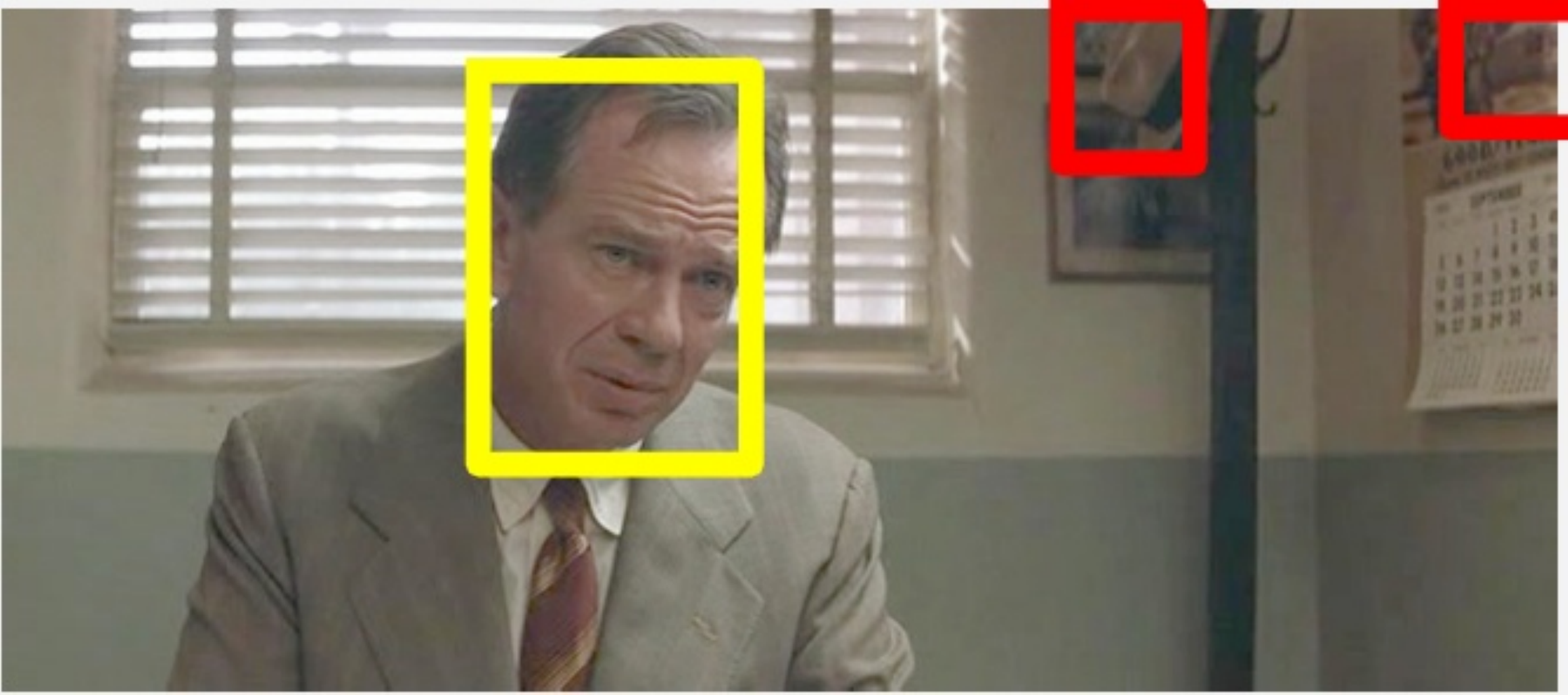}
&
\includegraphics[trim = 0mm 0mm 0mm 0mm, clip, width=0.24\linewidth]{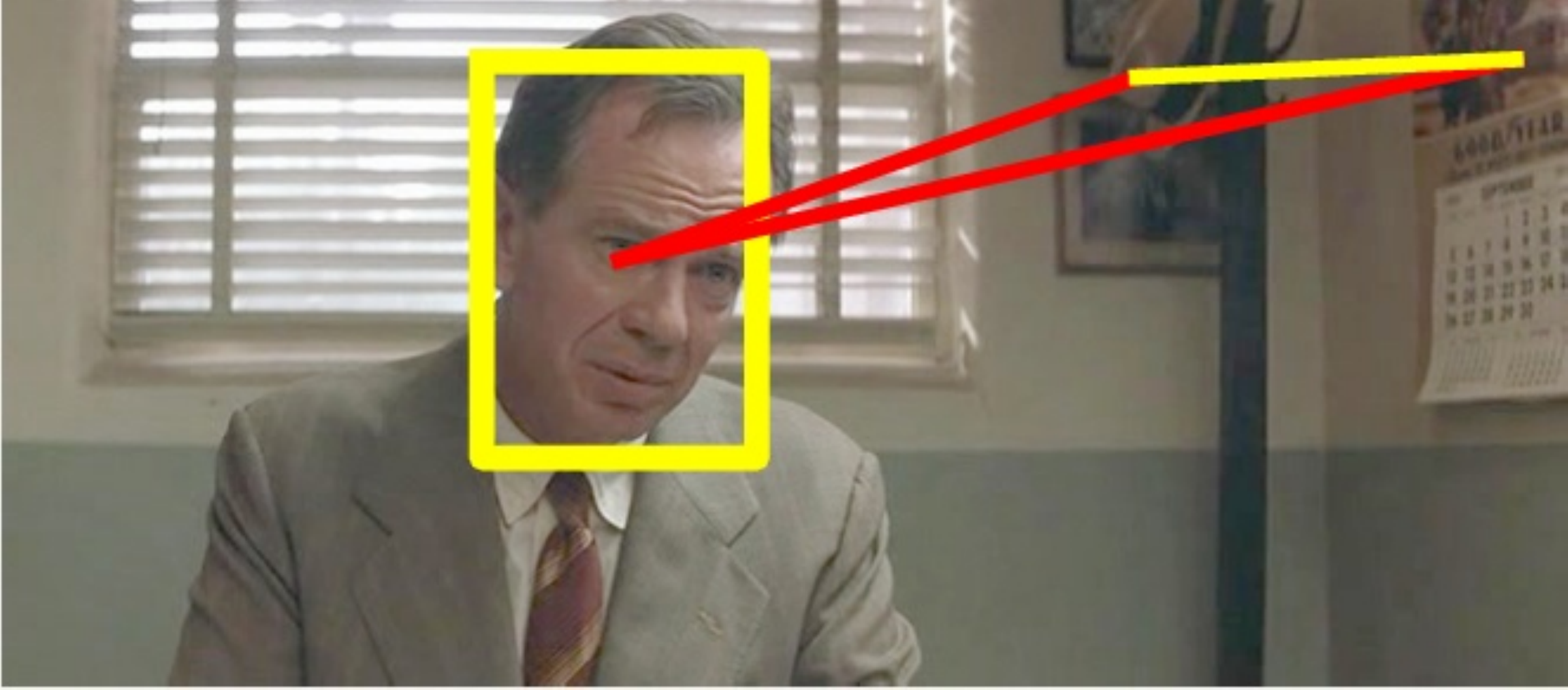}
 \\[-0.05cm]
\includegraphics[trim = 0mm 0mm 0mm 0mm, clip, width=0.24\linewidth]{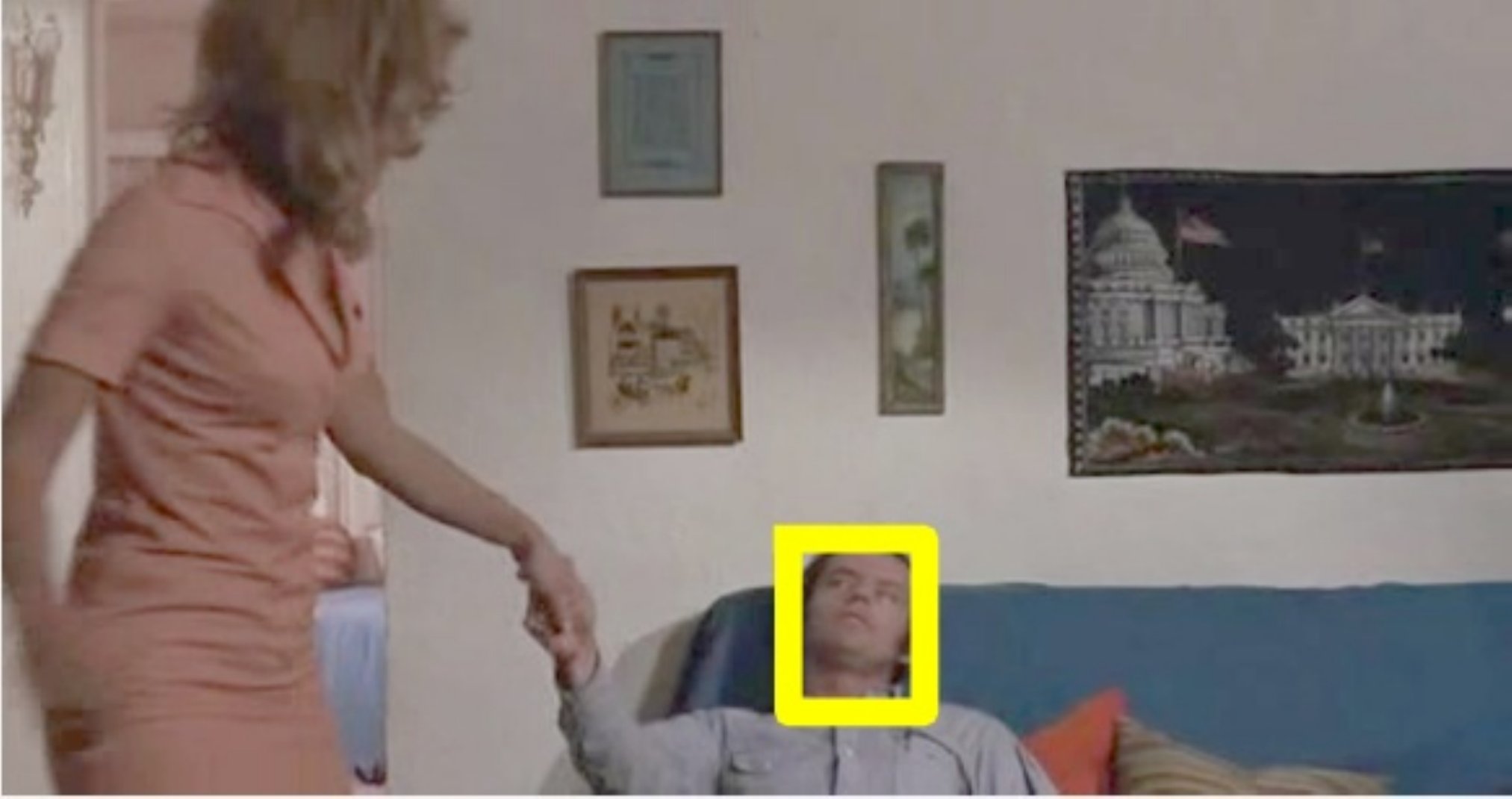}
&
\includegraphics[trim = 0mm 0mm 0mm 0mm, clip, width=0.24\linewidth]{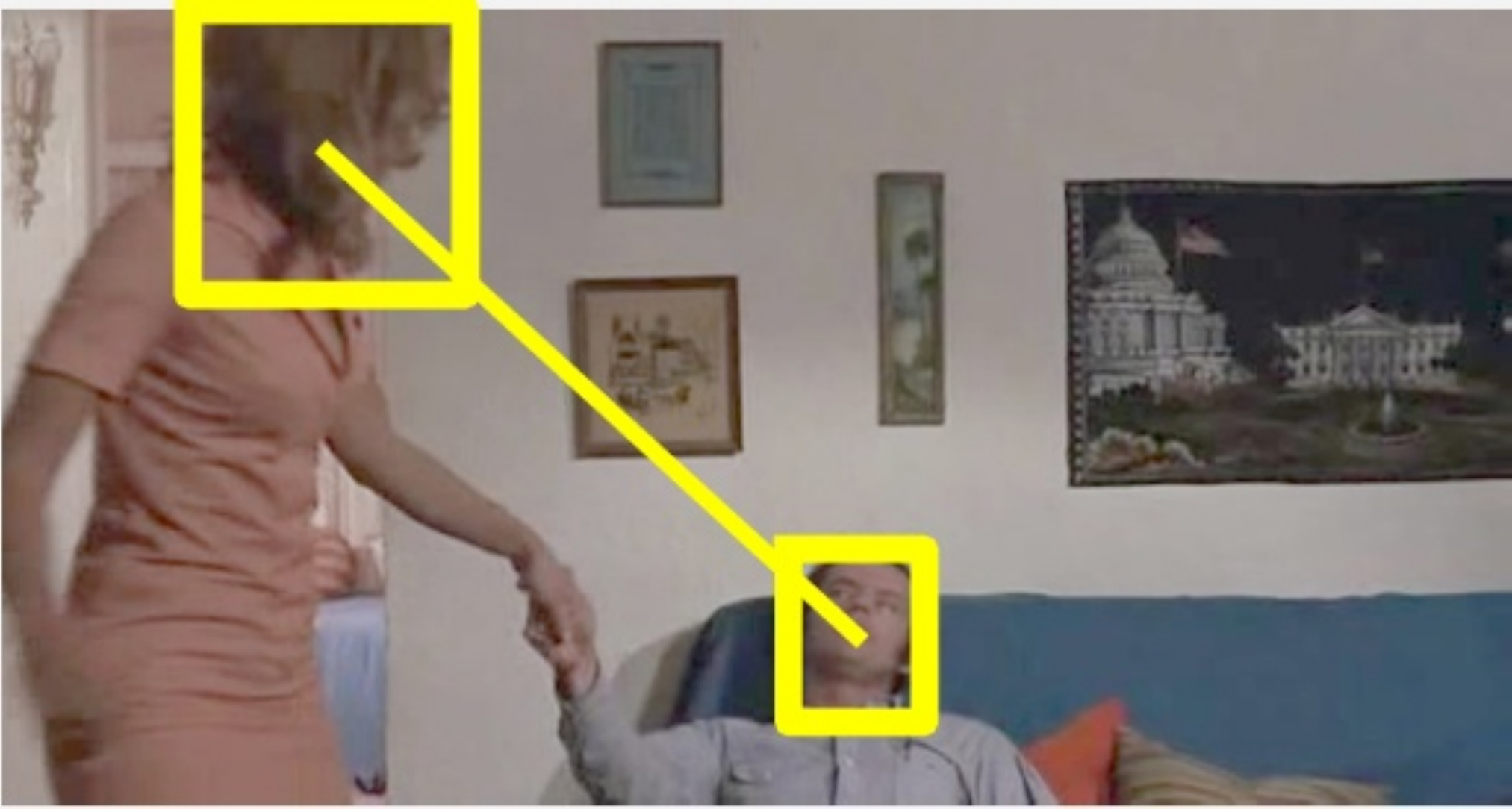}
&
\includegraphics[trim = 0mm 0mm 0mm 0mm, clip, width=0.24\linewidth]{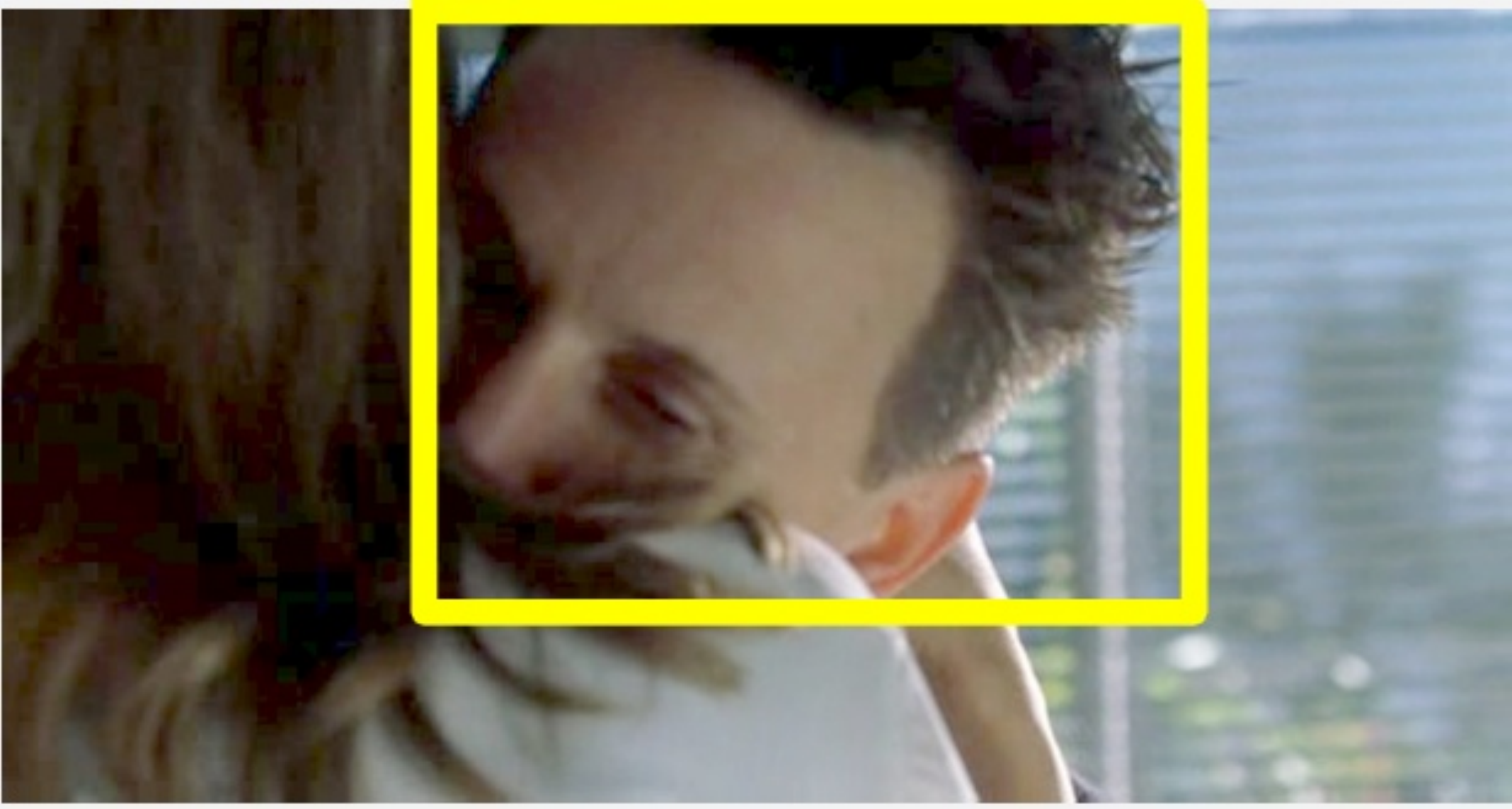}
&
\includegraphics[trim = 0mm 0mm 0mm 0mm, clip, width=0.24\linewidth]{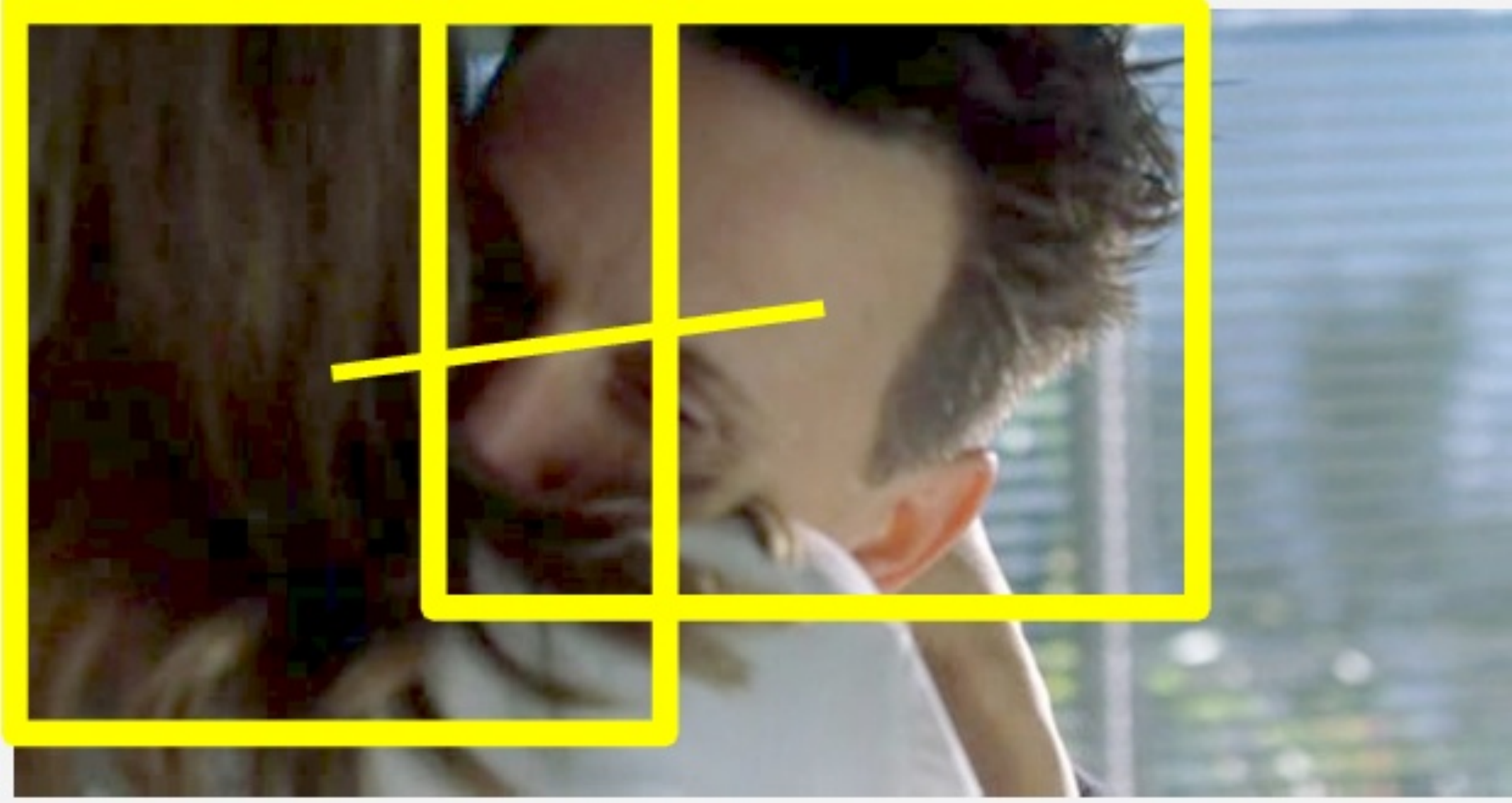}
\\[-0.05cm]
\includegraphics[trim = 2mm 24mm 4mm 23mm, clip, width=0.24\linewidth]{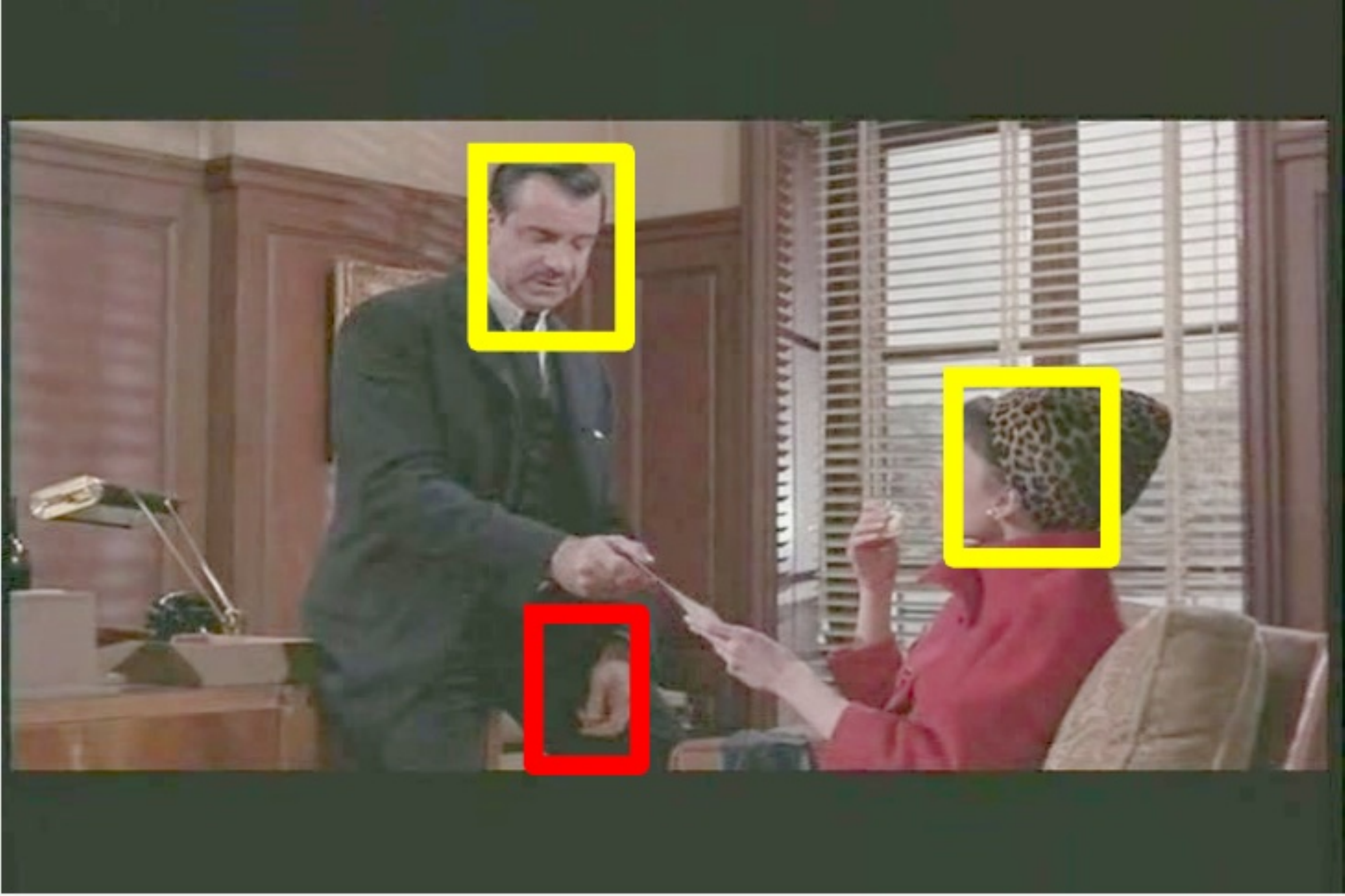}
&
\includegraphics[trim = 2mm 24mm 4mm 23mm, clip, width=0.24\linewidth]{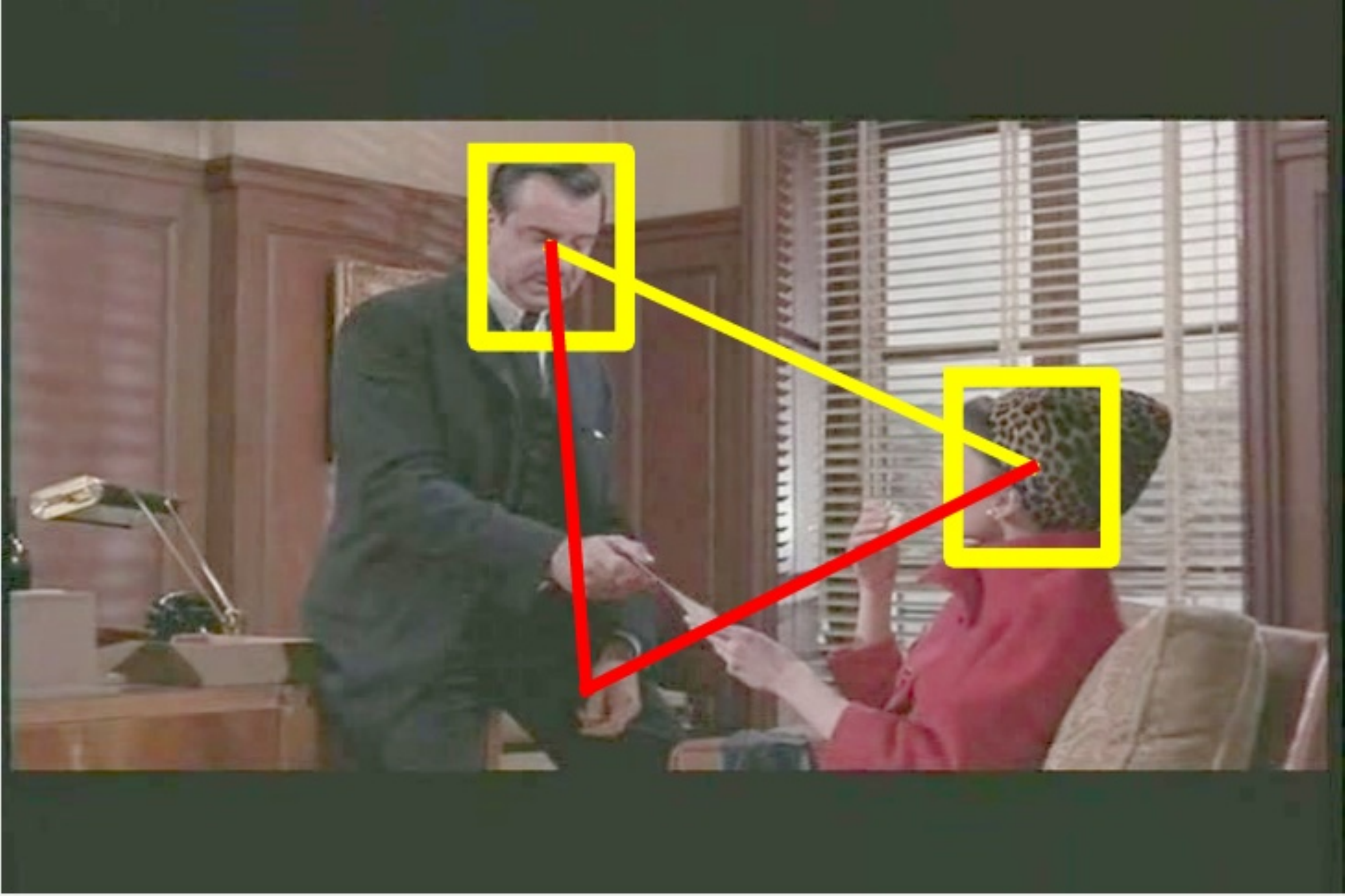}
&
\includegraphics[trim = 2mm 24mm 4mm 23mm, clip, width=0.24\linewidth]{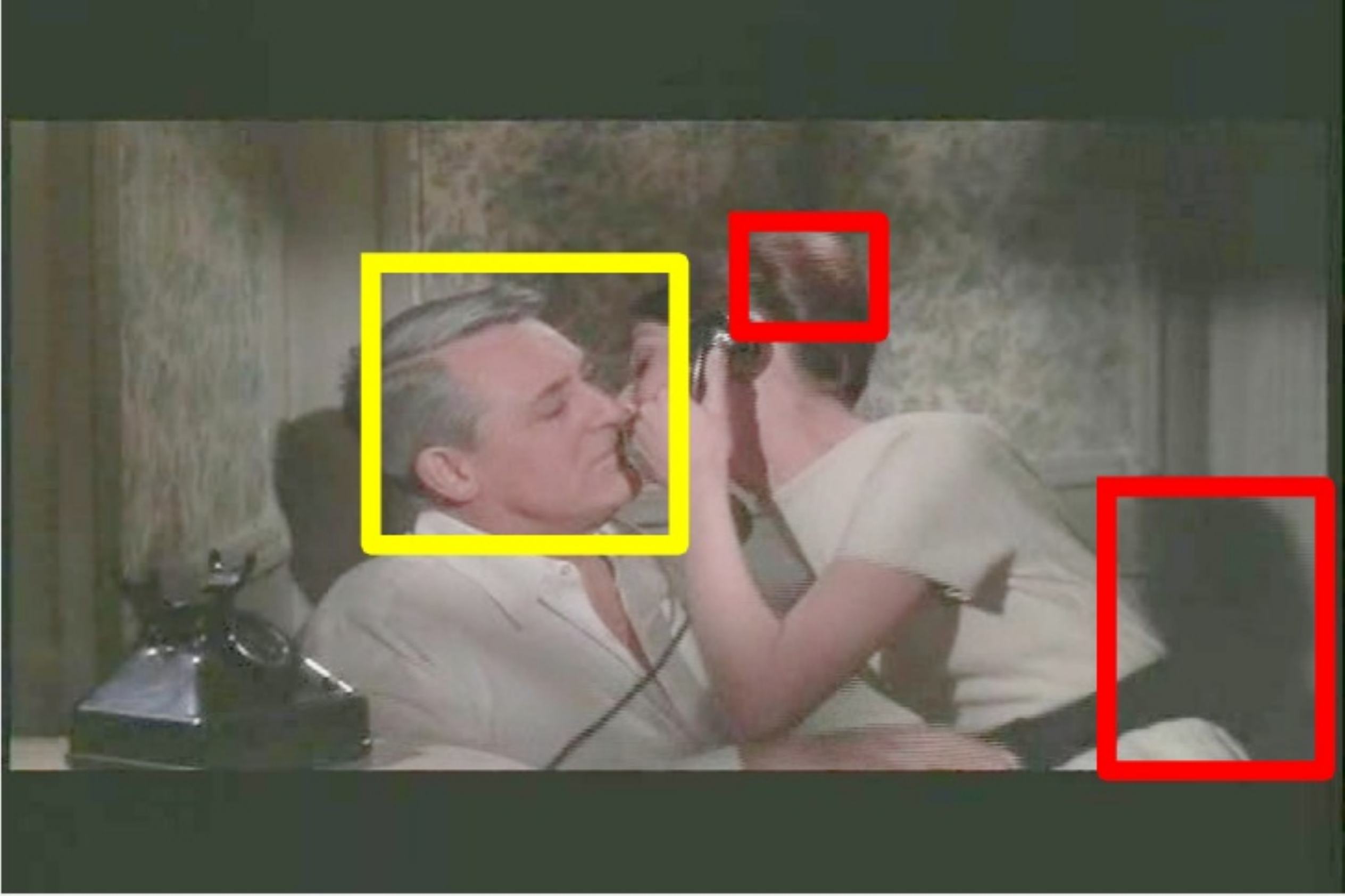}
&
\includegraphics[trim = 2mm 24mm 4mm 23mm, clip, width=0.24\linewidth]{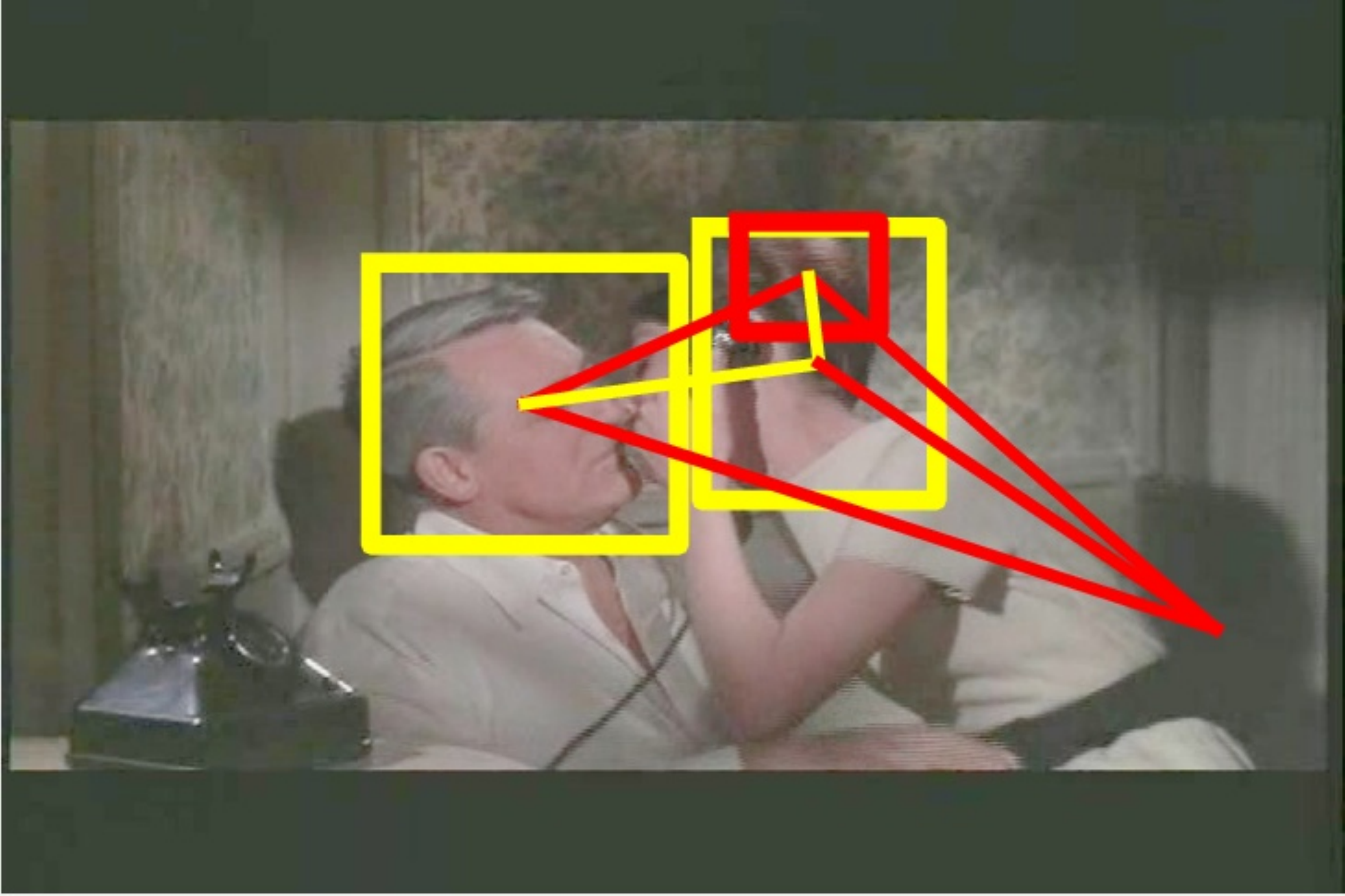} \\[-0.05cm]
\\[-0.0cm]
\end{tabular}
\caption{Qualitative results of our Pairwise model. For each frame we show the result of the local detector (left) and the result of the Pairwise model (right). For both methods we choose the threshold in such a way that precision equals recall on the validation set. We show only boxes with scores above the fixed threshold. The yellow bounding boxes correspond to correct detections, red~-- to false positive detections. For the Pairwise model we show links between candidates detected by the Local, Pairwise or both models. Links above a fixed strength threshold (attractive) are plotted with yellow, other (repulsive) links are marked by red.
}\vspace{-.2cm}
\label{fig:app:resultsPairwiseTerm}
\end{figure*}

\section{Qualitative results}
\label{sec:qual_res}
\subsection{Global model}
In this section we illustrate multi-scale grids of scores produced by the Global model (see Section~\ref{sec:globalmodel}).
Each output consists of $1\times1$, $3\times3$, $7\times7$ and $15\times15$ score grids corresponding to grids of cells with $28\times28$, $56\times56$, $112\times112$ and $224\times224$ pixels.
Figures~\ref{fig:globalModel_qual_examples} illustrates the output of the Global model for a few test examples. Note high responses at positions and
scales corresponding to human heads in the image.

\subsection{Pairwise model}
In Figure~\ref{fig:app:resultsPairwiseTerm} we provide a few qualitative results of our Pairwise model.
The bounding boxes and the links in this figure have the same meaning as the ones in Figure~4 of the main paper.
We use the same thresholds for the links and the candidates as in Figure~\ref{fig:resultsPairwiseTerm} of the main paper.

\end{document}